\documentclass[lettersize,journal]{IEEEtran}
\usepackage{amsmath,amsfonts}
\usepackage{algorithmic}
\usepackage{array}
\usepackage[caption=false,font=normalsize,labelfont=sf,textfont=sf]{subfig}
\usepackage{textcomp}
\usepackage{stfloats}
\usepackage{url}
\usepackage{verbatim}
\usepackage{graphicx}
\def\BibTeX{{\rm B\kern-.05em{\sc i\kern-.025em b}\kern-.08em
    T\kern-.1667em\lower.7ex\hbox{E}\kern-.125emX}}
\usepackage{balance}
\begin{document}
\title{Deep Reinforcement Learning for Digital Twin-Oriented Complex Networked Systems}

\author{\uppercase{Jiaqi Wen}, \uppercase{Bogdan Gabrys, and Katarzyna Musial}.\thanks{This work was supported by the Australian Research Council, “Dynamics and Control of Complex Social Networks” under Grant DP190101087.}}


\maketitle

\begin{abstract}
The Digital Twin Oriented Complex Networked System (DT-CNS) aims to build and extend a Complex Networked System (CNS) model with progressively increasing dynamics complexity towards an accurate reflection of reality -- a Digital Twin of reality. Our previous work proposed evolutionary DT-CNSs to model the long-term adaptive network changes in an epidemic outbreak. This study extends this framework by proposeing the temporal DT-CNS model, where reinforcement learning-driven nodes make decisions on temporal directed interactions in an epidemic outbreak. We consider cooperative nodes, as well as egocentric and ignorant "free-riders" in the cooperation. 
We describe this epidemic spreading process with the Susceptible-Infected-Recovered ($SIR$) model and investigate the impact of epidemic severity on the epidemic resilience for different types of nodes. Our experimental results show that (i) the full cooperation leads to a higher reward and lower infection number than a cooperation with egocentric or ignorant "free-riders"; (ii) an increasing number of "free-riders" in a cooperation leads to a smaller reward, while an increasing number of egocentric "free-riders" further escalate the infection numbers and (iii) higher infection rates and a slower recovery weakens networks' resilience to severe epidemic outbreaks. These findings also indicate that promoting cooperation and reducing "free-riders" can improve public health during epidemics.
\end{abstract}


\section{Introduction}
 \label{model}

\begin{figure*}[htp]
    \centering
    \includegraphics[width=4.5 in]{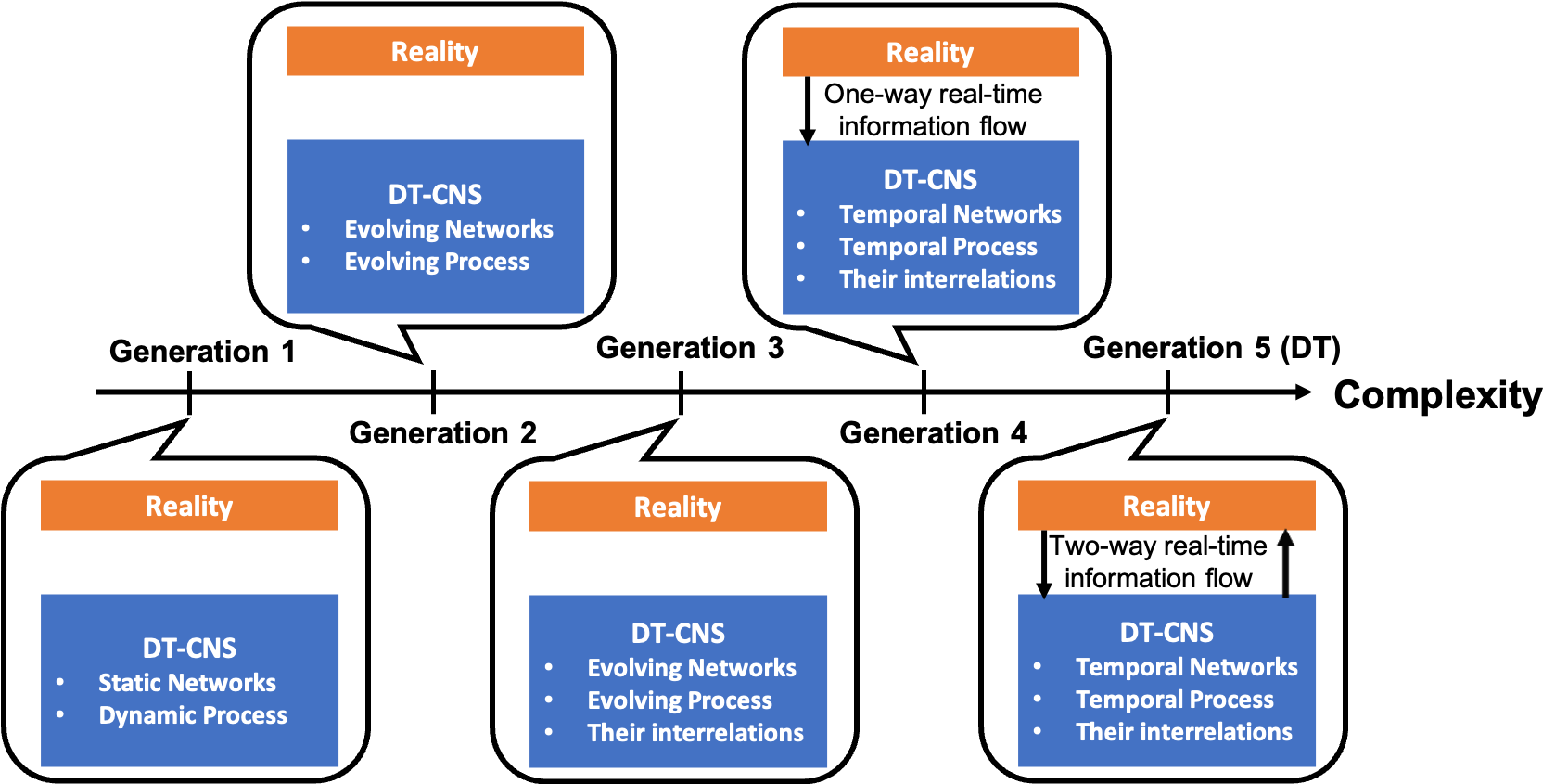}
    \caption{Generations of DT-CNSs, including generation 1: dynamic process on static networks, generation 2: evolving dynamic process on evolving networks, generation 3: evolving dynamic processes on evolving networks with interrelations between them, generation 4: temporal dynamic processes on temporal networks with interrelations between them and the acquisition of real time information, and generation 5 (a DT): Temporal dynamic processes on temporal networks with interrelations between them, as well as the real time two-way feedback between the reality and the CNSs, enabling an idealised state required by a DT.}
    \label{generation}
\end{figure*}

Accurate modelling of Complex Networked Systems (CNSs) that involves real-time interactions is crucial for addressing societal challenges such as gender inequality, crime, and epidemics. For example, organized crime and pandemic outbreak can cause significant economic losses. An advanced CNS model that mimics real-world social interactions, could help policymakers and authorities simulate various scenarios and develop better strategies to address these pressing issues. Thefore, our previous work proposed a new modelling paradigm: Digital Twin-Oriented Complex Networked Systems (DT-CNSs)~\cite{wen2022towards}, which aims to build and extend CNSs with increasing complexity of generations towards the ultimate goal -- a Digital Twin (DT) of real networked systems (Fig.~\ref{generation}).

As shown in Fig.~\ref{generation}, the complexity of generations of DT-CNSs depends on the evolvability of the dynamics in DT-CNSs, interrelations between the dynamics in DT-CNSs, as well as the real-time interplay between the DT-CNSs and the reality. From generation 1 to 5, the temporal scale of network representation and modelling transforms to be more instantaneous, ranging from static (no change), evolving (slow/temporal changes captured in snapshots) to temporal (temporal changes captured in real time) \cite{wen2022towards}. The generations of DT-CNSs, as they approach a DT in generation 5, model the temporal changes in networks and dynamic processes, together with their interrelations, while allowing for the real-time interplay between reality and the DT-CNSs \cite{wen2022towards}. Under this conceptual framework, we proposed an extendable DT-CNSs modelling framework for complex social networks by introducing heterogeneous node features (a.k.a node attributes) and nodes' preferences to create relationships, while allowing these preferences to evolve under the impact of dynamic processes towards rewarding interactions \cite{wen2023dtcns,wen2024evolutionary}. However, the proposed modelling framework builds DT-CNSs in generation 1, 2 and 3. Its extension towards generation 4 and 5 poses a challenge in data collection, processing and model updates considering the real-time information (Fig.~\ref{generation}). 
In this study, we progress this the DT-CNS modelling framework by modelling temporal networks and the temporal dynamic process on the networks, which brings the DT-CNS closer to the ultimate goal of a DT.

Reinforcement learning (RL) allows the agent to make decisions, observe the results, and then automatically adjust its strategy to achieve the optimal policy~\cite{luong2019applications}. To address the scalability and efficiency issues of traditional RL, 
researchers proposed to use Deep Reinforcement Learning (DRL), leveraging Deep Neural Networks to enhance learning speed and performance~\cite{mnih2013playing,luong2019applications}. Current studies have employed DRL to implement multiple tasks in CNSs, such as influencing maximisation~\cite{ma2022influence,yang2024balanced,ali2020addressing,jiang2023deep}, key-player identification~\cite{fan2020finding,zeng2023leveraging}, network topology optimisation~\cite{li2023network} and intelligent routing~\cite{liu2019intelligent,sun2021enabling,ali2021nedrl}. Some studies employed deep reinforcement learning to automatically make a decision in social networks, such as network resource allocation~\cite{he2018trust}, rumour mitigation~\cite{su2024rumor} and influence maximisation~\cite{li2022piano}. However, none of these studies employ deep reinforcement learning to optimally drive nodes' cooperation and free-riding behaviours in an epidemic outbreak.



This study proposes the temporal Digital Twin-Oriented Complex Networked System (DT-CNS) model driven by reinforcement learning algorithm. Our previous work ~\cite{wen2024evolutionary} enables to model evolutionary DT-CNSs regarding the evolving networks, a dynamic process on the networks and their interrelated changes.
This study extends this framework by introducing reinforcement learning-driven nodes to make decisions on temporal directed interactions in an epidemic outbreak. Under this framework, we consider nodes' heterogeneous features and changeable connection preferences, which combine the effects of preferential attachment and homophily. We consider three types of nodes: (i) cooperative nodes who 
maximise their total reward using a collective mind driven by reinforcement learning; (ii) egocentric nodes 
who maximise the individual reward under the assumption of being the only "free-rider" in the cooperation and (iii) the ignorant nodes who make random decisions. Given an epidemic outbreak, the healthy (sick) node, takes the risk of getting infected or infecting the neighbours via interactions. We describe this epidemic spreading process with the "susceptible-infected-recovered" ($SIR$) model and investigate the impact of epidemic severity on the epidemic resilience for different types of nodes. 
Our experiment results shows that (i) full cooperation leads to higher reward and lower infection number than cooperation with egocentric or ignorant free-riders; (ii) an increasing number of "free-riders" in cooperation leads to smaller reward, while an increasing number of egocentric "free-riders" further escalate the infection numbers and (iii) higher infection rates and slower recovery weakens networks' resilience to severe epidemic outbreaks.

Overall, this study contributes in the following aspects:
\begin{itemize}
     \item we propose a temporal DT-CNS framework related to nodes' temporal decisions on their preferences for connecting with others in an epidemic outbreak.

     \item we create heterogeneous preference mutation styles that characterise nodes' cooperation and free-riding behaviours.
     
    \item we introduce deep reinforcement learning algorithms to drive nodes' temporal decisions.

    \item we find that cooperation enhances rewards and reduces infections, while an increase in "free-riders" and higher infection rates with slower recovery diminish the network's epidemic resilience.
\end{itemize}

The rest of this study is structured as follows: Section~\ref{Model} presents the methodology of building a temporal DT-CNS. Following this, Section~\ref{Results} builds and evaluates the temporal DT-CNSs under more severe epidemic outbreaks. Finally, we conclude with Section~\ref{Conclusion}.

\section{Model}
\label{Model}
\subsection{Components}


Temporal directed network at time $t$ can be represented as $\mathbf{G}_t$ and is composed of two elements.
\begin{equation}
    \mathbf{G}_t(\mathbf{V}_t,\mathbf{E}_t)
\end{equation}

$\mathbf{V}_t$ represents the set of nodes. In this study, we assume a fixed number of nodes ($N$ nodes) for each temporal network $\mathbf{G}_t$. The set of nodes is defined as
\begin{equation}
\mathbf{V}_t=\{v_{0,t},v_{1,t},\cdots,v_{i,t},\cdots,v_{N,t}\},
\end{equation}

$\mathbf{E}_t$ represents the set of edges, which refer to the directed connections between the nodes. The set of edges can be defined as
\begin{equation}
\label{edgeSet}
   \mathbf{E}_t = \{e_{ij,t}|v_{i,t},v_{j,t}\in \mathbf{V}_t, i\neq j\},
\end{equation}
where $e_{ij,t}$ represents a directed connection from node $v_{i,t}$ to node $v_{j,t}$ at time $t$. 
We assume no self-links in this study ($e_{ii,t}$ does not exist).

\subsection{Dynamics}

\subsubsection{\textbf{Agents}}
In this study, the nodes also represent the agents who act in response to the interaction reward. They can be divided into a set of agents (nodes) driven by different action styles, including the ignorant agents: $\mathbf{V}^{Ign}=\{v_{i,t}^{Ign}\}$, the set of egocentric agents: $\mathbf{V}^{Ego}=\{v_{i,t}^{Ego}\}$, the set of cooperative agents: $\mathbf{V}^{Cop}=\{v_{i,t}^{Cop}\}$. The nodes and the agents are interchangeable concepts in this study.
\begin{equation}
    \mathbf{V} = \mathbf{V}^{Ign}\cup\mathbf{V}^{Ego} \cup \mathbf{V}^{Cop},
\end{equation}
where cooperative nodes cooperate to maximise the average value of the overall interaction reward. The egocentric nodes and the ignorant nodes are "free-riders" of the cooperation, as they assume the other nodes are cooperative so they betray the cooperation by ignorant or egocentric actions. The ignorant nodes keep ignorant of the interaction reward and randomly act. The egocentric nodes only consider individual interactions and their respective individual rewards. 

In the network evolution and the epidemic spreading processes, each agent observes the network states, makes decisions on temporal social contact and calculates the reward according to their action styles. The temporal changes in interaction patterns and infection patterns influence the interaction reward and reinforce the learning process of the agents.

\subsubsection{\textbf{States}}
We represent the states of the whole system at time $t$ with $\mathbb{S}_t$, which includes the states of the network $\mathbf{S}^{Net}_t$ and the epidemic process $\mathbf{S}^{Epi}_t$,
\begin{equation}
    \mathbb{S}_t = \mathbf{S}^{Net}_t\bigcup\mathbf{S}^{Epi}_t
\end{equation}

In this study, we model the epidemic spreading process with the "susceptible-infected-recovered" ($SIR$) model~\cite{volz2007susceptible}. 
The state $\mathbf{S}^{Epi}_{t}$ of an epidemic outbreak can be defined as

\begin{equation}
    \mathbf{S}^{Epi}_{t}(\mathbf{V}^{seed}_t,\zeta^s_t, \tau^r_t)
\end{equation}
where $\mathbf{V}^{seed}_t$ represents the seed set for the epidemic spread. $\zeta^s_t$ denotes the infection rate of the epidemic spread at time $t$. The node $v_{i,t}$ gets infected at a probability of $\zeta^s_t$ given a single interaction with an infected node and, once infected, recovers after a time length of $\tau^r_t$ before another interaction occurs.

The healthy node's infection probability can be calculated as 
\begin{equation}
    \left\{
    \begin{array}{ll}
    pr(\beta_{i,t}=1|\beta_{i,t-1}=0)&= \prod\limits_{v_{j,t}\in V_{t}}(1-\zeta^s_t)^{I_{ij,t}\beta_{j,t}}
    \end{array}
    \right.
\end{equation}
where $pr(\beta_{i,t}=1|\beta_{i,t-1}=0)$ represents the probability of a healthy node $v_{i,t}$ ($\beta_{i,t-1}=0$) getting infected ($\beta_{i,t}=1$). Its value depends on the epidemic transmissibility $\mathrm{pr}^s_t=\zeta^s_t$, the interactions $I_{ij,t}$ with each neighbour $v_{j,t}$ and the neighbour's infected condition $\beta_{j,t}=1$. 

The network state $\mathbf{S}^{Net}_t$ covers the states of each types of nodes and the state of the social connections.
\begin{equation}
    \mathbf{S}^{Net}_t = 
    \bigcup
    \limits_{v_{i,t}^{Ign}\in\mathbf{V}^{Ign}} \mathbf{S}_{v_{i,t},t}^{Ign}\bigcup
    \limits_{v_{i,t}^{Ego}\in\mathbf{V}^{Ego}} \mathbf{S}_{v_{i,t},t}^{Ego}\bigcup
    \limits_{v_{i,t}^{Cop}\in\mathbf{V}^{Cop}} \mathbf{S}_{v_{i,t},t}^{Cop}
\end{equation}
where $\mathbf{S}_{v_{i,t},t}^{Ign}$, $\mathbf{S}_{v_{i,t},t}^{Ign}$, $\mathbf{S}_{v_{i,t},t}^{Ego}$ and $\mathbf{S}_{v_{i,t},t}^{Cop}$ each represents the state of the ignorant node $v_{i,t}^{Ign}$, ignorant node $v_{i,t}^{Ign}$, egocentric node $v_{i,t}^{Ego}$ and cooperative node $v_{i,t}^{Cop}$ at time $t$. $\mathbf{S}_{E,t}$ represents the state of social connections between the nodes at time $t$.

The state $\mathbf{S}_{v_{i,t},t}^{Ign}$ for the ignorant node $v_{i,t}^{Ign}$ can be defined as
\begin{equation}
    \mathbf{S}_{v_{i,t},t}^{Ign}(\mathbf{f}^{Ign}_{i,t},sDNA^{Ign}_{i,t},\beta^{Ign}_{i,t},\mathrm{R}^{Ign}_{i,t},\$^{Ign}_{i,t},\hat{\$}^{Ign}_{i,t},\mathrm{Ri}^{Ign}_{i,t})
\end{equation}
where $\mathbf{f}^{Ign}_{i,t}$ and $sDNA^{Ign}_{i,t}$ each represent the features and the related preferences of node $v_{i,t}^{Ign}$ at time $t$. $\beta_{i,t}^{Ign}$ is a binary value that represents the node's health condition at time $t$. $\beta^{Ign}_{i,t}=0$ indicates the healthy condition and $\beta^{Ign}_{i,t}=1$ indicates the node's infected condition. The node $v_{i,t}^{Ign}$ suffers the interaction risk $\mathrm{Ri}^{Ign}_{i,t}$ in an epidemic outbreak and obtains the interaction reward $\mathrm{R}^{Ign}_{i,t}$ from interacting with others at time $t$. The social capital embedded in nodes' interactions is denoted by $\$^{Ign}_{i,t}$ and constrained by the social capital limit $\hat{\$}^{Ign}_{i,t}$. The node $v_{i,t}^{Ign}$ interacts under the social capital limit $\hat{\$}_{i,t}$ for objectives related to interaction reward $\mathrm{R}_{i,t}^{Ign}$. The social capital limit refers to people's maximum investment, due to personal capacity or social restrictions, in social interactions for expected returns \cite{chang2011social,lin2017building}.


The node's preferences $sDNA^{Ign}_{i,t}$ for the related features $\mathbf{f}^{Ign}_{i,t}$ is defined as
\begin{equation}
    sDNA^{Ign}_{i,t}(\mathbf{p}^{Ign}_{i,t},\mathbf{w\_p}^{Ign}_{i,t},\mathbf{h}^{Ign}_{i,t},\mathbf{w\_h}^{Ign}_{i,t},\mathrm{c}^{Ign}_{i,t},\mathrm{w\_c}^{Ign}_{i,t})
\end{equation}
where $\mathbf{p}{i,t}^{Ign}$ and $\mathbf{h}^{Ign}{i,t}$ are preference vectors. These vectors represent preferences for features and feature differences. The values of the preferences are denoted as $-1$, $0$, and $1$, corresponding to negative, neutral, and positive preferences. 
These two vectors are followed with the same length weighting vectors $\mathbf{w\_p}^{Ign}_{i,t}$ and $\mathbf{w\_h}^{Ign}_{i,t}$, which include the weight of preference with value within $(0,1]$. Similarly, $\mathrm{c}^{Ign}_{i,t}$ determines the negative, zero and positive preference for connecting with nodes with common friends by $-1$, $0$ and $1$, followed by the weight of preference $\mathrm{w\_c}^{Ign}_{i,t}$. 

The epidemic risk $\mathrm{Ri}^{Ign}_{i,t}$ of a healthy node $v_{i,t}^{Ign}$ at time $t$ is determined as
\begin{equation}
   \mathrm{Ri}_{i,t}^{Ign}= \left\{
   \begin{array}{ll}
     1-\prod\limits_{v_{j,t}\in V_{t}}(1-\zeta^s)^{I_{ij,t}\beta_{j,t}} &\textit{if $\beta_i^{Ign}=0$}\\
  1-\prod\limits_{v_{j,t}\in V_{t}}(1-\zeta^s)^{I_{ij,t}(1-\beta_{j,t})}&\textit{if $\beta_i^{Ign}=1$}\\
   \end{array}
    \right.
\end{equation}
where $\beta_i^{Ign}=0$ and $\beta_i^{Ign}=1$ each represents the healthy and the infected states of the node $v_{i,t}^{Ign}$. $1-\prod\limits_{v_{j,t}\in V_{t}}(1-\zeta^s)^{I_{ij,t}\beta_{j,t}}$ represents the probability of a healthy node $v_{i,t}^{Ign}$'s infection given its interactions $I_{ij,t}$ with its neighbour $v_{j,t}$ and the neighbour's infected condition $1-\beta_{j,t}=1$. $\prod\limits_{v_{j,t}\in V_{t}}(1-\zeta^s)^{I_{ij,t}(1-\beta_{j,t})}$ represents the probability of an epidemic spread from the infected node $v_{i,t}^{Ign}$ to its neighbours at time $t$.


The state of the directed connection $e_{ij,t}$ from node $v_{i,t}$ to node $v_{j,t}$ at time $t$ is defined as:
\begin{equation}
\label{edgeState}
    \mathbf{S}_{e_{ij,t}}(I_{ij,t},w^I_{ij,t},B_{ij,t},w^B_{ij,t})
\end{equation}
where $I_{ij,t}$ represents the no interaction or interaction from node $v_{i,t}$ to node $v_{j,t}$ at time $t$ with $0$ and $1$. $B_{ij,t}$ determines whether this node pair get closely bonded with $0$ and $1$, each indicating the unbonded and the bonded node pair. These two factors are followed by the respective intensity factor $w^I_{ij,t}$ and $w^B_{ij,t}$ a.k.a relationship strength.




\subsubsection{\textbf{Observation}}
The nodes observe the related social contact and epidemic spread. The observation $\mathbf{O}_{v_{i,t},t}^{Ign}$ of an ignorant node $v_{i,t}^{Ign}$ at time $t$ can be defined as
\begin{equation}
    \mathbf{O}_{v_{i,t},t}^{Ign}(\mathbf{S}^{Epi}_{t},\mathbf{S}_{v_{i,t},t}^{Ign},\mathbb{S}_{e_{i\cdot,t}}^{Ign},\mathbb{S}_{e_{\cdot i,t}}^{Ign})
\end{equation}
where $\mathbf{S}^{Epi}_{t}$ represents the state of the epidemic spread at time $t$. $\mathbf{S}_{v_{i,t},t}^{Ign}$ represents the state of the ignorant node $v_{i,t}^{Ign}$ at time $t$.
$\mathbb{S}_{e_{i\cdot,t}}^{Ign}$ and $\mathbb{S}_{e_{\cdot i,t}}^{Ign}$ each represents the states of the directed interactions related to the ignorant node $v_{i,t}^{Ign}$ until the observation time at $t$. They can be defined as
\begin{equation}
\left\{
    \begin{array}{l}
\mathbb{S}_{e_{i\cdot,t}}^{Ign} = \bigcup\limits_{v_{j,t}\in\mathbf{V},\tau\leq t} \mathbf{S}_{e_{ij},\tau}     \\[6mm]
\mathbb{S}_{e_{\cdot i,t}}^{Ign} = \bigcup\limits_{v_{j,t}\in\mathbf{V},\tau\leq t} \mathbf{S}_{e_{ji},\tau}   
    \end{array}
    \right.
\end{equation}
where $\bigcup\limits_{v_{j,t}\in\mathbf{V},\tau\leq t} \mathbf{S}_{e_{ij},\tau}$ represents the set of states for the directed interaction from the ignorant node $v_{i,t}^{Ign}$ to its neighbour $v_{j,t}$ at time $\tau\leq t$ before the observation time at $t$. Similarly, $\bigcup\limits_{v_{j,t}\in\mathbf{V},\tau\leq t} \mathbf{S}_{e_{ji},\tau}$ represents the set of states for the directed interaction from the node $v_{j,t}$ to the ignorant node $v_{i,t}^{Ign}$ at time $\tau\leq t$ before the observation time at $t$.

the observation $\mathbf{O}_{v_{i,t},t}^{Ego}$ for the egocentric node $v_{i,t}^{Ego}$ and the observation $\mathbf{O}_{v_{i,t},t}^{Cop}$ for the cooperative node $v_{i,t}^{Cop}$. Based on these observations, the node make social contact (See section~\ref{action}).


\subsubsection{\textbf{Action}}
\label{action}
The agents (nodes) mutate their preferences for features and feature differences and make social contact based on the observations of the system. The action set $\mathbb{A}_{i,t}^{Ign}$ for the ignorant node $v_{i,t}^{Ign}$ at time $t$ can be defined as 
\begin{equation}
    \mathbb{A}_{i,t}^{Ign}(\mathbf{p}_{i,t}^{Ign},\mathbf{h}_{i,t}^{Ign}),
    \label{eqAction}
\end{equation}
where $\mathbf{p}_{i,t}^{Ign}$ represents the preference of node $v_{i,t}^{Ign}$ for given features. $\mathbf{h}_{i,t}^{Ign}$ represents the preference of node $v_{i,t}^{Ign}$ for feature differences. 



The nodes take actions related to the preferences for features based on the observations of the changes in social connections. This results in temporal changes in the states $\mathbf{S}_{e_{ij},t}(I_{ij,t},w^I_{ij,t})$ (equation~\ref{edgeState}) of directed interactions between node pairs, including the existence of directed interaction $I_{ij,t}$ from node $v_{i,t}$ to node $v_{j,t}$ and the corresponding interaction intensity $w^I_{ij,t}$.

Taking the ignorant node $v_{i,t}^{Ign}$ as an example, the temporal directed interaction formation process between node $v_{i,t}^{Ign}$ and any other node $v_{j,t}$ can be defined as 
\begin{equation}
    I_{ij,t}=\left\{
    \begin{array}{ll}
     1   &\textit{if $\Pi_{ij,t}^{Ign}>\Pi^{Ign,*}_{i,t}$ and $\$^{Ign}_{i,t}<\hat{\$}^{Ign}_{i,t}$}  \\[3mm]
     0    & \textit{if else}  \\
    \end{array}
    \right.
\end{equation}
where $\Pi_{ij,t}^{Ign}$ represents the score of the directed interaction from node $v_{i,t}^{Ign}$ to any node $v_{j,t}$ based on the evaluation from the perspective of node $v_{i,t}^{Ign}$. $\Pi^{Ign,*}_{i,t}$ represents the interaction score threshold of node $v_{i,t}^{Ign}$ at time $t$. $\$^{Ign}_{i,t}$ represents the social capital invested by node $v_{i,t}^{Ign}$ in its directed interactions. $\hat{\$}^{Ign}_{i,t}$ denotes the social capital limit imposed on node $v_{i,t}^{Ign}$ for its directed interactions. The node $v_{i,t}^{Ign}$ takes a directed interaction $I_{ij,t}$ when the interaction score $\Pi_{ij,t}^{Ign}$ is higher than the interaction score threshold $\Pi^{Ign,*}_{i,t}$.

Specifically, the interaction score threshold $\Pi_{i,t}^{Ign,*}$ for the ignorant node $v_{i,t}^{Ign}$  depends on its health status $\beta_{i,t}^{Ign}$.
\begin{equation}
    \Pi^{Ign,*}_{i,t} = \eta^{\beta_{i,t}^{Ign}}\Pi^{Ign,*}
\end{equation}
where $\eta>1$ denotes a penalty indicator. It increases the interaction score threshold when the ignorant node is infected $\beta_{i,t}^{Ign}=1$.

The interaction intensity $w^I_{ij,t}$ for the directed interaction from node $v_{i,t}^{Ign}$ to node $v_{j,t}$ is defined as
\begin{equation}
    w^I_{ij,t}=\left\{
    \begin{array}{ll}
    b+\alpha (\Pi_{ij,t}^{Ign}-\Pi_{i,t}^{Ign,*}) &\textit{if $I_{ij,t}=1$}  \\[3mm]
        0 & \textit{if $I_{ij,t}=0$}
    \end{array}
    \right.
\end{equation}
where $\Pi_{ij,t}^{Ign}$ represents the interaction score for the directed interaction from node $v_{i,t}^{Ign}$ to node $v_{j,t}$. $b>0$ and $\alpha>0$ represent the base and scaling parameters for the interaction intensity. The interaction intensity $w^I_{ij,t}$ scales up with the corresponding interaction score $\Pi_{ij,t}^{Ign}$ given an interaction $I_{ij,t}$ from node $v_{i,t}^{Ign}$ to node $v_{j,t}$.

The interaction score $\Pi_{ij,t}^{Ign}$ of the directed interaction from the ignorant node $v^{Ign}_{i,t}$ to the node $v_{j,t}$ depends on the unilateral evaluation of node $v^{Ign}_{i,t}$ based on the features and related preferences. The score $\Pi^{Ign}_{ij,t}$ that node $v^{Ign}_{i,t}$ assigns to its directed interaction with $v_{j,t}$ is calculated based on the preferential attachment and homophily effects, each represented as $\pi\_h^{Ign}_{ij,t}$ and $\pi\_p^{Ign}_{ij,t}$.
\begin{eqnarray}
\Pi^{Ign}_{ij,t} = \left\{ 
\begin{array}{ll}
\frac{1}{2}(\pi\_h^{Ign}_{ij,t}+ \pi\_p^{Ign}_{ij,t}) + \epsilon_{ij,t} & \textrm{\footnotesize{if $v^{Ign}_{i,t}$ and $v_{j,t}$ encounters.}}\\[3mm] 
0 & \textrm{\footnotesize{if else.}}\\
\end{array}
\right.
\end{eqnarray}
where $v^{Ign}_{i,t}$ evaluates its directed interaction with others by averaging the homophily score $\pi\_h^{Ign}_{ij,t}$ and preferential attachment score $\pi\_p^{Ign}_{ij,t}$ based on their encounters, ranging between $[0,1]$. In contrast, the subjective node assigns $0$ to this unilateral evaluation of the directed interaction without an encounter. $\epsilon_{ij,t}\sim N(0,0.01^2)$ is a random interference that follows a normal distribution $N(0,0.01^2)$, which characterises the interference in nodes' evaluation on interactions~\cite{wen2024evolutionary}.

The homophily effect $\pi\_h^{Ign}_{ij, t}$ describes the preference of the ignorant node $v_{i,t}^{Ign}$ for connecting with similar or dissimilar others~\cite{wen2024evolutionary}. 
\begin{equation}
    \pi\_h^{Ign}_{ij,t} = |\mathbf{f}_{i,t}^{Ign}-\mathbf{f}_{j,t}|^\tau (\mathbf{p}_{i,t}^{Ign} \odot \mathbf{w\_p}^{Ign}_{i,t})
\end{equation}
where $\mathbf{f}_{i,t}^{Ign}$ and $\mathbf{f}_{j,t}$ represent the feature vectors for ignorant node $v_{i,t}^{Ign}$ and any other node $v_{j,t}$. $\mathbf{h}_{i,t}^{Ign}$ and $\mathbf{w\_h}^{Ign}_{i,t}$ represent the preference vectors of the ignorant node $v_{i,t}^{Ign}$ for feature differences between the ignorant node $v_{i,t}^{Ign}$ and the node $v_{j,t}$.

The preferential attachment effect $\pi\_p^{Ign}_{ij,t}$ describes the node's preference for connecting with other nodes with specific features~\cite{wen2024evolutionary}.
\begin{equation}
\footnotesize
\begin{aligned}
 \pi\_p^{Ign}_{ij,t} = \mathbf{f}_{j,t}^\tau (\mathbf{p}_{i,t}^{Ign} \odot \mathbf{w\_p}^{Ign}_{i,t})
\end{aligned}
\end{equation}
where $\mathbf{f}_{j,t}$ denotes the feature vector of node $v_{j,t}$. $\mathbf{p}_{i,t}^{Ign}$ and $\mathbf{w\_p}^{Ign}_{i,t}$ represent the preference vectors of the ignorant node $v_{i,t}^{Ign}$ for features.


The node pair $v_{i,t}^{Ign}$ and $v_{j,t}$ gets closely bonded at time $t$ given a one-way pairwise interaction within time $(t-\Delta t_i, t]$ and an another-way pairwise interaction as response at time $t$.
\begin{equation}
B_{ij,t}=\left\{
\begin{array}{ll}
1     & \textit{if $\exists I_{ij,t}^{Ign}*I_{ji,\tau_j}=1$, $\tau_j \in (t-\Delta t_i, t]$,} \\[3mm]
1     & \textit{if $\exists I_{ij,\tau_i}^{Ign}*I_{ji,t}=1$, $\tau_i\in (t-\Delta t_i, t]$,} \\[3mm]
0     & \textit{if $\forall I_{ij,\tau_i}^{Ign}*I_{ji,\tau_j}=0$, $\tau_i,\tau_j \in (t-\Delta t_i, t]$,}
\end{array}
\right.
\end{equation}
where $I_{ij,\tau_i}^{Ign}$ represents the directed interaction from node $v_{i,\tau_i}^{Ign}$ at time $\tau_i$ to node $v_{j,\tau_i}$. $I_{ji,\tau_i}^{Ign}$ represents the directed interaction from node $v_{j,\tau_i}$ at time $\tau_j$ to node $v^{Ign}_{i,\tau_j}$. 

The bonding intensity is calculated based on the average value of the directed interaction intensities for the ignorant node $v_{i,t}^{Ign}$ and the other node $v_{j,t}$.
\begin{equation}
    w^B_{ij,t} =\frac{1}{2}(\frac{\sum\limits_{\tau_i\in (t-\Delta t,t]}w^I_{ij,\tau_i}}{\sum\limits_{\tau_i\in (t-\Delta t,t]}I_{ij,\tau_i}}+ \frac{\sum\limits_{\tau_j\in (t-\Delta t,t]}w^I_{ji,\tau_j}}{\sum\limits_{\tau_j\in (t-\Delta t,t]}I_{ji,\tau_j}})
\end{equation}
where $w^I_{ij,\tau_i}$ and $w^I_{ji,\tau_j}$ each represents the directed interaction intensities. $\sum\limits_{\tau_i\in (t-\Delta t,t]}w^I_{ij,\tau_i}$ represents the sum of the directed interaction intensities from the ignorant node $v_{i,\tau_i}^{Ign}$ to node $v_{j,\tau_i}$ at time $\tau_i\in(t-\Delta t,t]$. $\sum\limits_{\tau_i\in (t-\Delta t,t]}I_{ij,\tau_i}$ represents the number of the directed interactions from the ignorant node $v_{i,\tau_i}^{Ign}$ to node $v_{j,\tau_i}$ at time $\tau_i\in(t-\Delta t,t]$. Based on that, we calculate the average interaction intensities for the directed interactions from node $v_{i,\tau_i}^{Ign}$ to node $v_{j,\tau_i}$ within the time $(t-\Delta t,t]$. The average interaction intensities for the directed interactions from node $v_{j,\tau_j}$ to the ignorant node $v_{i,\tau_j}^{Ign}$ within the time $(t-\Delta t,t]$ is similarly calculated.


\subsubsection{\textbf{Reward}}
\label{reward}

The nodes' individual rewards differ from nodes' action styles.

\paragraph{Cooperative} nodes cooperate and mutate preferences and social capital constraint to maximise overall benefit.  These cooperative nodes use a collective mind for decision making and calculate the interaction reward based on the interaction intensities and risks of all individuals. The cooperative nodes assume that all nodes are cooperative and make group decisions using a collective mind driven by the reinforcement learning algorithm. 
\begin{equation}
\begin{aligned}
    \mathrm{R}^{Cop}_{i,t} = & \frac{1}{N} \sum\limits_{v_{i,t} \in \mathbf{V_t}} \sum\limits_{v_{j,t} \in \mathbf{V_t}} w^{B}_{ij,t} B_{ij,t} (1 - \mathrm{Ri}^{H}_{i,t}) (1 - \beta_{i,t}^{Ign}) \\
    & + \frac{\delta}{N} \sum\limits_{v_{i,t} \in \mathbf{V_t}} \sum\limits_{v_{j,t} \in \mathbf{V_t}} w^{B}_{ij,t} B_{ij,t} (1 - \mathrm{Ri}^{I}_{i,t}) \beta_{i,t}^{Ign},
\end{aligned}
\end{equation}
where $w_{ij,t}^I$ represents the interaction intensity from the cooperative node $v_{i}^{Cop}$ to the node $v_{j,t}$. The reward for cooperative node $v_{i,t}^{Cop}$ is calculated as the average value of the aggregated interaction intensities of all nodes in the node set $\mathbf{V}_t$, respectively accounting for spreading and infection risks, which reduce the reward accordingly.

\paragraph{Egocentric} nodes optimise their interaction reward by social preference mutation shortly after interactions and based on their observations of their related interactions. Each egocentric node assumes itself as the only "free-rider" of the cooperation process and makes decisions towards higher individual rewards based on the reinforcement learning algorithm. Referring to \cite{gosak2021endogenous,wen2024evolutionary}, we calculate the interaction reward $\mathrm{R}^{Ego}_{i,t}$ of the egocentric node $v^{Ego}_{i,t}$ based on the interaction intensity and the risk avoidance. 
\begin{eqnarray}
 \mathrm{R}^{Egn}_{i,t} =\left\{
\begin{array}{ll}
\sum\limits_{v_{j,t}\in\mathbf{V_t}}w^{B}_{ij,t}B_{ij,t}(1-\mathrm{Ri}^{H}_{i,t}) & \textrm{if $\beta_{i,t}=0$.}\\
\delta  \sum\limits_{v_{j,t}\in\mathbf{V_t}}w^{B}_{ij,t}B_{ij,t}(1-\mathrm{Ri}^{B}_{i,t}) & \textrm{if $\beta_{i,t}=1$.}\\
\end{array}
\right.
\end{eqnarray}
where $w_{ij,t}^I$ represents the interaction intensity from the egocentric node $v_{i}^{Ego}$ to the node $v_{j,t}$. The reward for egocentric node $v^{Ego}_{i,t}$, given its healthy condition $\beta_{i,t}=1$, is dependent on the infection risk $\mathrm{R}_{i,t}^H$ and the interaction intensity $w^I_{ij,t}$. In contrast, in the infected condition, the node $v_{i,t}$'s interaction reward is directly correlated with bonding intensity $w^B_{ij,t}$, excluding the spreading risk $\mathrm{Ri}_{i,t}^I$ and depreciated by $\delta$ due to the infected condition.

\paragraph{Ignorant} style describes the active nodes who are unaware of the individual interactions and thus randomly select their social preferences related to preferential attachment and homophily effects between positive, zero and negative.
\begin{equation}
    \mathrm{R}^{Ign}_{i,t}=\mathcal{C}
\end{equation}
where $\mathcal{C}$ denotes a constant reward value for ignorant node $\mathrm{R}^{Ign}_{i,t}$.

We model the network dynamics based on the interaction reward driven by the features and related preferences under the impact of epidemic spread. Based on the set-ups of agents, agents' states, observation spaces, action spaces, and reward functions, we employ the Twin-Delayed Deep Deterministic (TD3) reinforcement learning algorithm \cite{fujimoto2018addressing} to update the social preferences and the social capital constraint towards a higher overall social reward.

\section{Results}
\label{Results}
In this section, we conduct experiments on the proposed temporal DT-CNS model. We simulate temporal networks in an epidemic outbreak and investigate the impact of various preference action styles on the networked systems. We first consider network simulations in an epidemic outbreak respectively driven by the cooperative agents (nodes), the ignorant agents and the egocentric agents. 
The cooperative nodes assume that all nodes are cooperative and make group decisions using a collective mind driven by the reinforcement learning algorithm. The ignorant nodes act randomly, while each egocentric node assumes itself as the only "free-rider" of the cooperation process and makes decisions towards higher individual rewards. We initialise the experiment set-ups in section~\ref{initialization} and present the network simulation results in section~\ref{NetSimulations}. In section~\ref{resilience}, we investigate the epidemic disaster resilience level of the heterogeneous nodes through what-if analyses of network simulations given various infection rates and infection recovery time values.

\subsection{Parameter Setup}
\label{initialization}
The temporal DT-CNS model simulates the social interactions of cooperative, ignorant and egocentric nodes based on their respective strategies. We assume $30$ nodes for network simulations and consider two node features, including the health state (binary value, $0$ for a healthy state and $1$ for an infection state) and a numeric feature that is simulated using a random uniform distribution between $0$ and $1$. We also assume no interactions between these agents at the first time step and allow these agents to interact in a directed way. The epidemic spread starts from the node denoted by $0$ and propagates over the network through mutual interactions. We identify the other parameters for the network simulations in Table~\ref{tabParam}.

\begin{table}[htp]
\scriptsize
	\centering
	\caption{Parameter Set-up of the temporal DT-CNS model.\label{tabParam}}
	\begin{tabular}{|p{165pt}|l|l|}
		\hline
Parameter & Symbol &Value \\
		\hline
Number of nodes & --& 30\\
\hline 
Episode Length & --& 100-day \\
\hline 
Social frequency & $\Delta t_{i}$& 0.8-hour\\
\hline 
Initial preferences regarding preferential attachment effect & $\mathbf{p}_{i,t_0}$ & [0,0]\\
\hline
Initial preference weights regarding preferential attachment effect  & $\mathbf{w\_p}_{i,t_0}$& [0,0] \\
\hline 
Initial preferences regarding homophily effect & $\mathbf{h}_{i,t_0}$ & [0,0]\\
\hline 
Initial preference weights regarding homophily effect  & $\mathbf{w\_h}_{i,t_0}$& [0,0] \\
 \hline 
Interaction score threshold for ignorant nodes&$\Pi^{Ign,*}_{i,t}$& 20 \\
\hline 
Interaction score threshold for cooperative nodes&$\Pi^{Cop,*}_{i,t}$& 20 \\
\hline 
Interaction score threshold for egocentric nodes&$\Pi^{Ego,*}_{i,t}$& 20 \\
\hline 
Base parameter for the interaction intensity&$b$& 0.2500 \\
\hline 
Scaling parameter for the interaction intensity&$\alpha$ &  0.1250\\
\hline 
Transmissibility & $\zeta^s_t$& 0.10\\
\hline
Recovery time  &$\tau^r$& 5\\
\hline
\end{tabular}
\end{table}

We also specify the parameter setups of the reinforcement learning algorithm when respectively training the collective mind for the group decisions of cooperative agents and the individual mind of the egocentric "free-rider" (Table~\ref{RLparam}).

\begin{table}[htp]
\scriptsize
	\centering
	\caption{Parameter settings of the reinforcement learning algorithm\label{RLparam}}
	\begin{tabular}{|l|l|l|}
		\hline
Node Type&Parameter &Value \\
		\hline
Cooperative nodes& Learning rate & 0.01\\
\cline{2-3}
&Batch size& 256 \\
\cline{2-3}
&Total time steps& 50000 \\
\cline{2-3}
&Discount factor& 0.99 \\
\cline{2-3}
&Policy noise& 0.2 \\
\cline{2-3}
&Exploration noise& 0.1 \\
\hline
Egocentric node& Learning rate & 0.01\\
\cline{2-3}
&Batch size& 256 \\
\cline{2-3}
&Total time steps& 20000 \\
\cline{2-3}
&Discount factor& 0.99 \\
\cline{2-3}
&Policy noise& 0.2 \\
\cline{2-3}
&Exploration noise& 0.1 \\
\hline
\end{tabular}
\end{table}

\subsection{Network Simulations}
\label{NetSimulations}
We conduct network simulations based on heterogeneous preference mutation styles and the parameter setups in section~\ref{initialization}. We calculate the cumulative total reward and the infection occurrence of each node type in the network simulations that are each driven by a single preference mutation style (See section~\ref{single1}) and a mixed preference mutation style (see section~\ref{mix1}). The single mutation style involves all nodes sharing the same mutation preference, while the mixed style includes a combination of cooperative nodes and free-riders.

\subsubsection{Single Preference Mutation Style}
\label{single1}

\begin{figure}[htp]
    \centering
    \subfloat[\footnotesize  Infection occurrences]{%
\includegraphics[width=0.50\columnwidth]{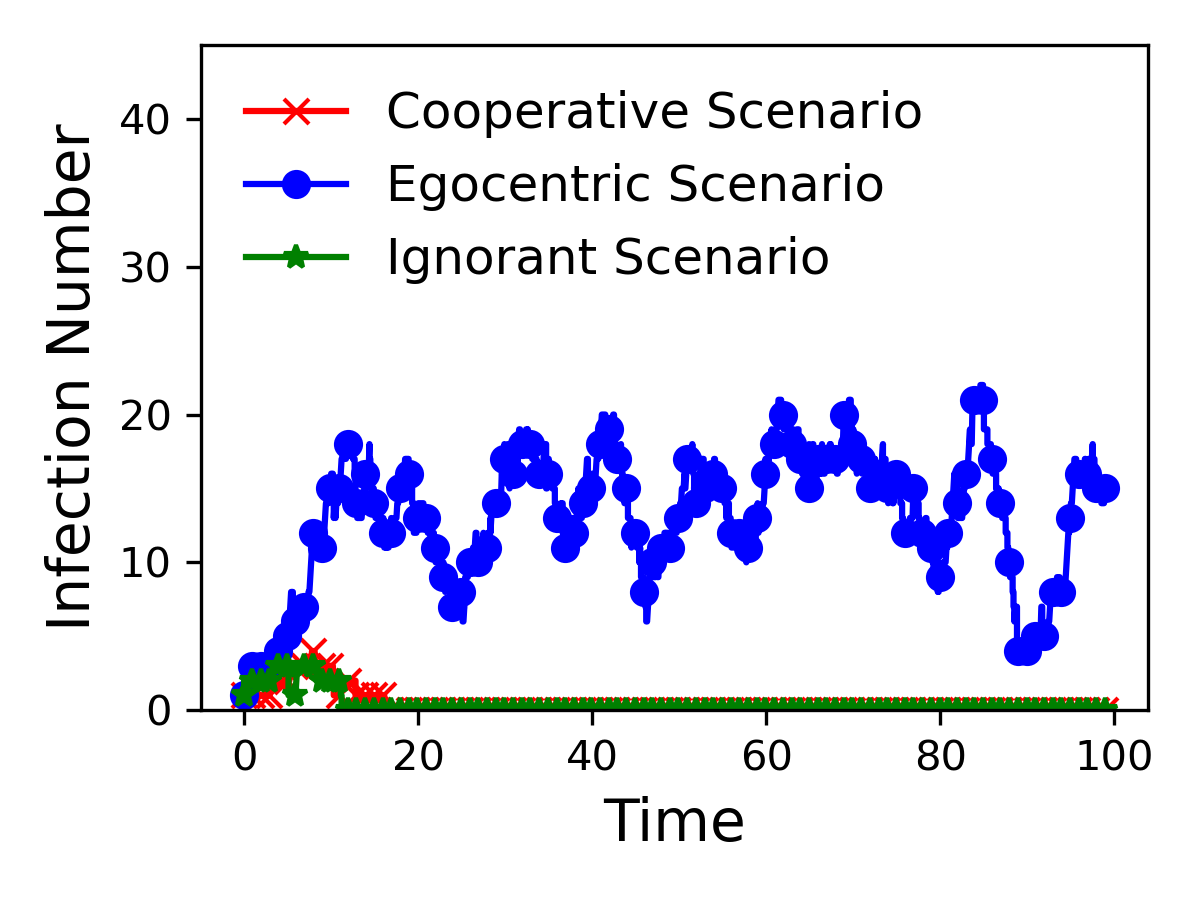}%
    }%
    \subfloat[\footnotesize  Cumulative total reward]{%
\includegraphics[width=0.50\columnwidth]{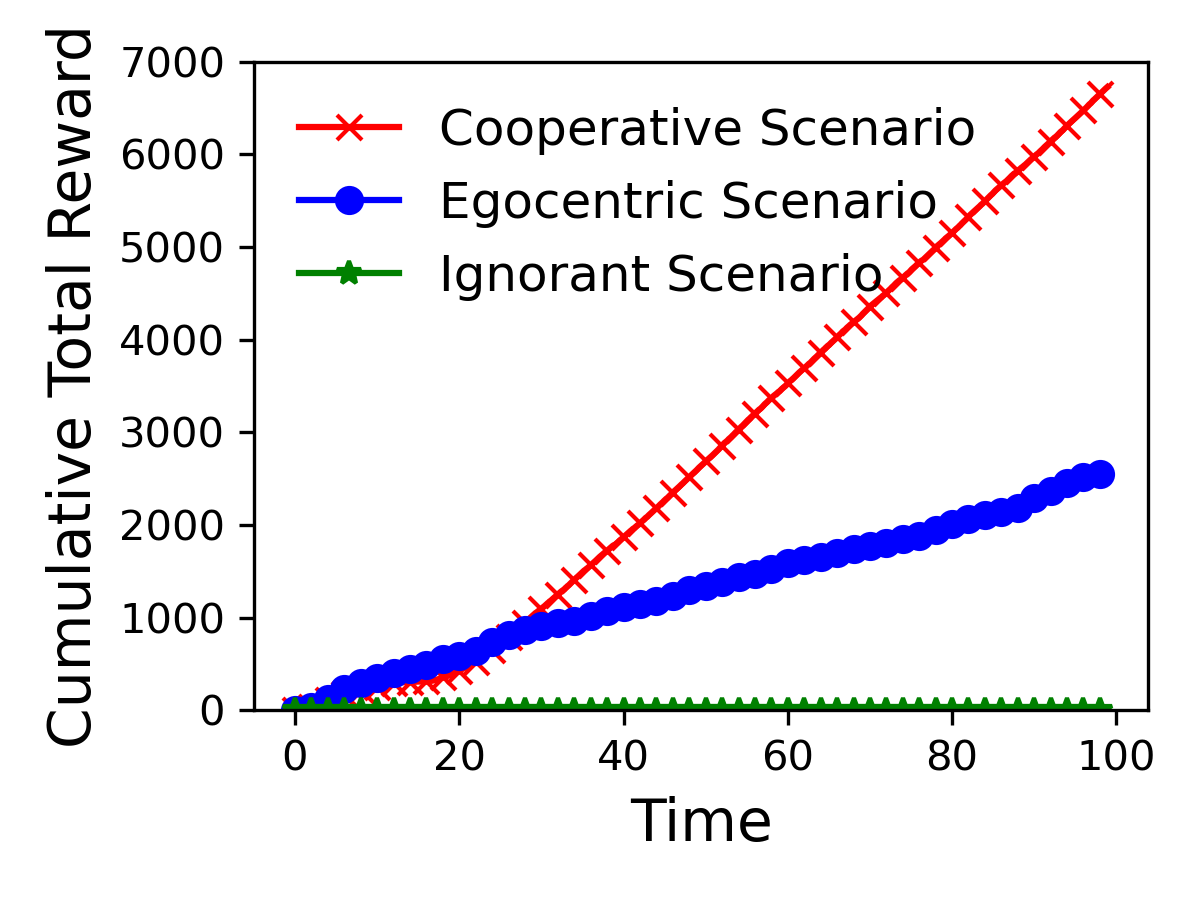}%
    }%
	\caption{The cumulative reward in cooperative, ignorant and egocentric scenarios given various setups of infection rate and recovery time.}
 \label{NetSimulationFig1}
\end{figure}

Fig.~\ref{NetSimulationFig1} shows the values of infection occurrence (Fig.~\ref{NetSimulationFig1}(a)) and the cumulative total reward (Fig.~\ref{NetSimulationFig1}(b)) over time. As shown in Fig.~\ref{NetSimulationFig1}(a), there are more infection occurrences in the network simulations driven by the egocentric nodes than those in the network simulations driven by the other node types. The infection numbers for egocentric nodes fluctuate regularly every $10$ days, with the infection number peaking at about 20 and dropping to about 7 to 10. This phenomenon might result from the recovery time of $5$ days from the infection to the healthy state. The infection occurrences of the egocentric nodes increase due to the infected node's active interactions with the healthy nodes without considering the overall benefit. The infection occurrences of these egocentric nodes decrease as they start to recover and avoid interactions with the infected others. In contrast, there are limited infection number in the network simulations driven by the cooperative and ignorant nodes in the first $20$ days and then infection vanishes afterwards. This indicates that the cooperative and ignorant nodes manage to avoid further epidemic spread. In the meanwhile, as shown in Fig.~\ref{NetSimulationFig1}(b), the cooperative nodes achieve higher cumulative total reward through cooperation using a collective mind than the other two node types in the respective network simulations. This indicates that cooperation using a collective mind enlarges the overall benefit of all involved. In addition, the egocentric nodes achieve higher cumulative total rewards than the ignorant nodes. The egocentric decisions towards higher individual benefits lead to more benefits than random interactions.

\subsubsection{Mixed Preference Mutation Style}
\label{mix1}

\begin{figure}[htp]
    \centering
\subfloat[\footnotesize  Cumulative infection number]{%
\includegraphics[width=0.50\columnwidth]{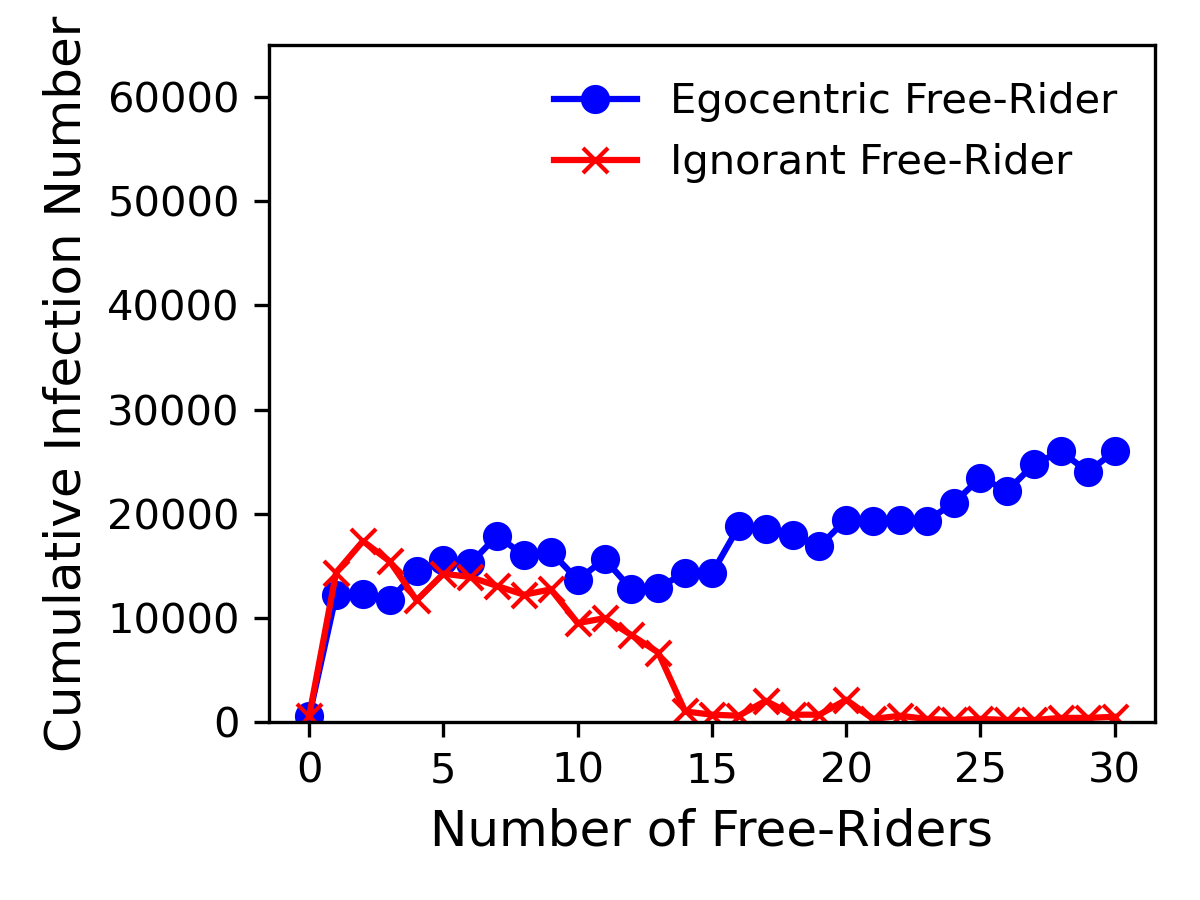}%
    }%
    \subfloat[\footnotesize  Cumulative total reward]{%
\includegraphics[width=0.50\columnwidth]{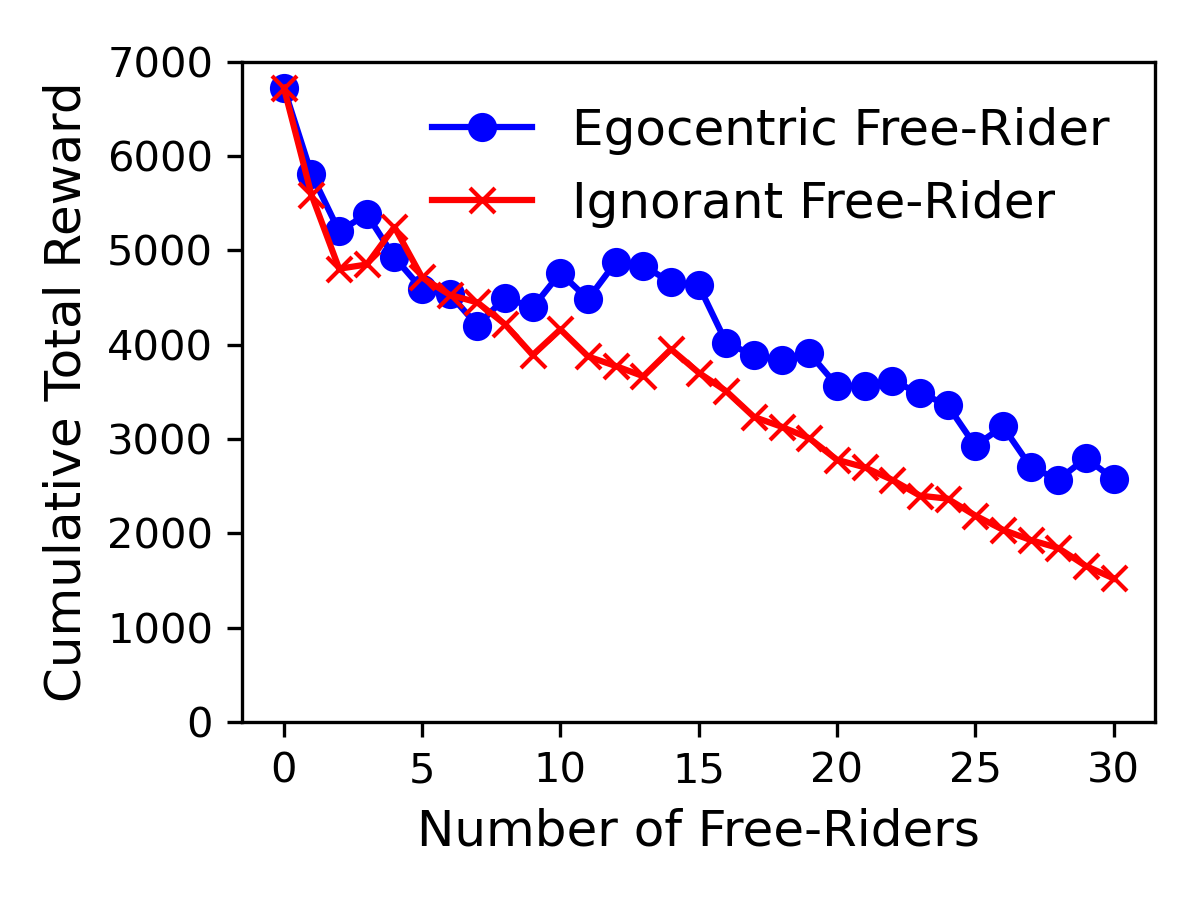}%
    }%
    \caption{The cumulative reward in the network simulations driven by the mixed preference mutation styles}
 \label{NetSimulationFig2}
\end{figure}

As shown in Fig.~\ref{NetSimulationFig2}(a), the fully cooperative scenario results in lower cumulative infection numbers compared to scenarios with egocentric or ignorant free-riders, while Fig.~\ref{NetSimulationFig2}(b) shows that it also leads to higher cumulative total rewards. This implies the introduction of "free-riders" significantly undermines the cooperation between the nodes. In addition, with an increased proportion of egocentric "free-riders" in the cooperation scenario, the cumulative infection numbers gradually increase with minor fluctuations. This indicates that the introduction of more egocentric "free-riders" intensifies the detrimental impact of free-riding on the cooperation. In contrast, more ignorant "free-riders" lead to a smaller cumulative interaction number but lower cumulative total reward. This is because an increasing number of ignorant "free-riders" leads to more random directed interactions among the population, decreasing the social bonds that are built upon the mutual interactions. This suppresses the epidemic spread and lowers the cumulative infection number. However, it also leads to a smaller cumulative reward compared to both full cooperation and scenarios with an increasing number of egocentric free-riders.

In summary, cooperation among fully cooperative nodes leads to a higher cumulative total reward and fewer infections compared to scenarios with egocentric or ignorant free-riders. An increasing number of "free-riders" (either egocentric or ignorant) leads to more infections, while an increasing number of egocentric "free-riders" further increases the cumulative infection numbers even more.

\subsection{Disaster Resilience}
\label{resilience}
We further investigate people's resilience level to a more severe epidemic spread when they are respectively driven by the single preference mutation style (see section~\ref{single2}) and the mixed preference mutation (see section~\ref{mix2}) styles. We consider the transmissibility $\zeta^s_t$ ranging from $0.05$, $0.10$, $0.15$ to $0.20$. 
We also extend recovery time $\tau^r$ from $5$ to $10$, $15$ and $20$ days (see initial parameter setups in Table~\ref{tabParam}). 

\subsubsection{Single Preference Mutation Style}
\label{single2}
\begin{figure*}[htp]
    \centering
    \subfloat[\footnotesize $\tau^r=5$ and $\zeta^s_t=0.05$]{%
\includegraphics[width=0.45\columnwidth]{ResiliencePlots/Compare_EachInfectionRT5Inf0.05.png}%
    }%
    \subfloat[\footnotesize $\tau^r=5$ and $\zeta^s_t=0.10$]{%
    \includegraphics[width=0.45\columnwidth]{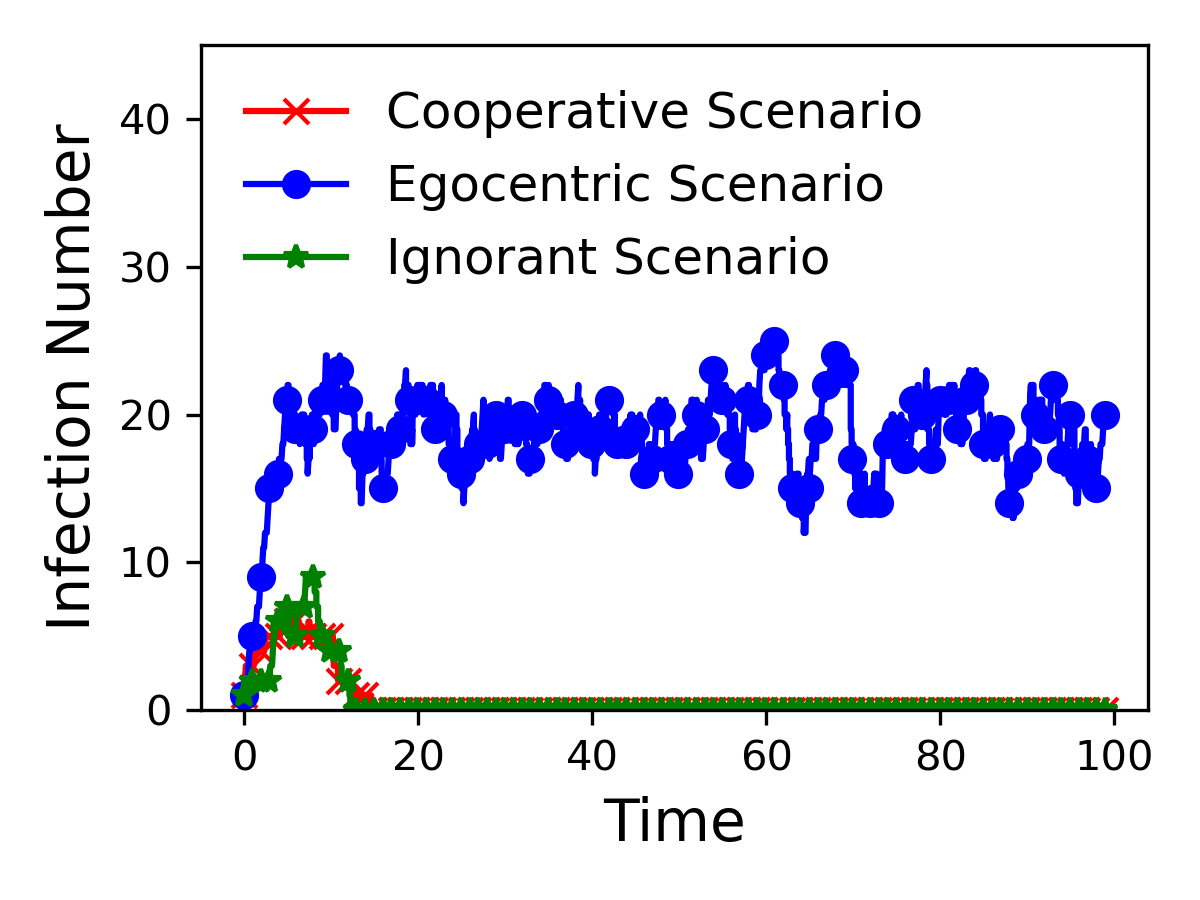}}
    \subfloat[\footnotesize $\tau^r=5$ and $\zeta^s_t=0.15$]{%
        \includegraphics[width=0.45\columnwidth]{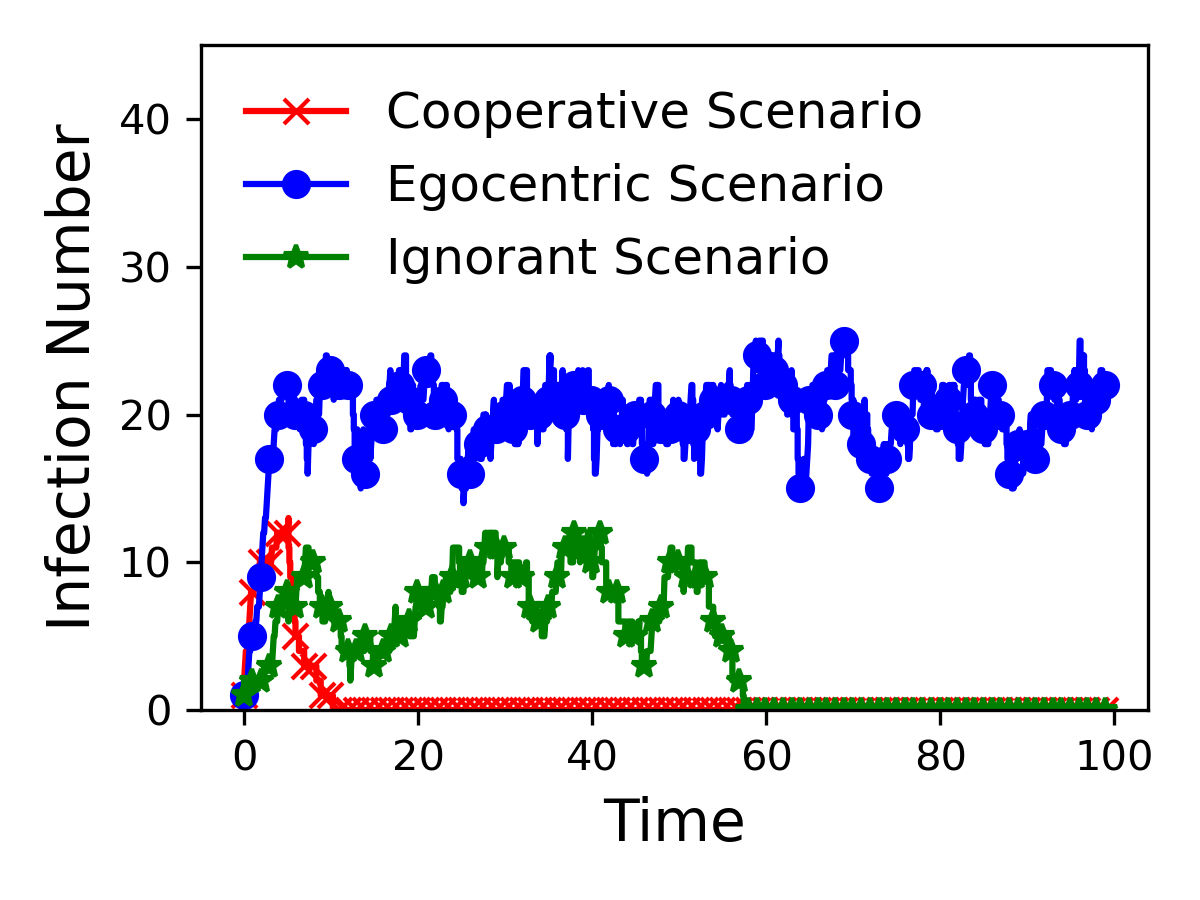}}
    \subfloat[\footnotesize $\tau^r=5$ and $\zeta^s_t=0.20$]{%
        \includegraphics[width=0.45\columnwidth]{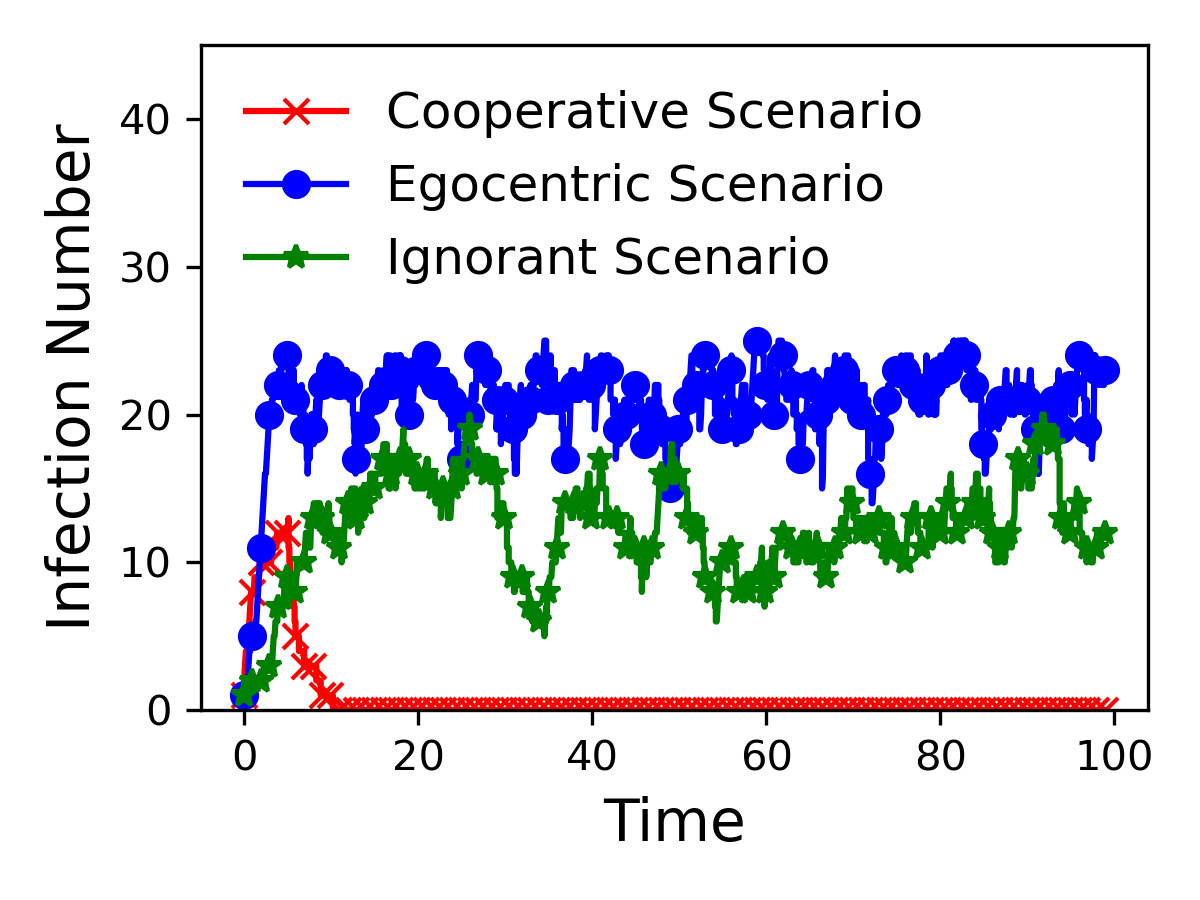}}
\\
        \subfloat[\footnotesize $\tau^r=10$ and $\zeta^s_t=0.05$]{%
\includegraphics[width=0.45\columnwidth]{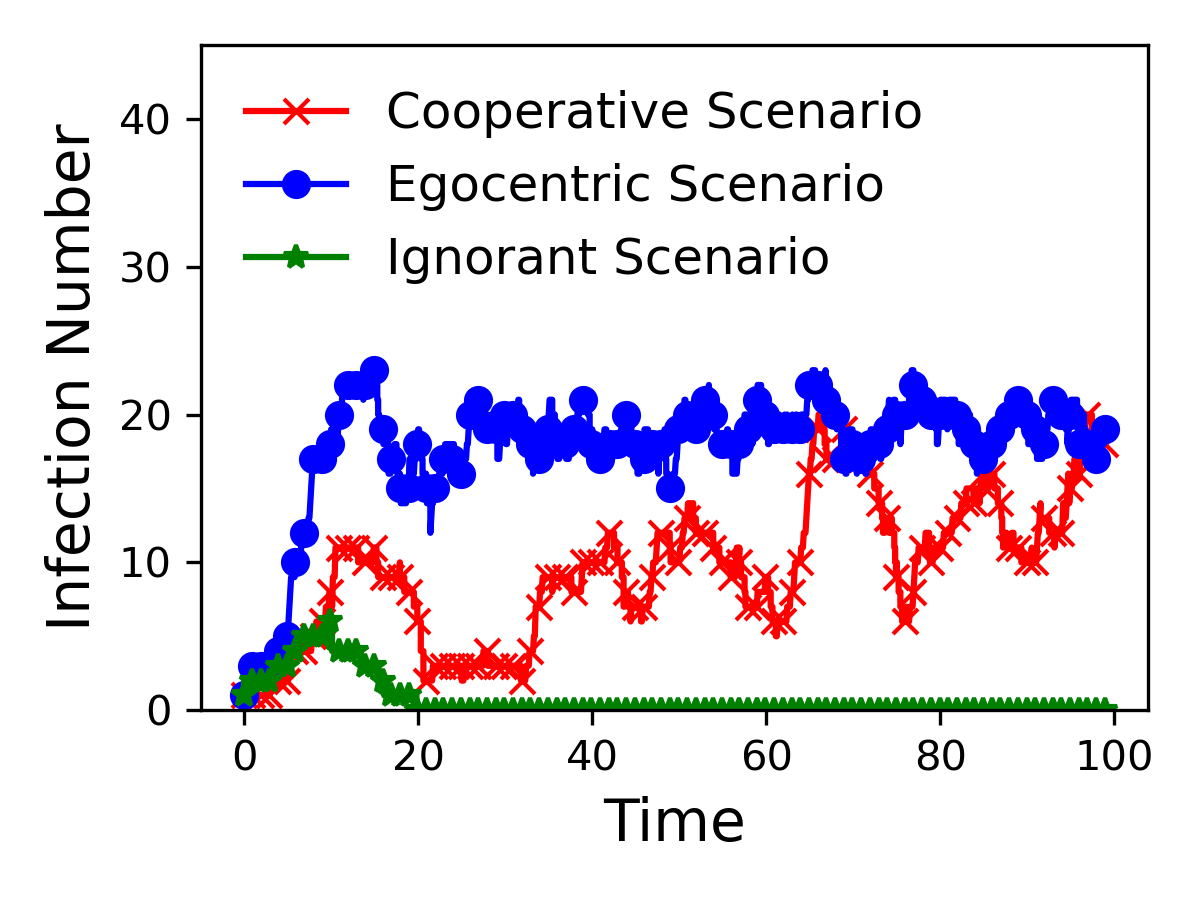}%
    }%
    \subfloat[\footnotesize $\tau^r=10$ and $\zeta^s_t=0.10$]{%
        \includegraphics[width=0.45\columnwidth]{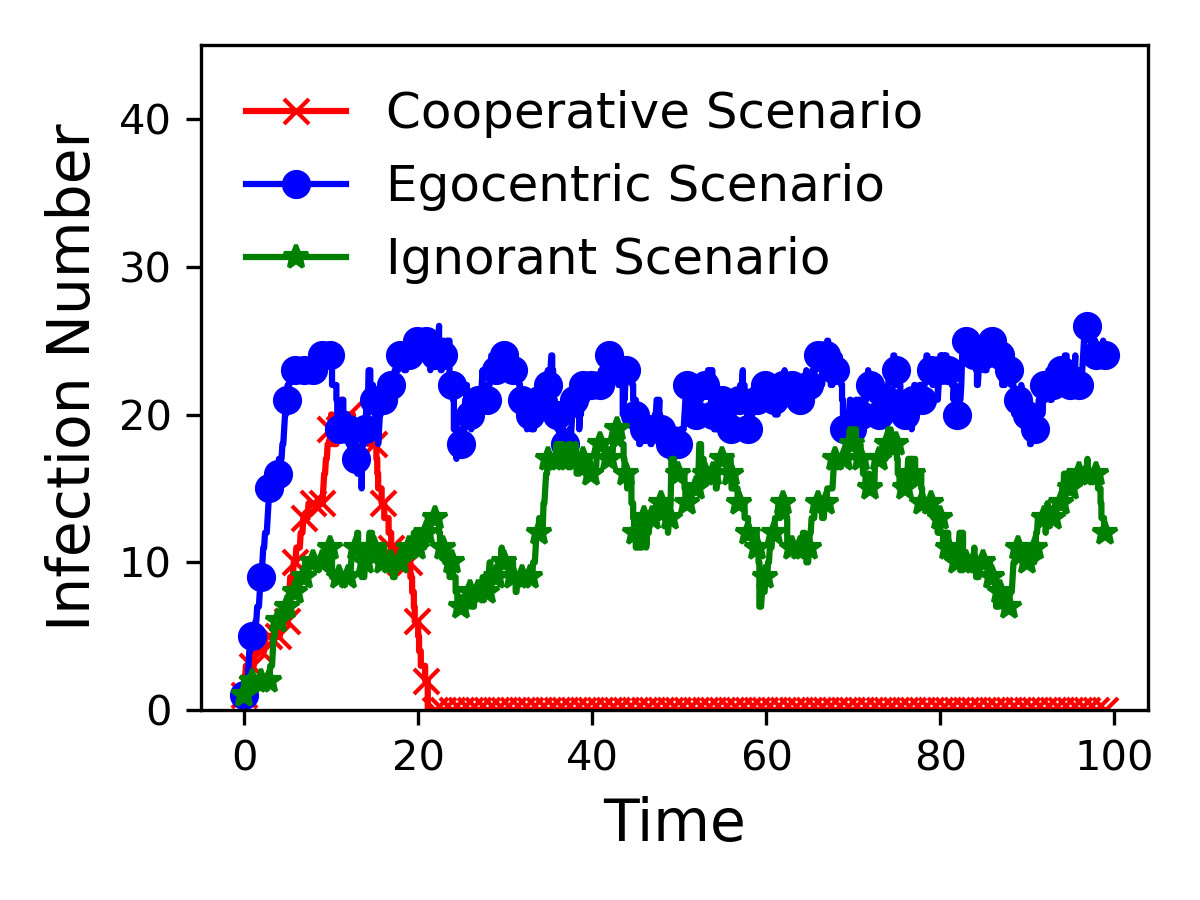}}
    \subfloat[\footnotesize $\tau^r=10$ and $\zeta^s_t=0.15$]{%
        \includegraphics[width=0.45\columnwidth]{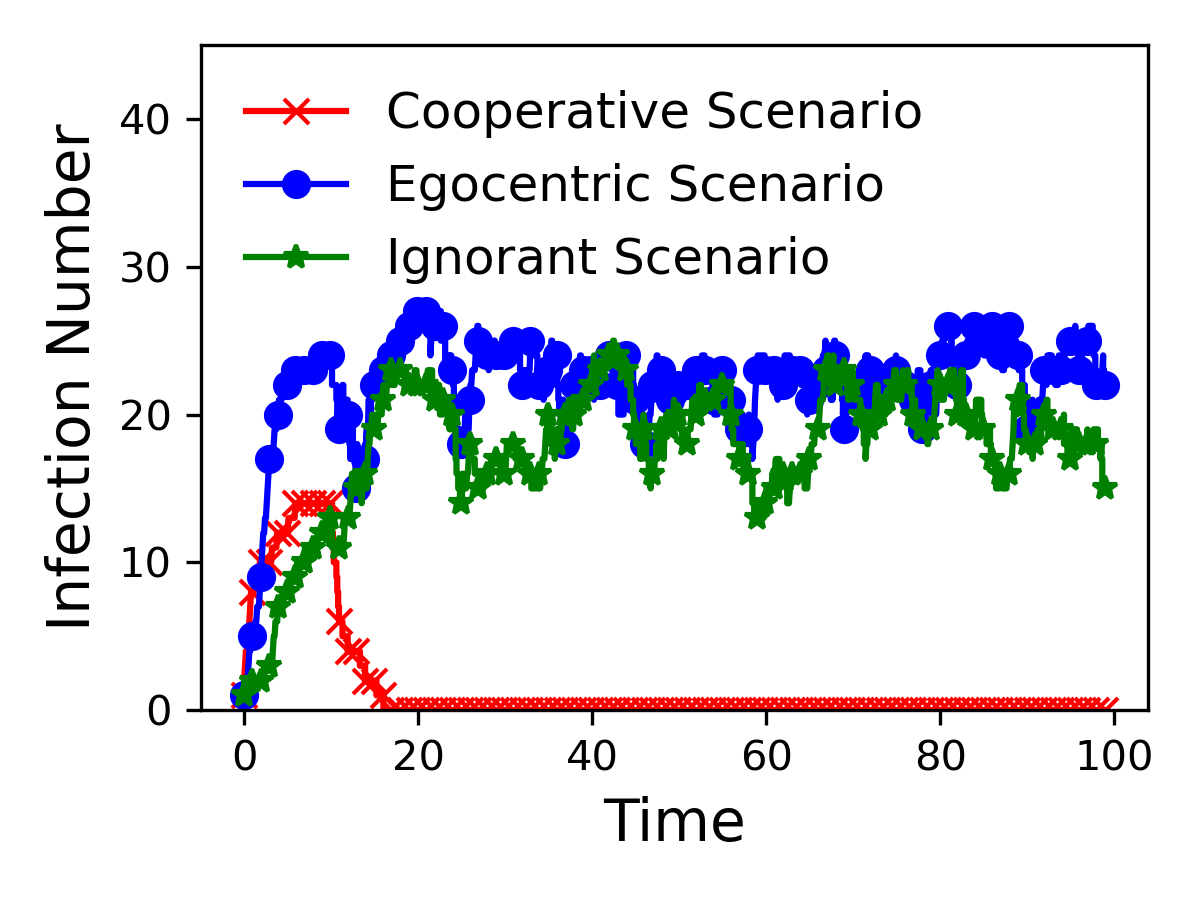}}
    \subfloat[\footnotesize $\tau^r=10$ and $\zeta^s_t=0.20$]{%
        \includegraphics[width=0.45\columnwidth]{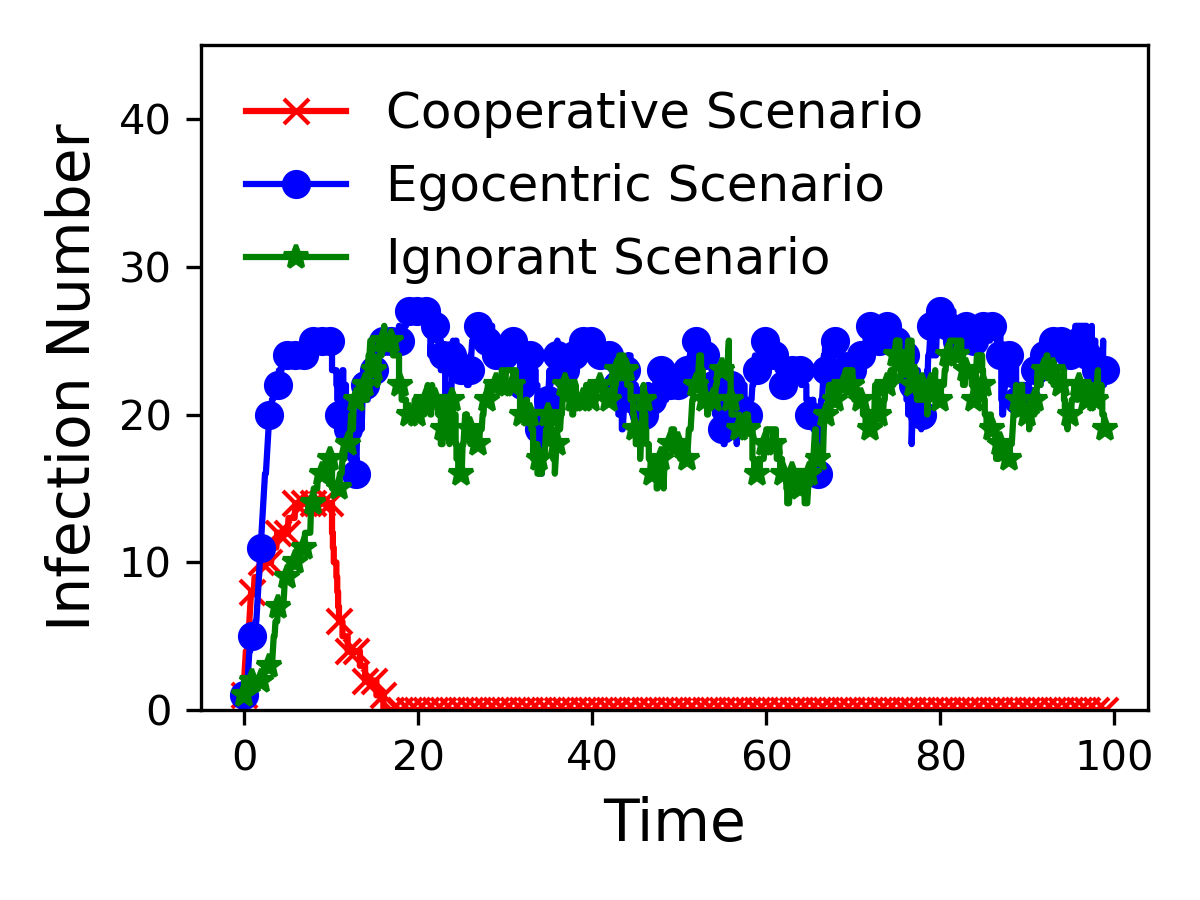}}
    \\
        \subfloat[\footnotesize $\tau^r=15$ and $\zeta^s_t=0.05$]{%
\includegraphics[width=0.45\columnwidth]{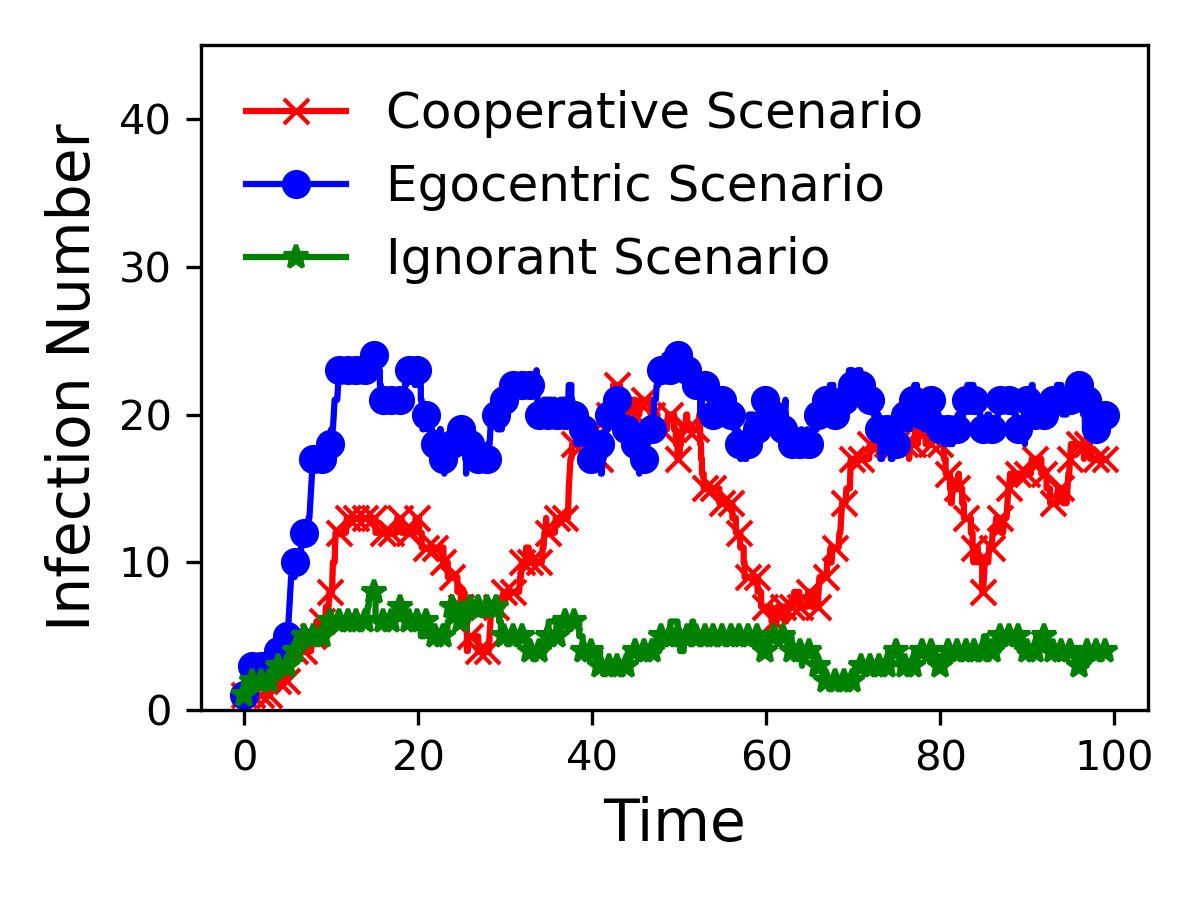}%
    }%
    \subfloat[\footnotesize $\tau^r=15$ and $\zeta^s_t=0.10$]{%
    \includegraphics[width=0.45\columnwidth]{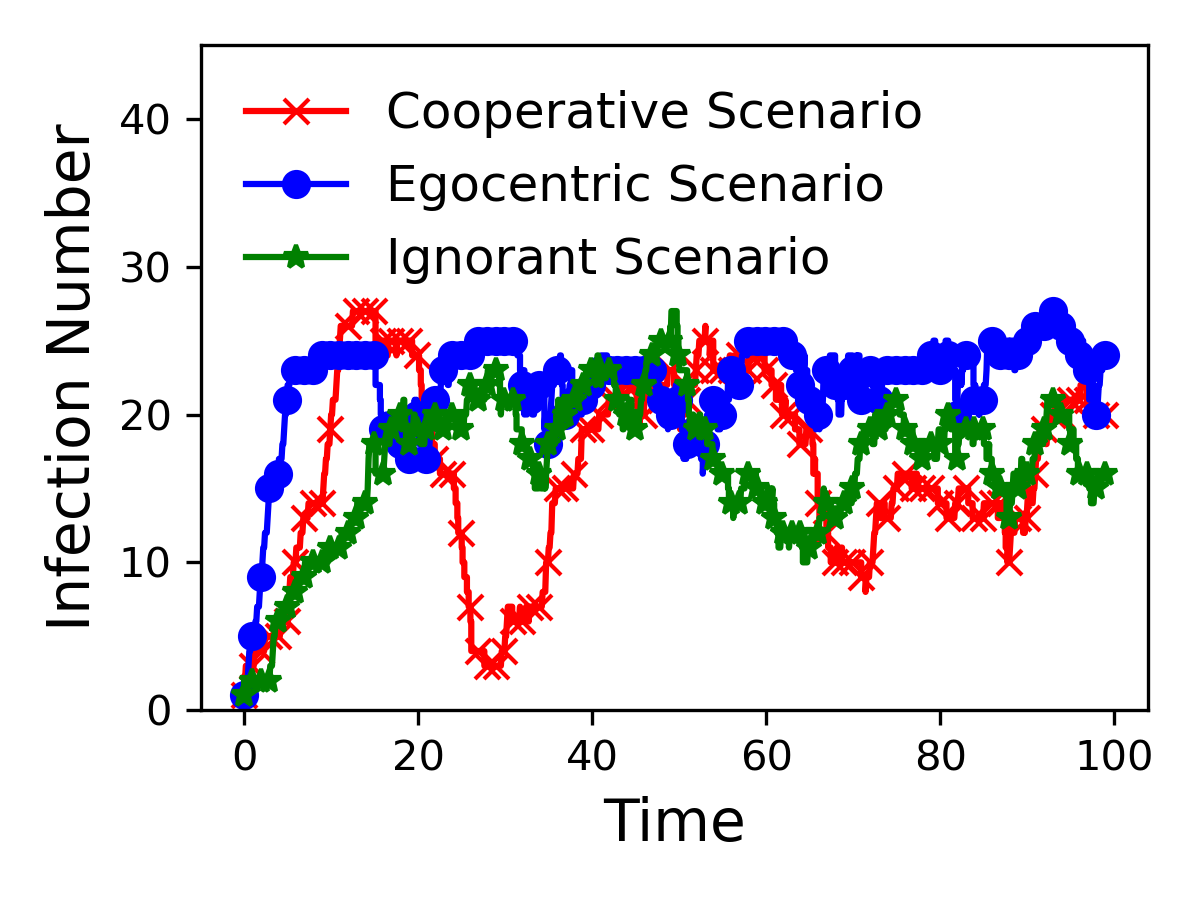}}
    \subfloat[\footnotesize $\tau^r=15$ and $\zeta^s_t=0.15$]{%
        \includegraphics[width=0.45\columnwidth]{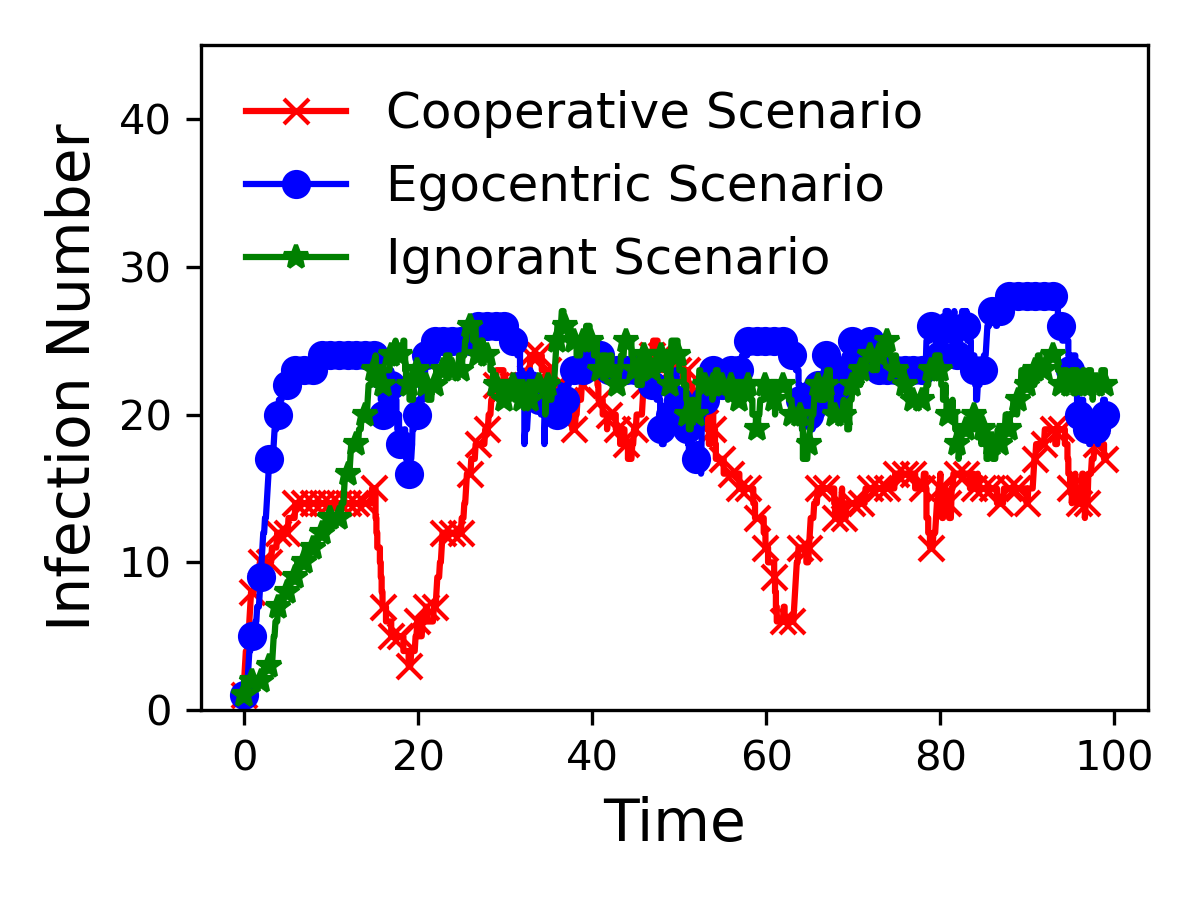}}
    \subfloat[\footnotesize $\tau^r=15$ and $\zeta^s_t=0.20$]{%
        \includegraphics[width=0.45\columnwidth]{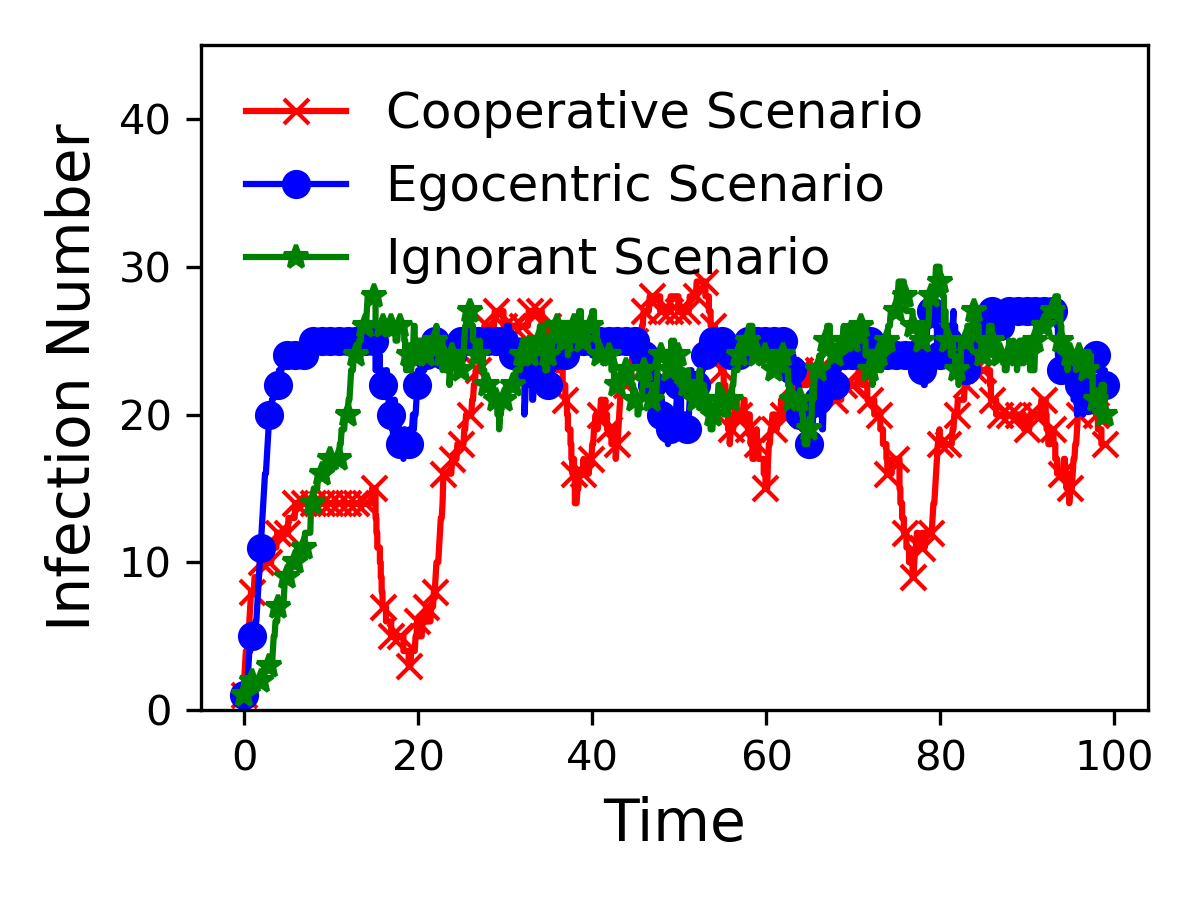}}
    \\
        \subfloat[\footnotesize $\tau^r=20$ and $\zeta^s_t=0.05$]{%
\includegraphics[width=0.45\columnwidth]{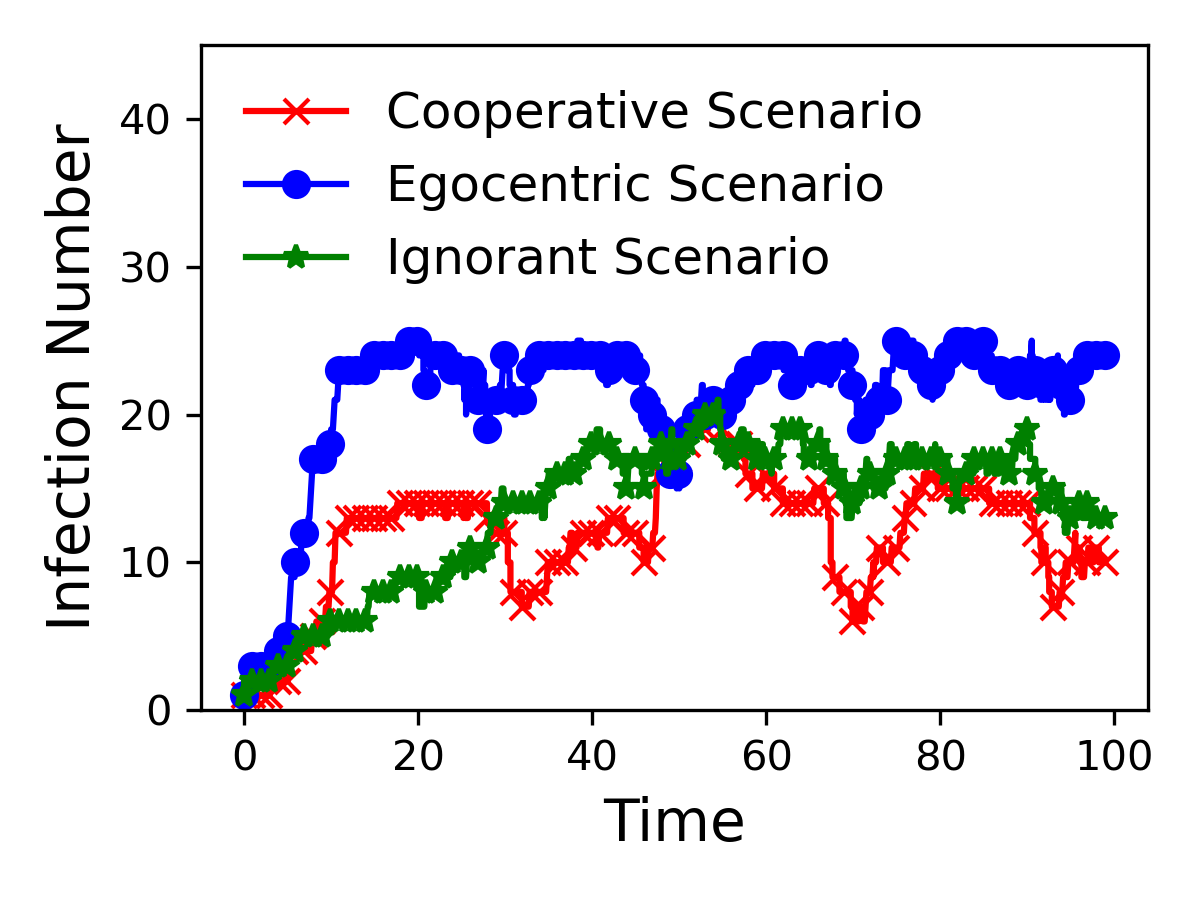}%
    }%
    \subfloat[\footnotesize $\tau^r=20$ and $\zeta^s_t=0.10$]{%
        \includegraphics[width=0.45\columnwidth]{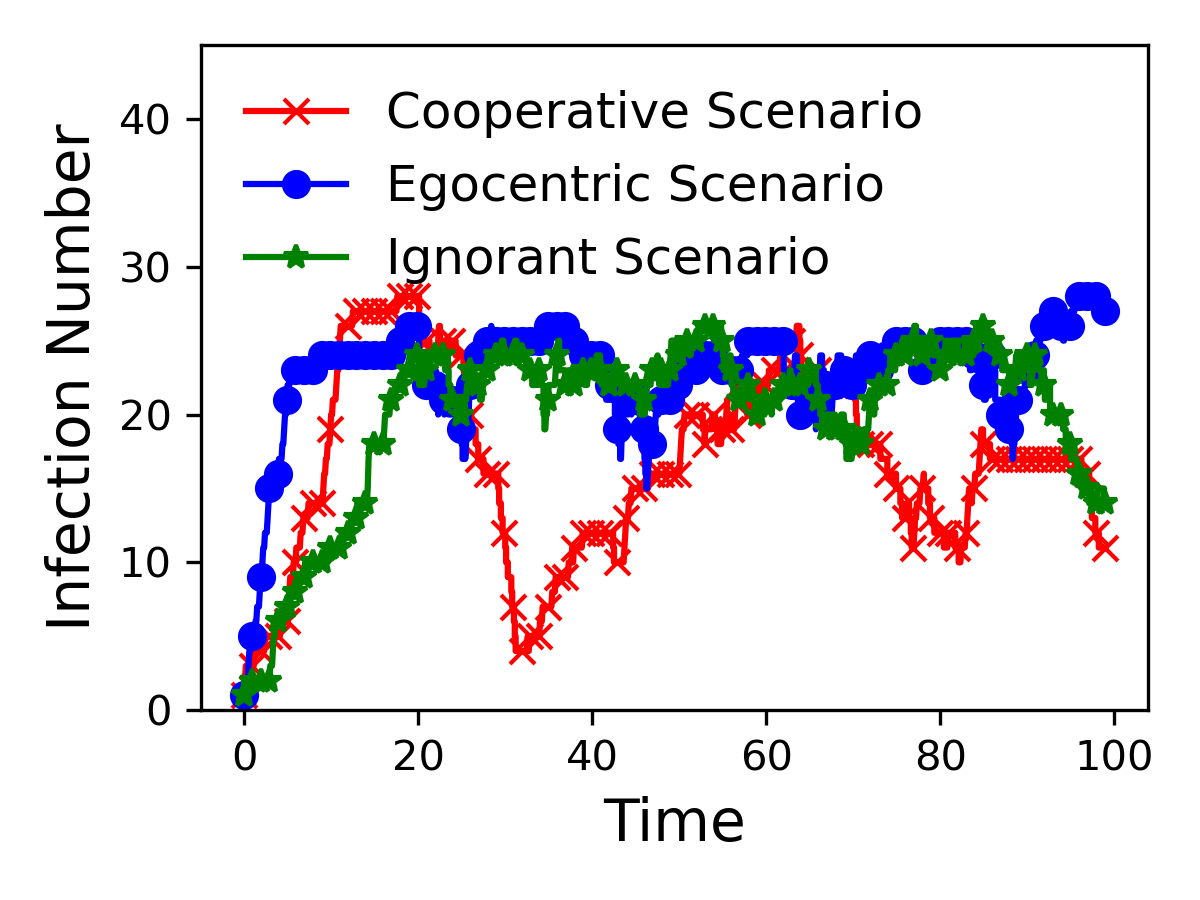}}
    \subfloat[\footnotesize $\tau^r=20$ and $\zeta^s_t=0.15$]{%
        \includegraphics[width=0.45\columnwidth]{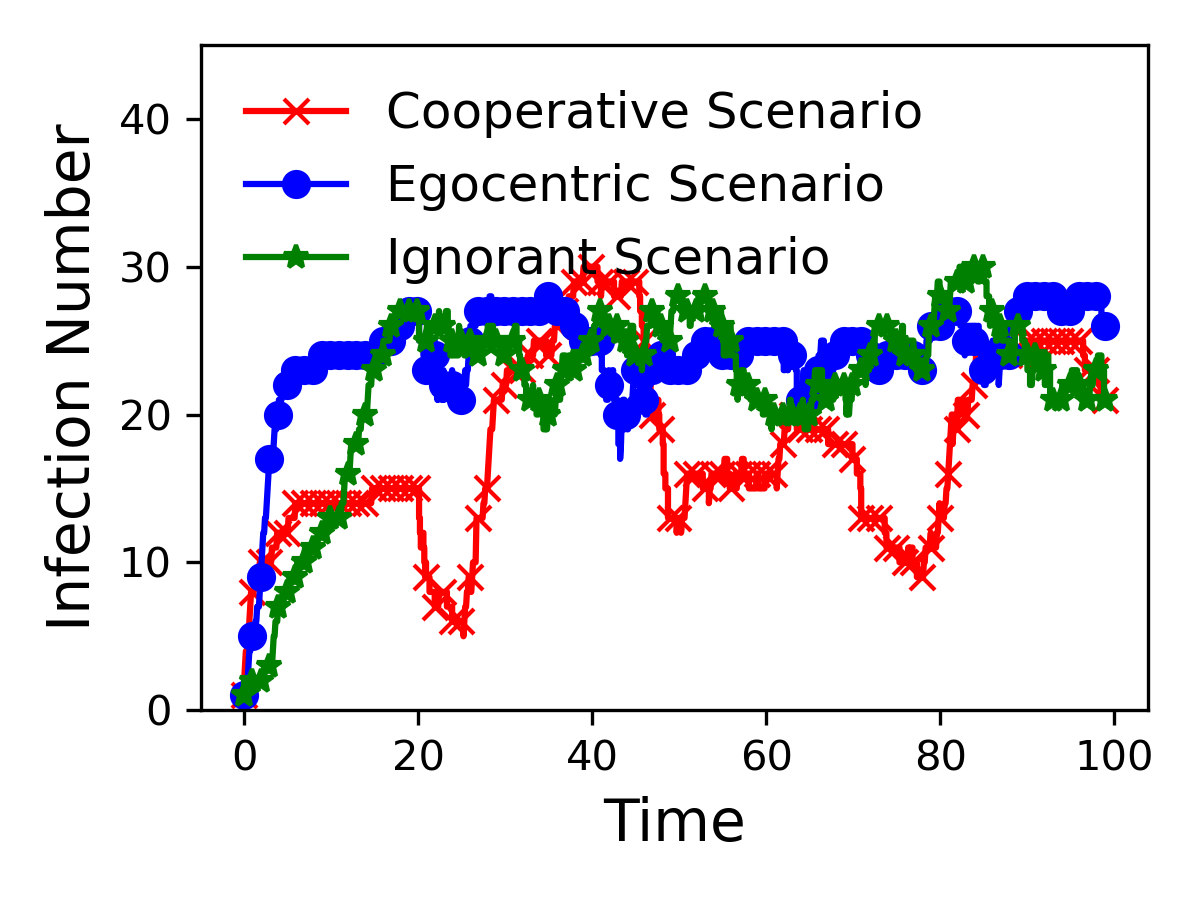}}
    \subfloat[\footnotesize $\tau^r=20$ and $\zeta^s_t=0.20$]{%
        \includegraphics[width=0.45\columnwidth]{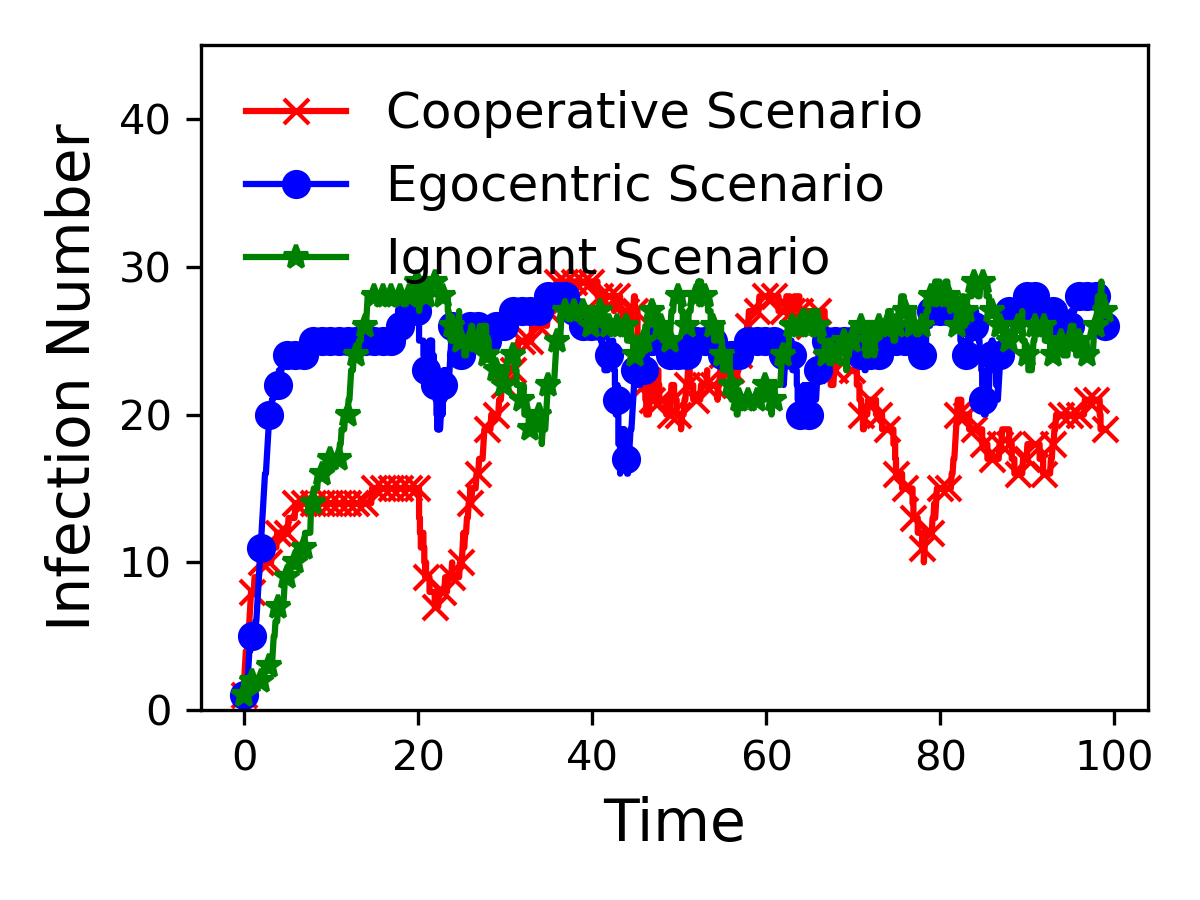}}
	\caption{The cumulative infection occurrences in cooperative, ignorant and egocentric scenarios given various setups of infection rate and recovery time.}
 \label{infectionnumber0}
\end{figure*}

Fig.~\ref{infectionnumber0} shows the infection occurrences over time given various infection rates and recovery time values in the network simulations driven by a single preference mutation style. Overall, the infection number increases given a higher infection rate and a longer recovery time. The network simulations, which are purely driven by the egocentric free-riders, generally have the highest infection number throughout the interaction period. This suggests that networks purely comprised of egocentric "free-riders" are less resilient to the epidemic spread than the networks driven by the other node types. In addition, the infection number generally decreases to zero in the fully cooperative networks given a recovery time of $5$ and $10$, which in contrast, significantly increases with large fluctuations given the recovery time at $15$ and $20$. This suggests that cooperation remains resilient against epidemic spread and prevent widespread contagion given low infection rates and recovery time values. Given a high infection rate and recovery time value, the cooperative nodes soon get infected early and engage in mutual interactions, impeding recovery and increasing infections. In addition, changes in recovery time have a more pronounced impact on infection numbers than changes in infection rates. This suggests that recovery time must be closely monitored when making epidemic mitigation strategies.

\begin{figure*}[htp]
    \centering
    \subfloat[\footnotesize $\tau^r=5$ and $\zeta^s_t=0.05$]{%
\includegraphics[width=0.45\columnwidth]{ResiliencePlots/Compare_EachRewardRT5Inf0.05.png}%
    }%
    \subfloat[\footnotesize $\tau^r=5$ and $\zeta^s_t=0.10$]{%
    \includegraphics[width=0.45\columnwidth]{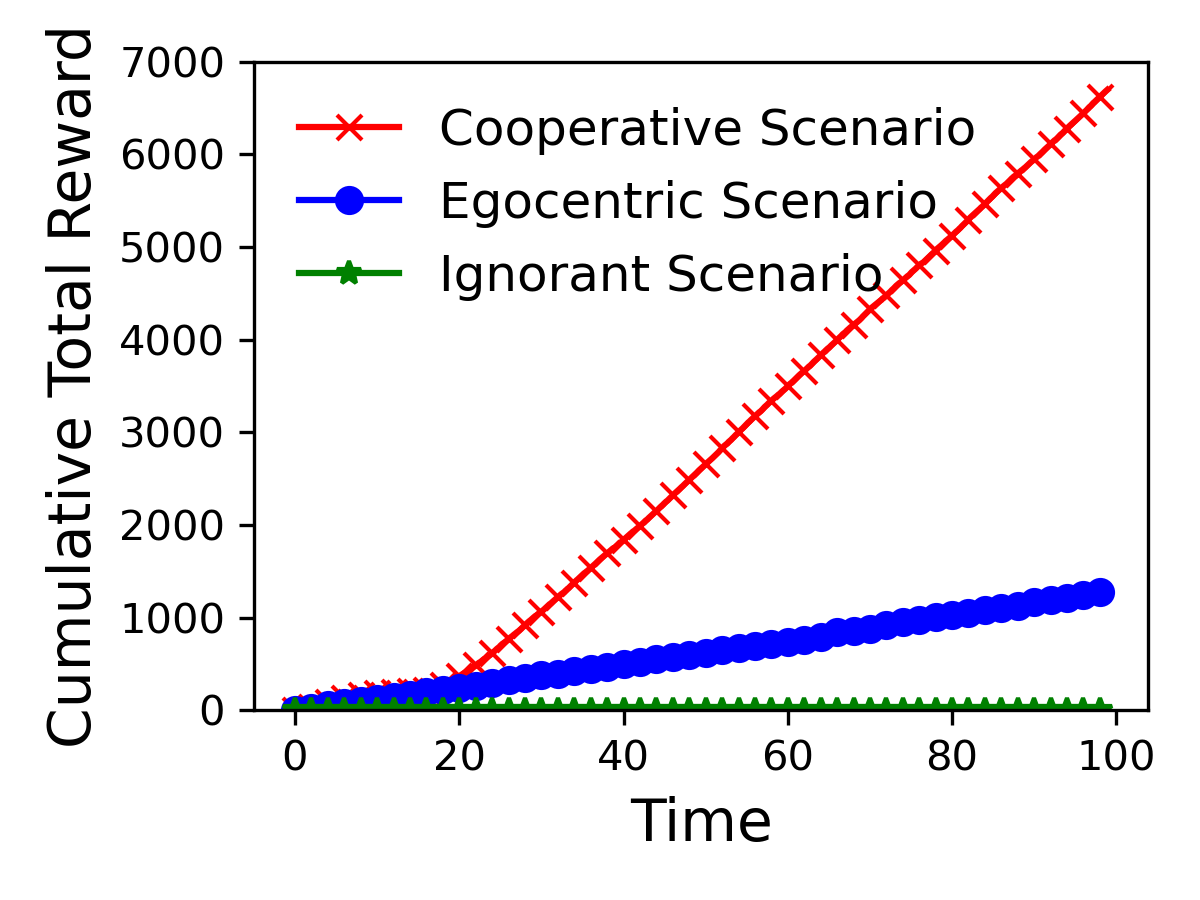}}
    \subfloat[\footnotesize $\tau^r=5$ and $\zeta^s_t=0.15$]{%
        \includegraphics[width=0.45\columnwidth]{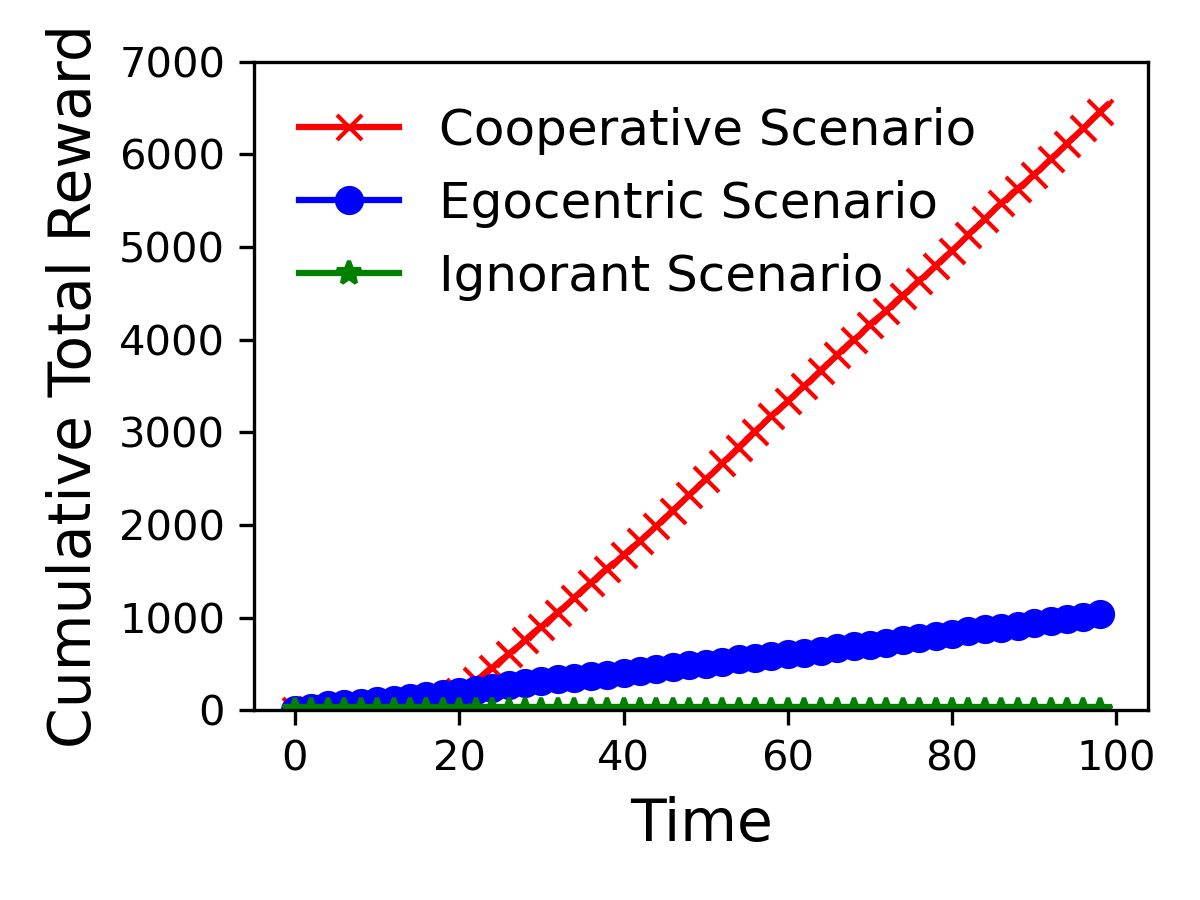}}
    \subfloat[\footnotesize $\tau^r=5$ and $\zeta^s_t=0.20$]{%
        \includegraphics[width=0.45\columnwidth]{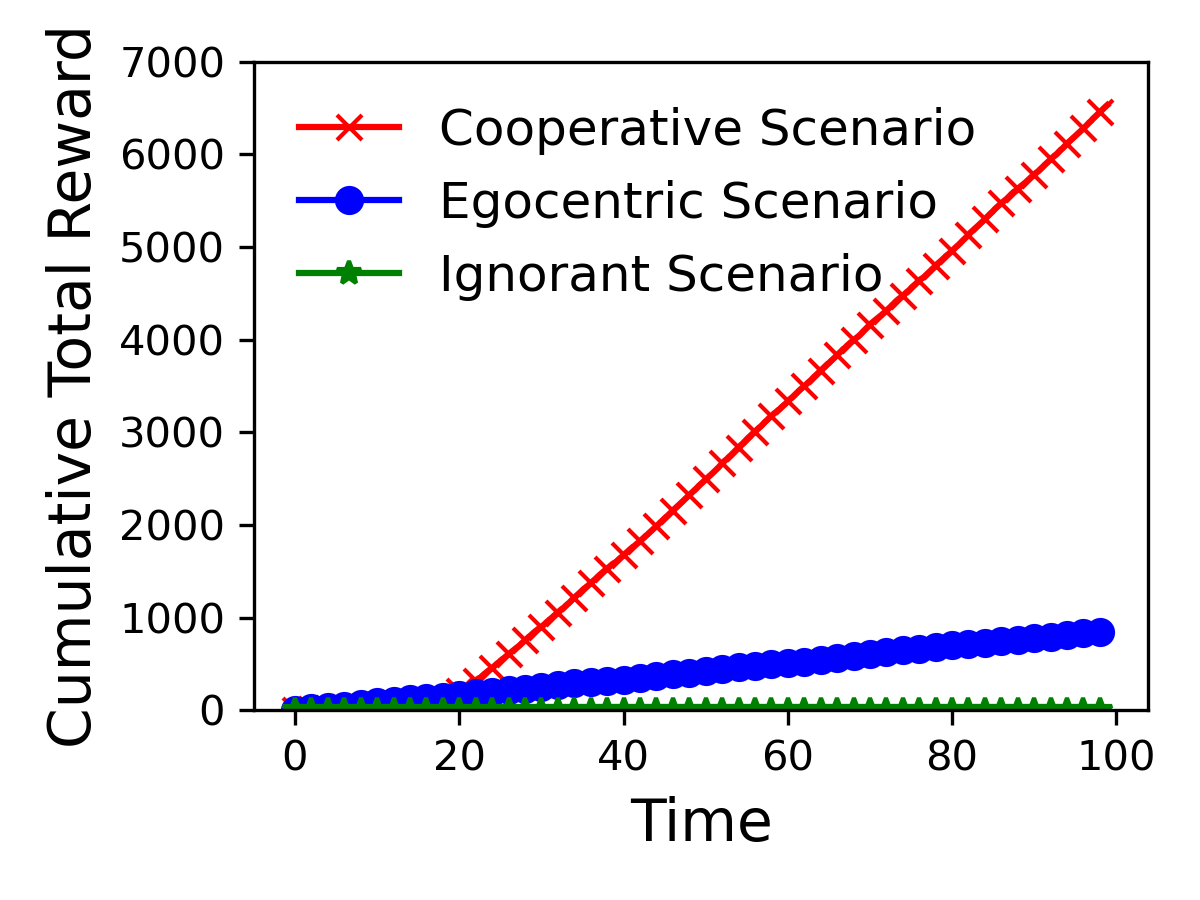}}
    \\
        \subfloat[\footnotesize $\tau^r=10$ and $\zeta^s_t=0.05$]{%
\includegraphics[width=0.45\columnwidth]{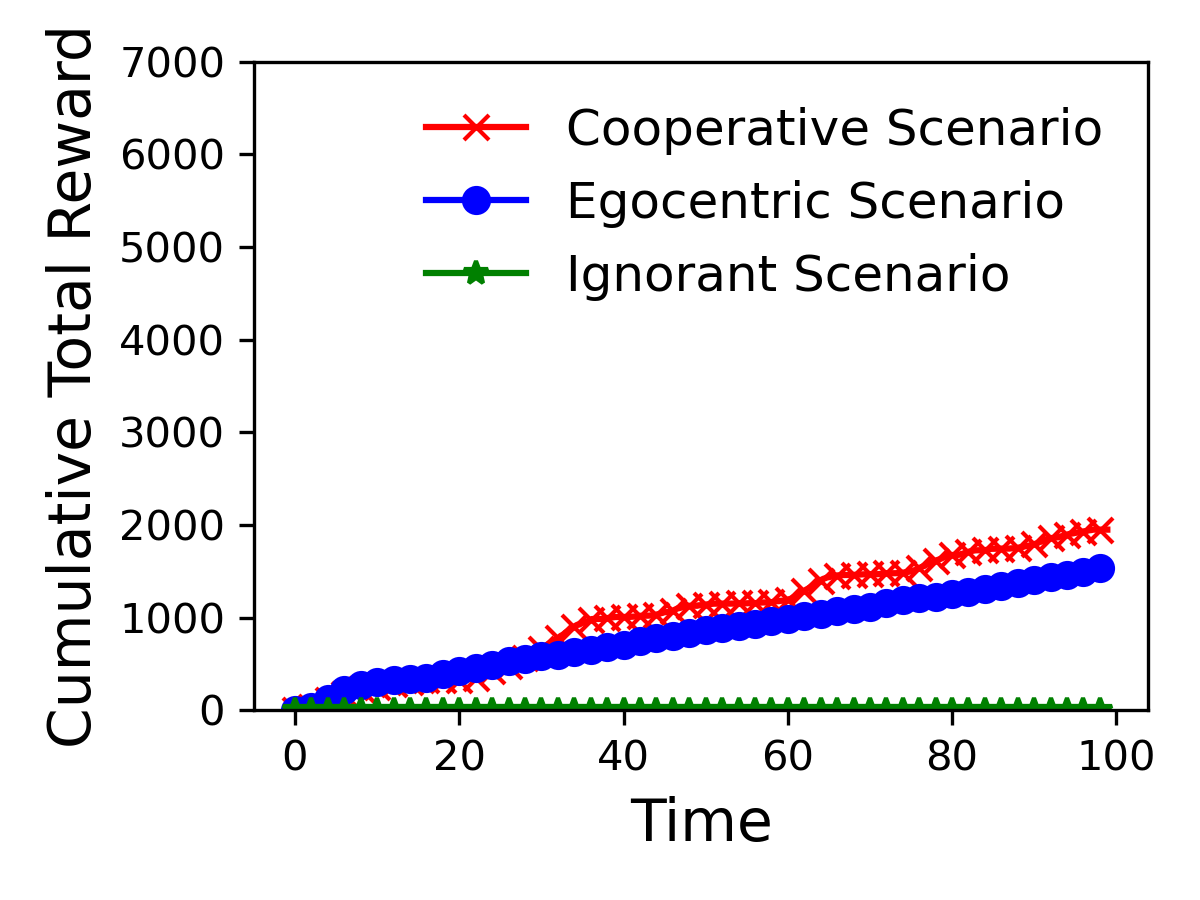}%
    }%
    \subfloat[\footnotesize $\tau^r=10$ and $\zeta^s_t=0.10$]{%
        \includegraphics[width=0.45\columnwidth]{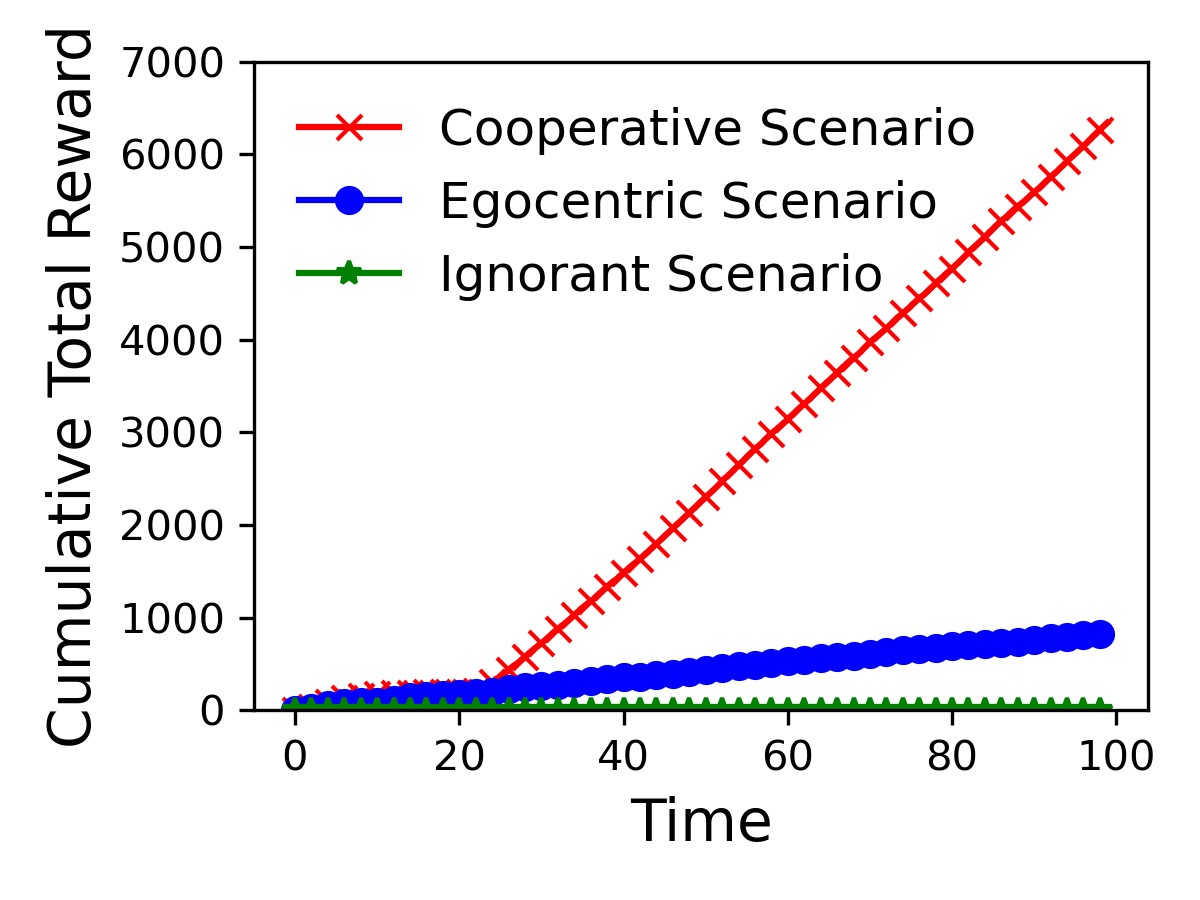}}
    \subfloat[\footnotesize $\tau^r=10$ and $\zeta^s_t=0.15$]{%
        \includegraphics[width=0.45\columnwidth]{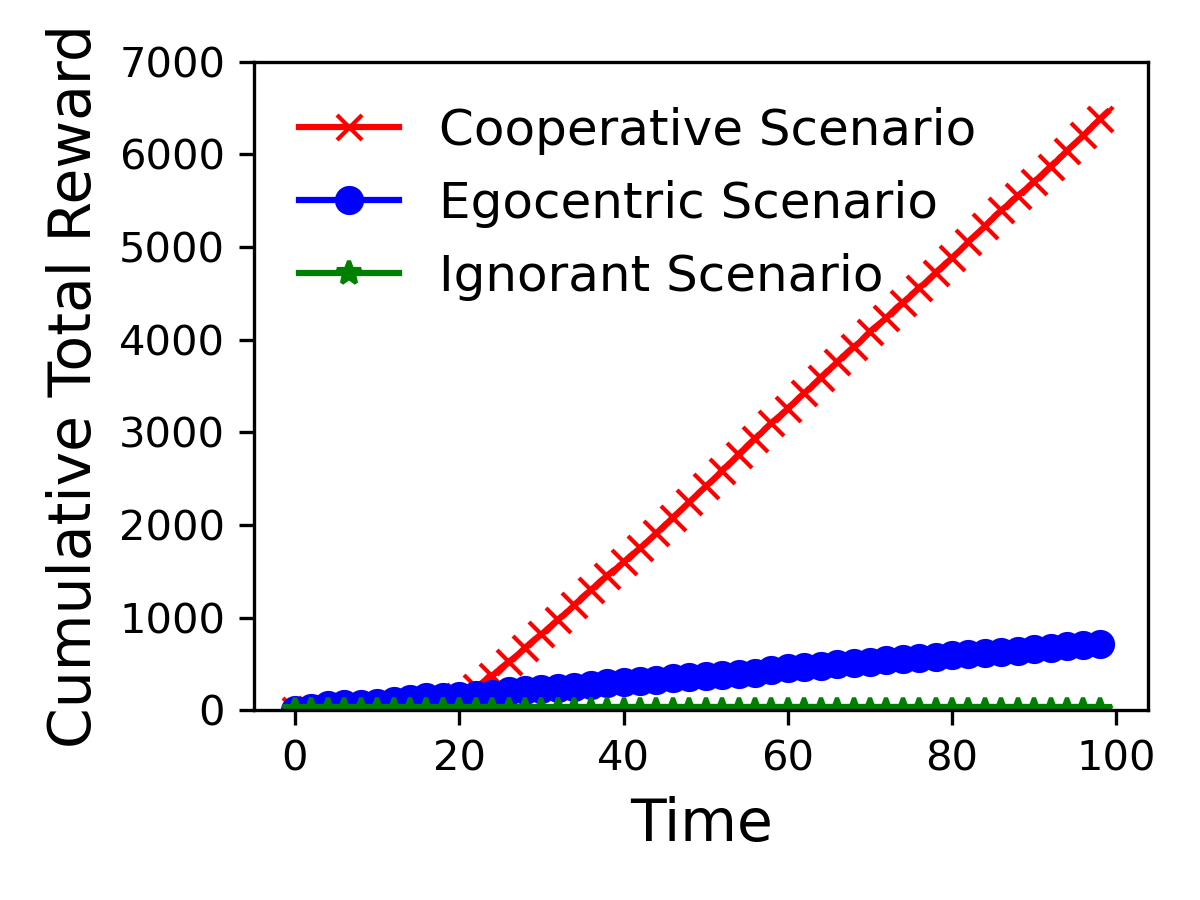}}
    \subfloat[\footnotesize $\tau^r=10$ and $\zeta^s_t=0.20$]{%
        \includegraphics[width=0.45\columnwidth]{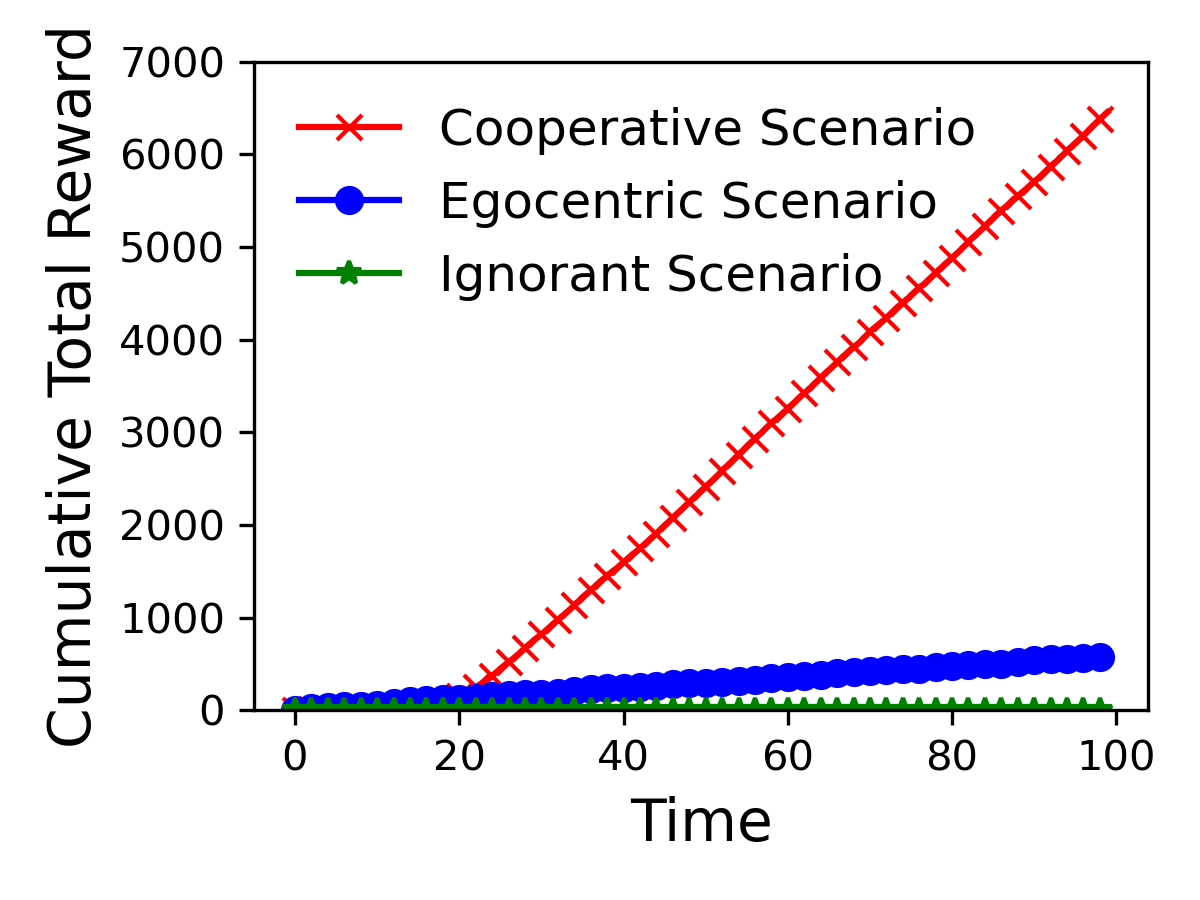}}
        \\
        \subfloat[\footnotesize $\tau^r=15$ and $\zeta^s_t=0.05$]{%
\includegraphics[width=0.45\columnwidth]{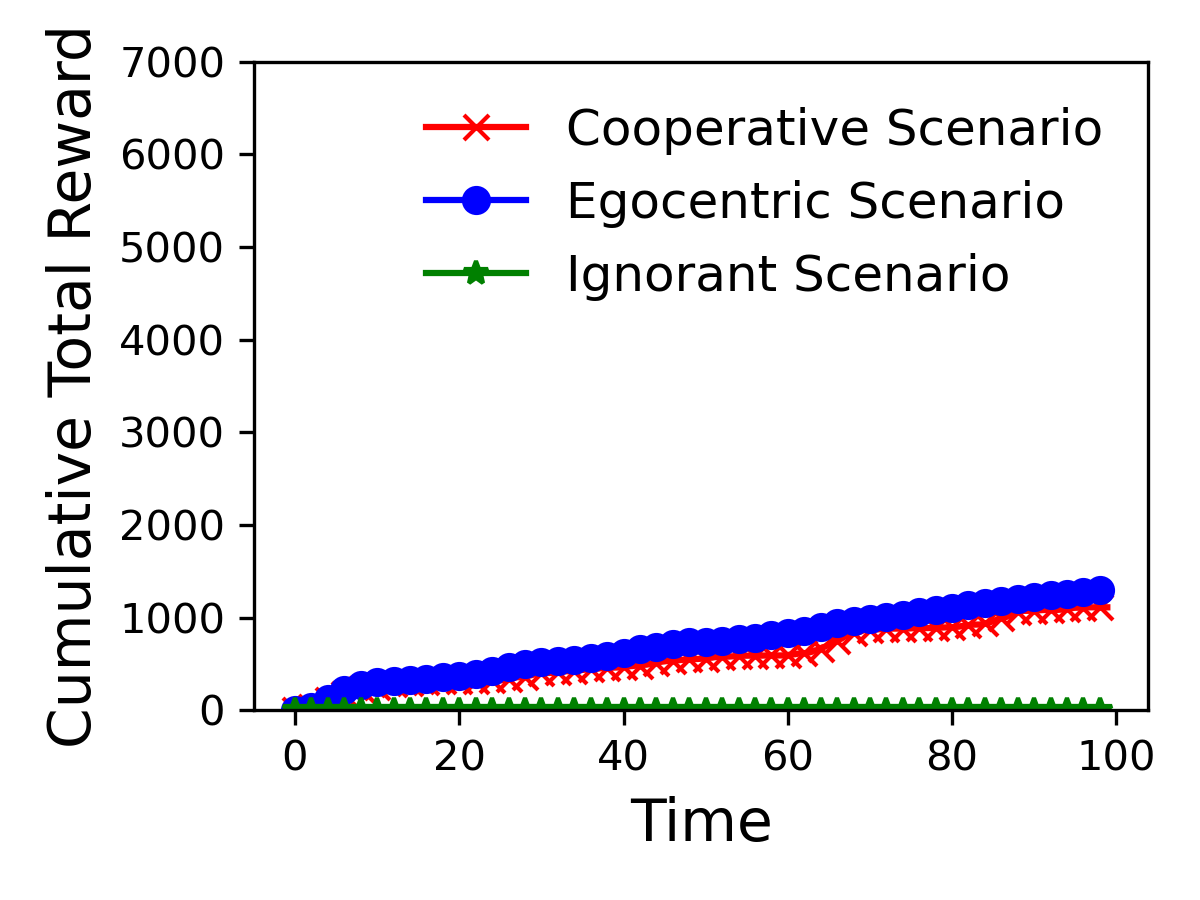}%
    }%
    \subfloat[\footnotesize $\tau^r=15$ and $\zeta^s_t=0.10$]{%
    \includegraphics[width=0.45\columnwidth]{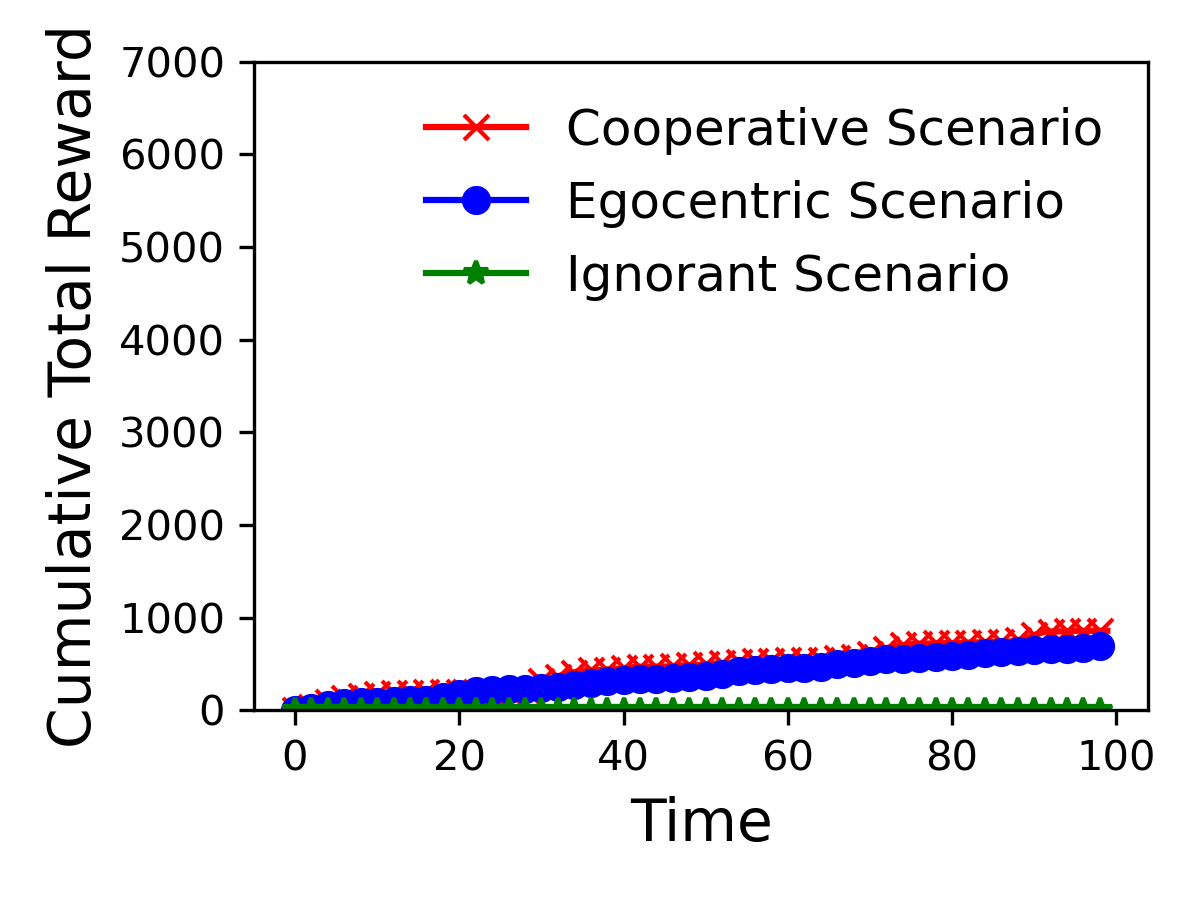}}
    \subfloat[\footnotesize $\tau^r=15$ and $\zeta^s_t=0.15$]{%
        \includegraphics[width=0.45\columnwidth]{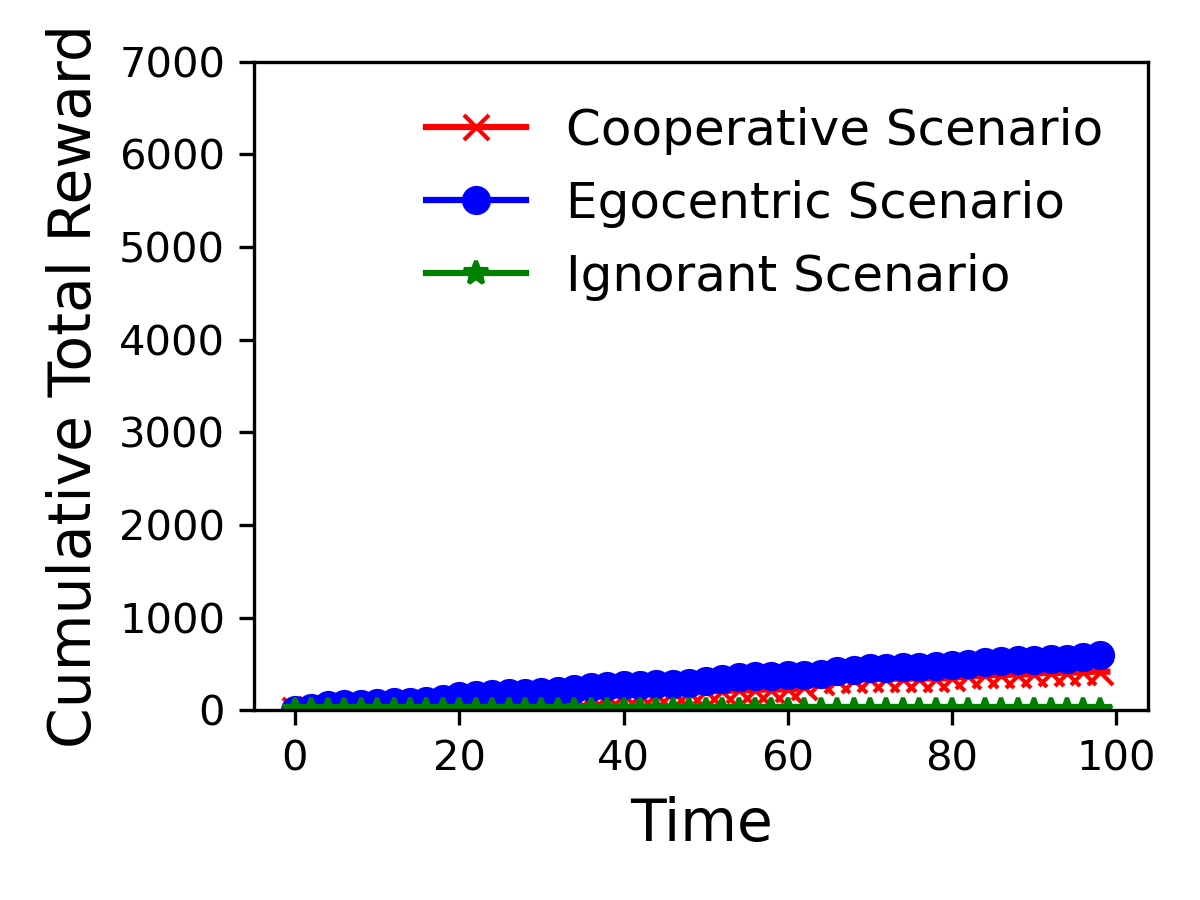}}
    \subfloat[\footnotesize $\tau^r=15$ and $\zeta^s_t=0.20$]{%
        \includegraphics[width=0.45\columnwidth]{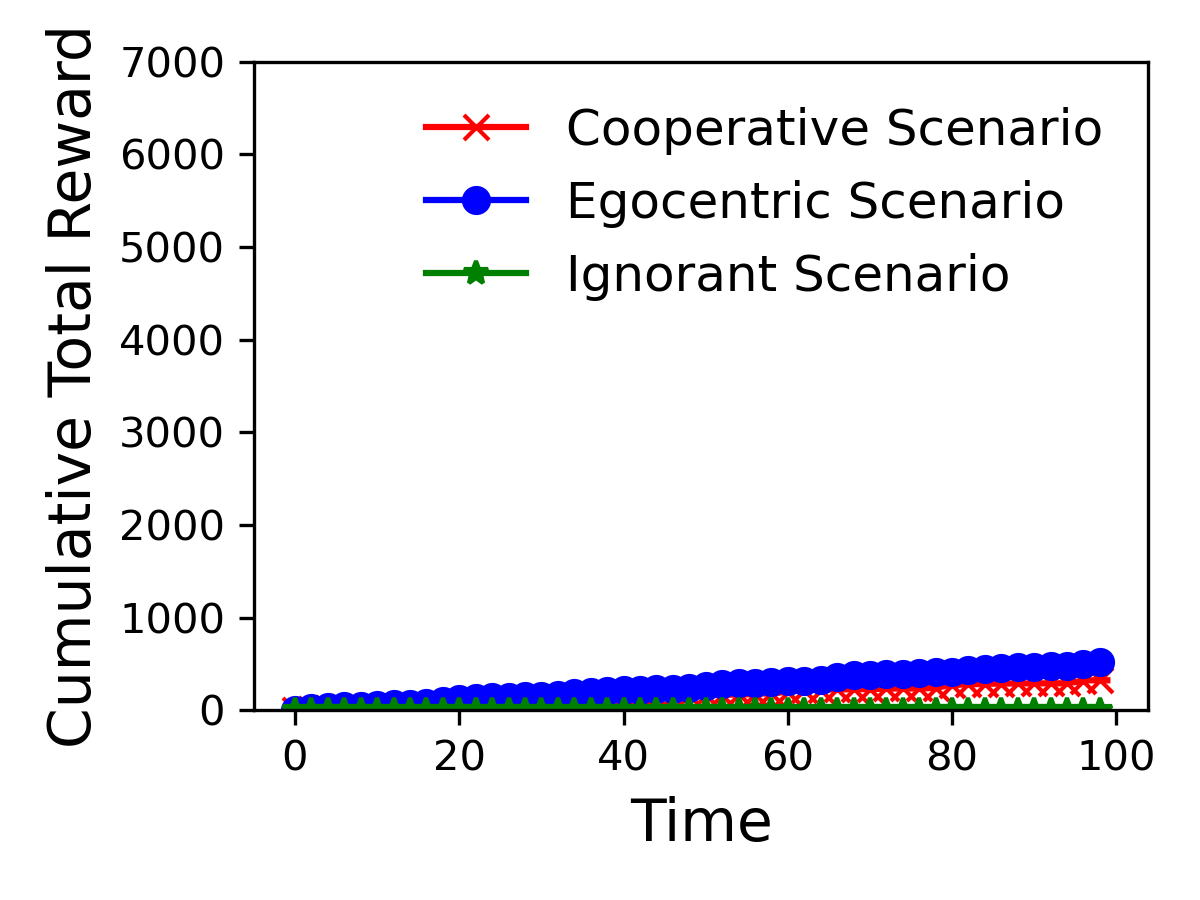}}
    \\
        \subfloat[\footnotesize $\tau^r=20$ and $\zeta^s_t=0.05$]{%
\includegraphics[width=0.45\columnwidth]{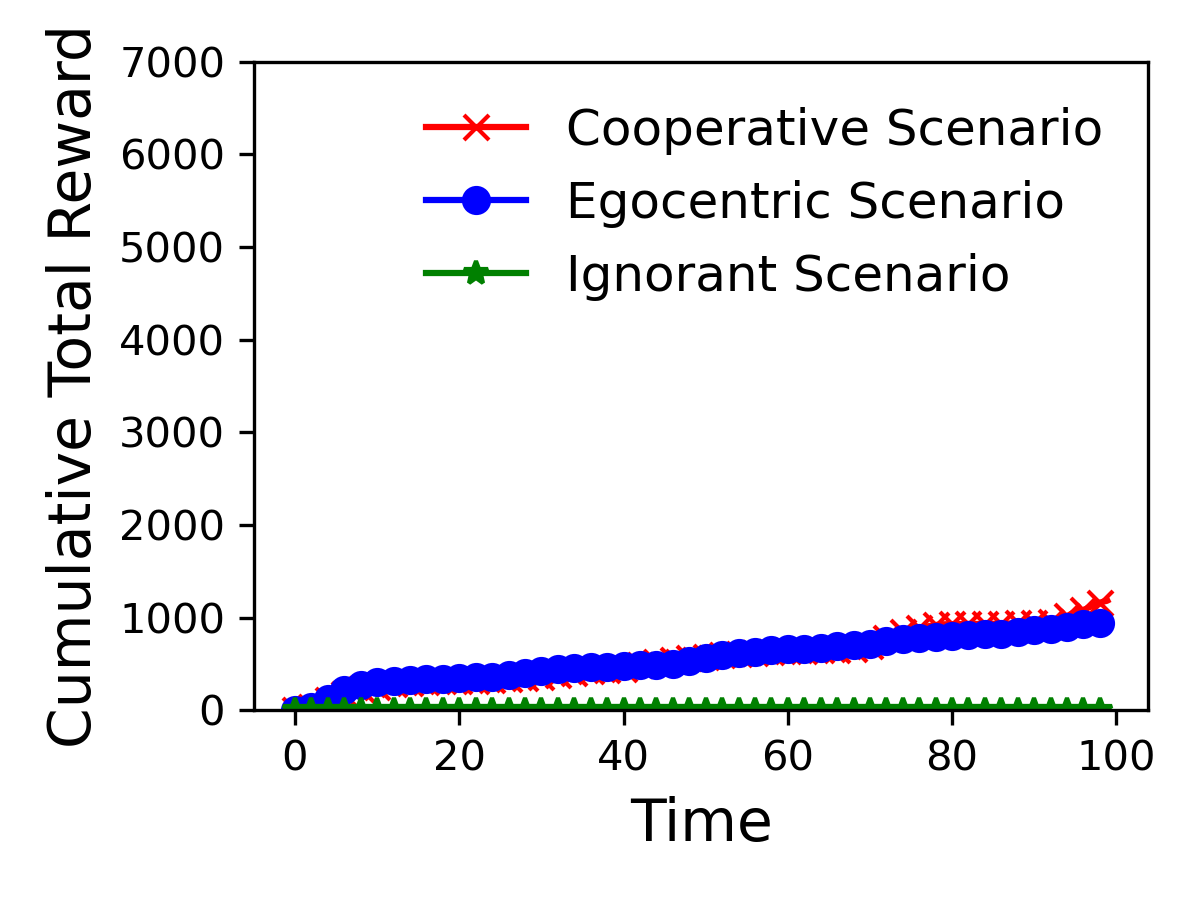}%
    }%
    \subfloat[\footnotesize $\tau^r=20$ and $\zeta^s_t=0.10$]{%
        \includegraphics[width=0.45\columnwidth]{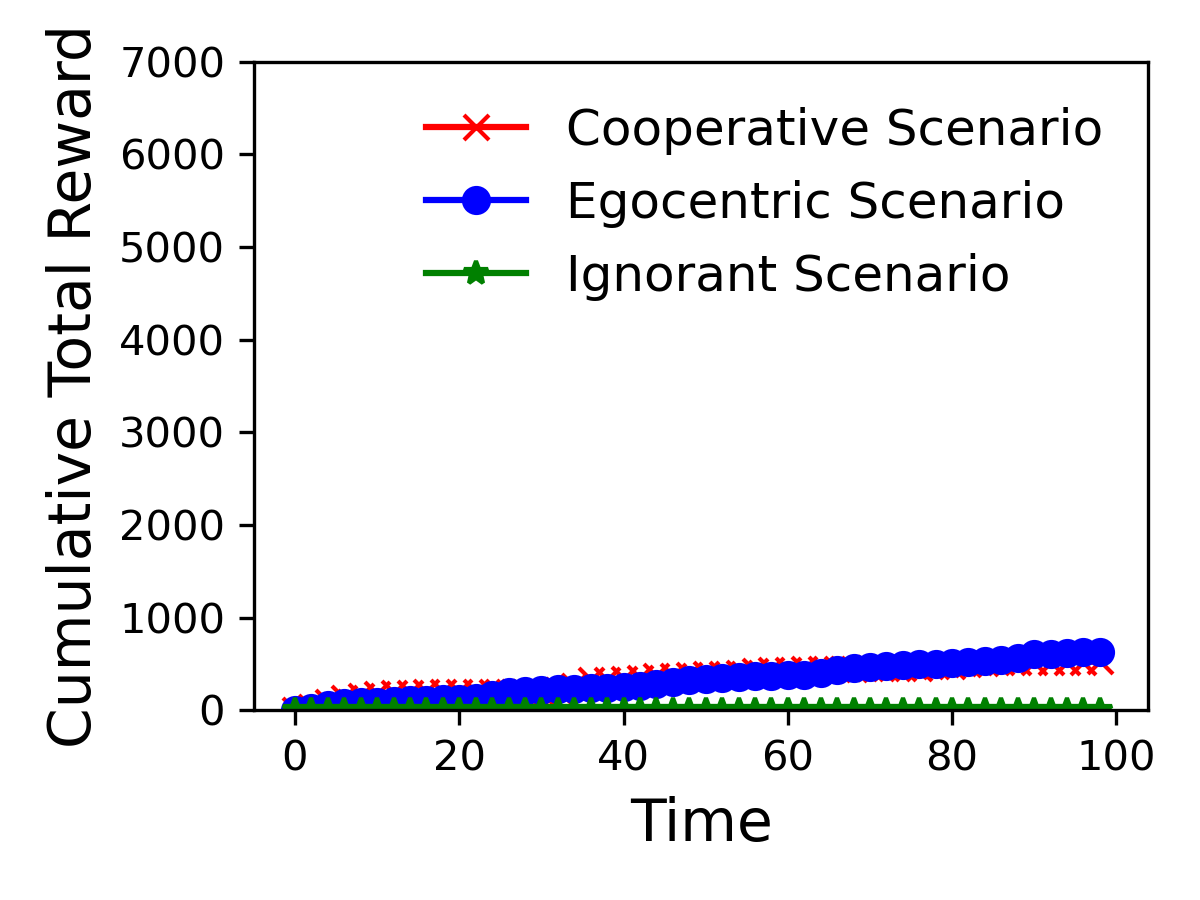}}
    \subfloat[\footnotesize $\tau^r=20$ and $\zeta^s_t=0.15$]{%
        \includegraphics[width=0.45\columnwidth]{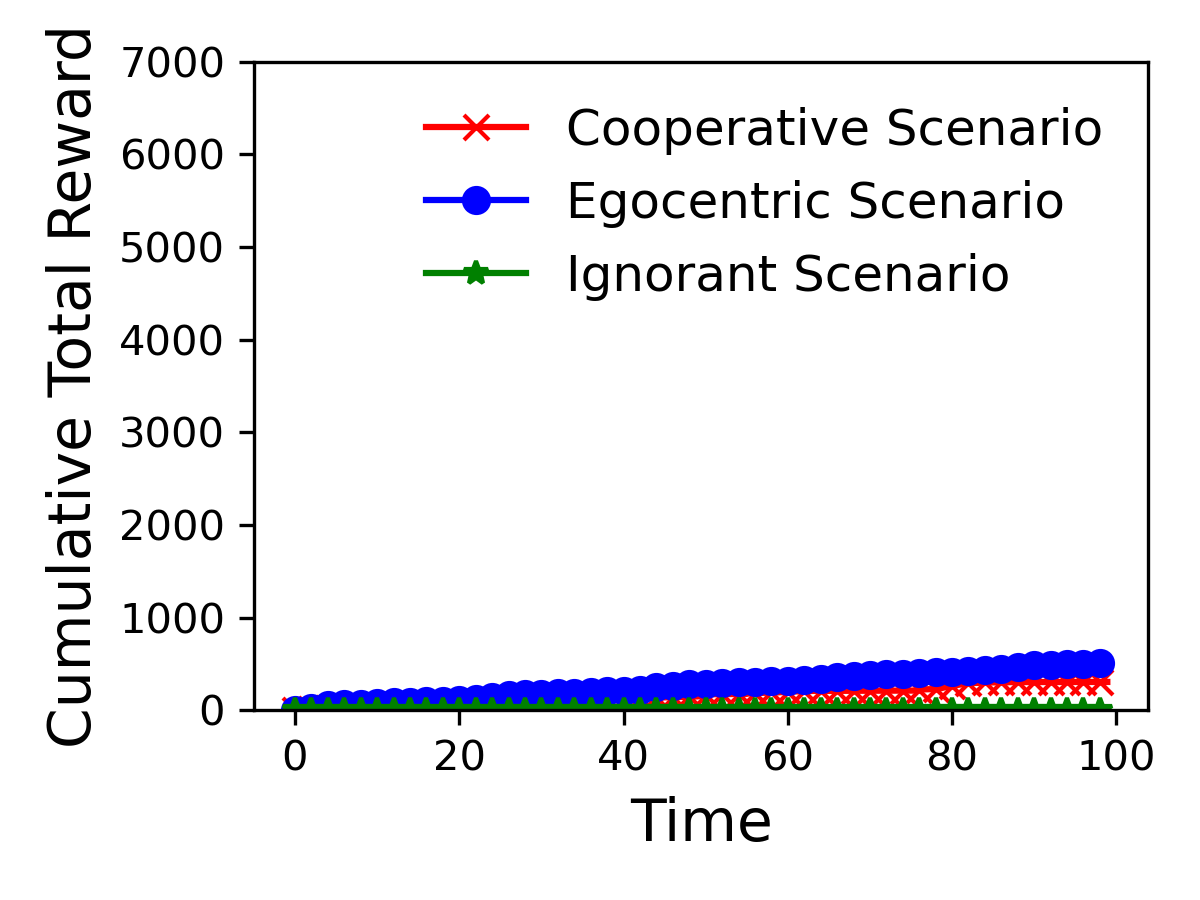}}
    \subfloat[\footnotesize $\tau^r=20$ and $\zeta^s_t=0.20$]{%
        \includegraphics[width=0.45\columnwidth]{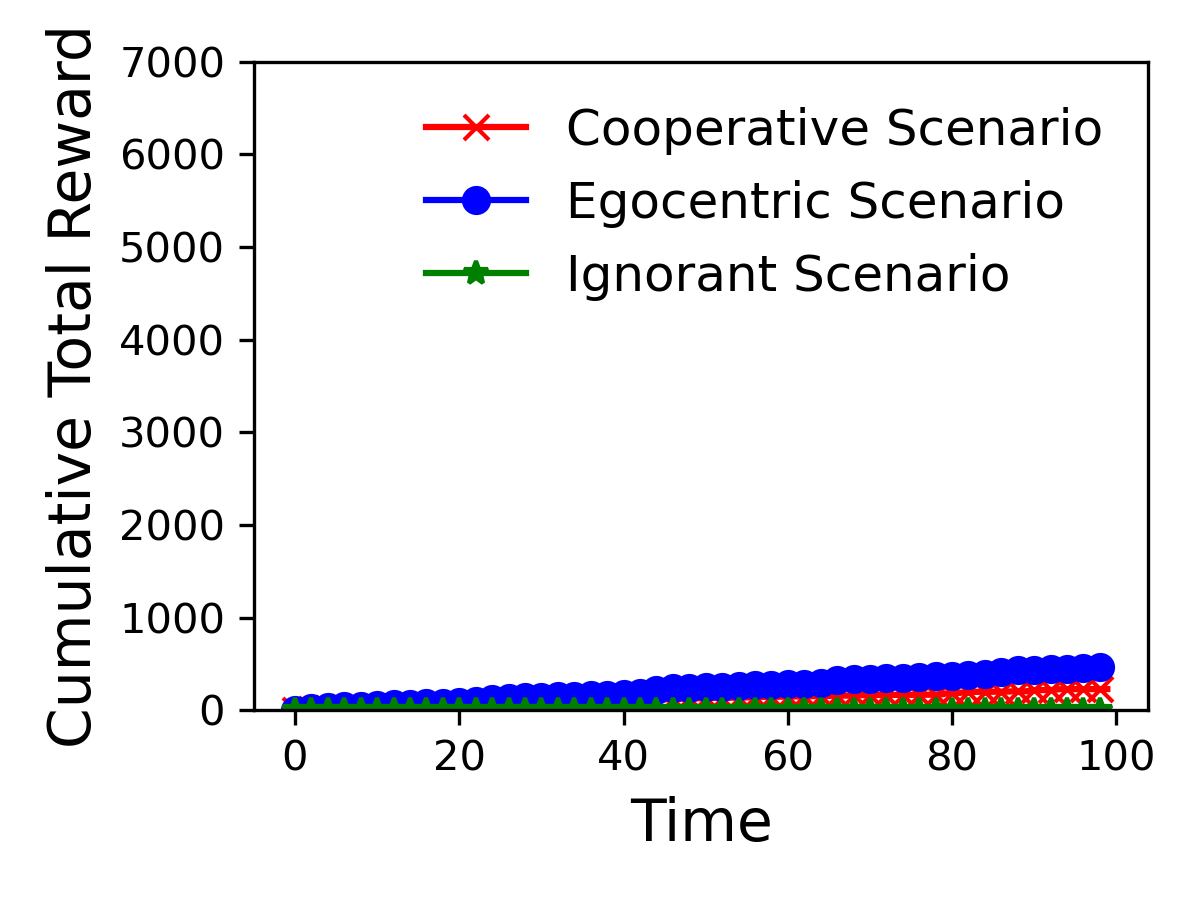}}
	\caption{The cumulative total reward in cooperative, ignorant and egocentric scenarios given various setups of infection rate and recovery time.}
 \label{CumulativeReward1}
\end{figure*}

Fig.~\ref{CumulativeReward1} shows the impact of varying infection rates and recovery times on the cumulative total reward in the network simulations that are driven by single preference mutation styles. With the increase in infection rates and recovery time, the cumulative total reward of the egocentric and ignorant nodes decreases. This can be caused by the increased infection numbers (Fig. \ref{infectionnumber0}). The cumulative total reward of the cooperative nodes in the fully cooperative network simulations decreases significantly when the recovery time increases from $10$ to higher values at $15$ and $20$. This also results from the major increase in the infection number among the fully cooperative agents. The abovementioned phenomenon indicates that cumulative total rewards are sensitive to infection dynamics and recovery times. The cumulative total reward of fully cooperative nodes can be substantially impacted by increased recovery time values.

\subsubsection{Mixed Preference Mutation Style}
\label{mix2}

\begin{figure*}[htp]
    \centering
    \subfloat[\footnotesize $\tau^r=5$ and $\zeta^s_t=0.05$]{%
\includegraphics[width=0.45\columnwidth]{ResiliencePlots/Compare_RiderInfectionRT5Inf0.05.png}%
    }%
    \subfloat[\footnotesize $\tau^r=5$ and $\zeta^s_t=0.10$]{%
    \includegraphics[width=0.45\columnwidth]{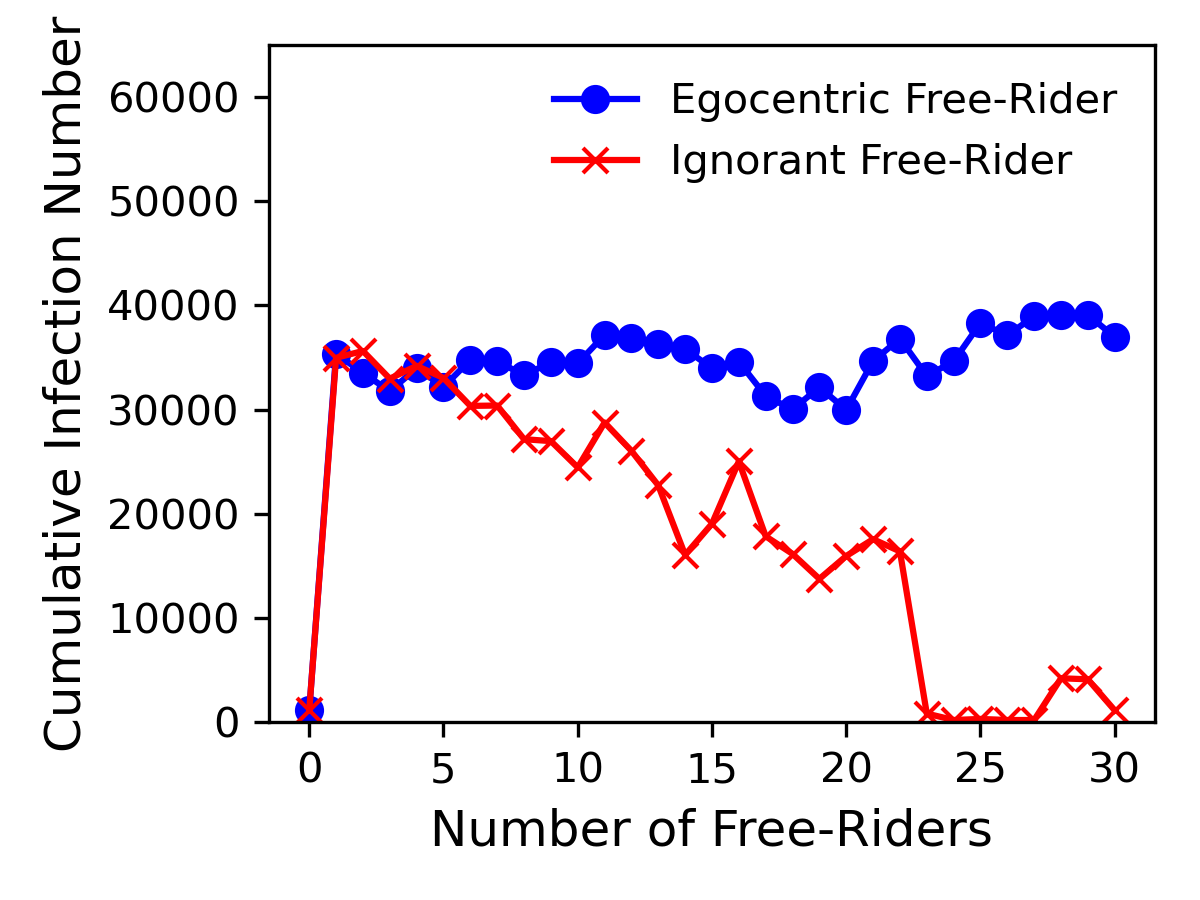}}
    \subfloat[\footnotesize $\tau^r=5$ and $\zeta^s_t=0.15$]{%
        \includegraphics[width=0.45\columnwidth]{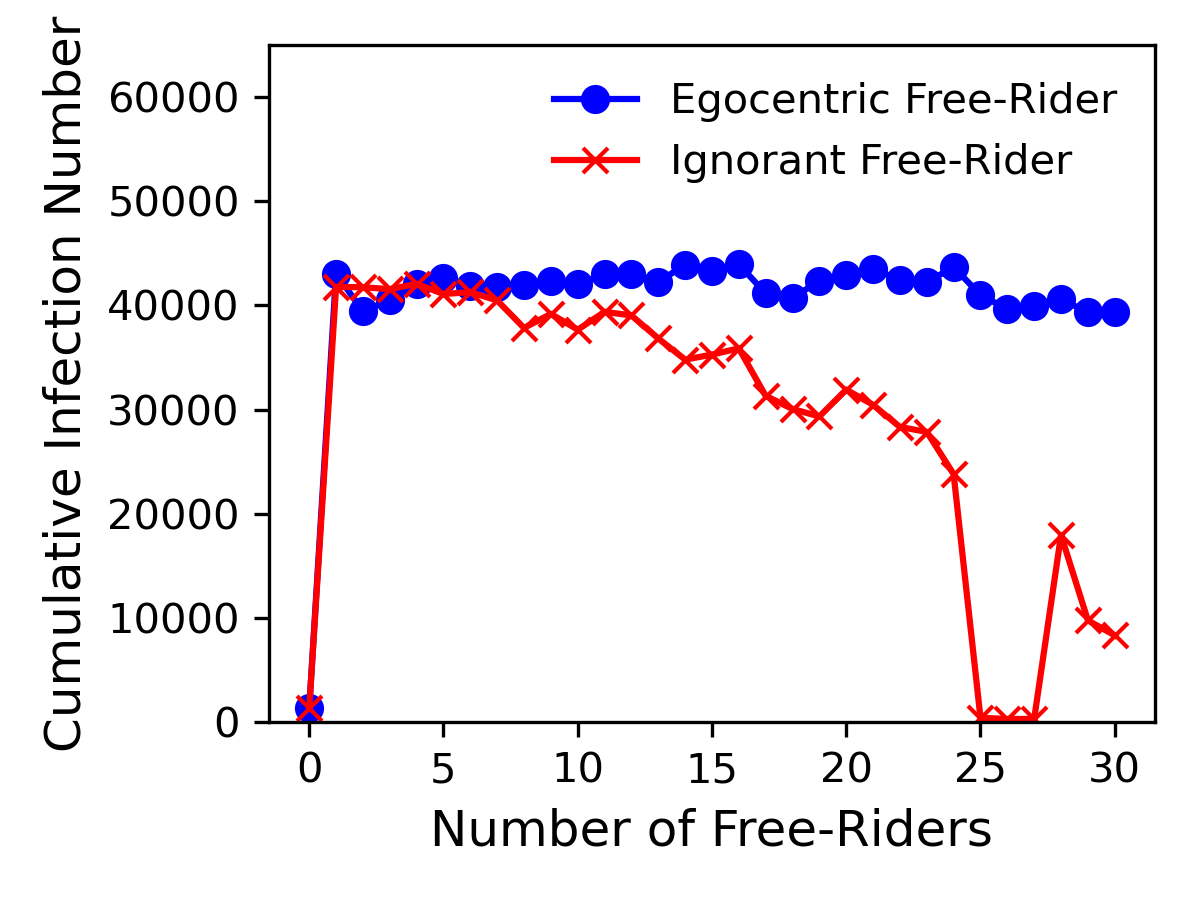}}
    \subfloat[\footnotesize $\tau^r=5$ and $\zeta^s_t=0.20$]{%
        \includegraphics[width=0.45\columnwidth]{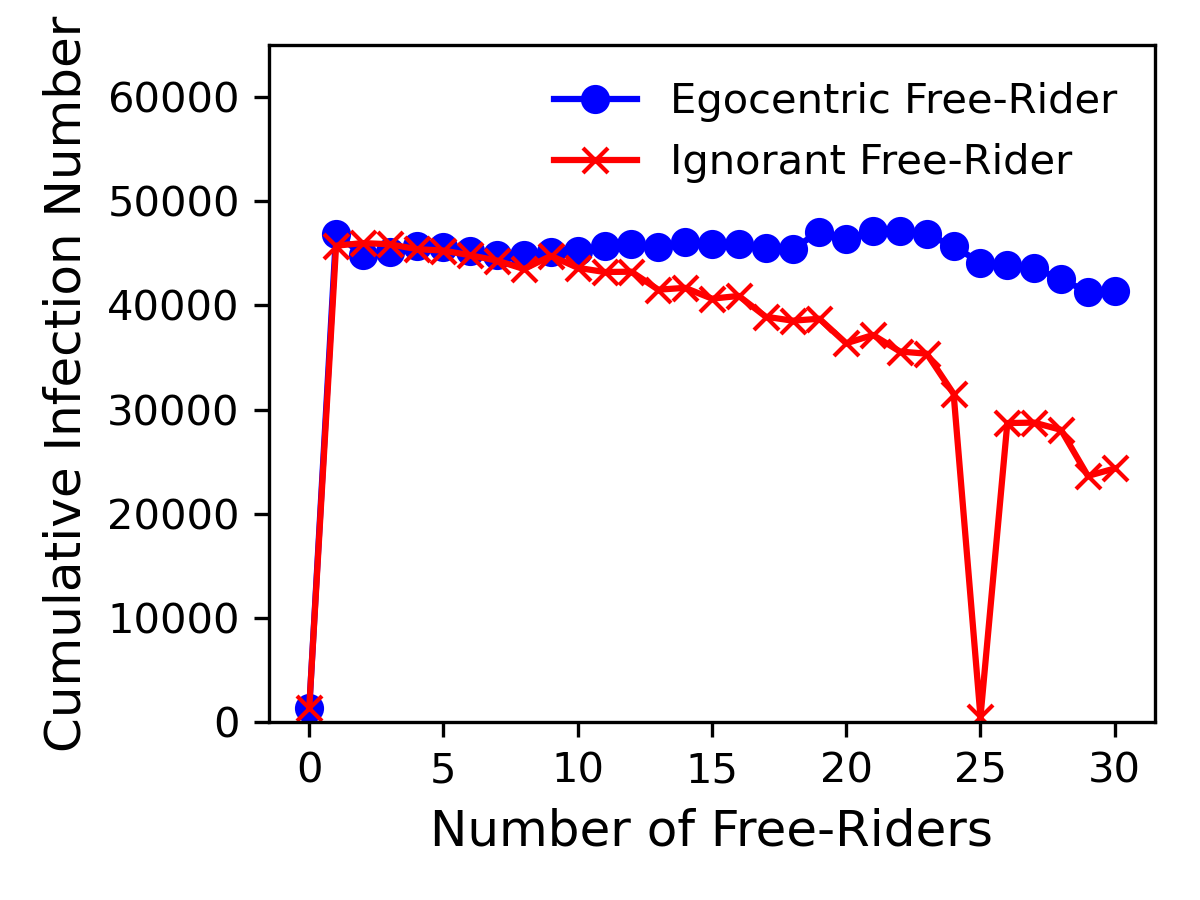}}
    \\
        \subfloat[\footnotesize $\tau^r=10$ and $\zeta^s_t=0.05$]{%
\includegraphics[width=0.45\columnwidth]{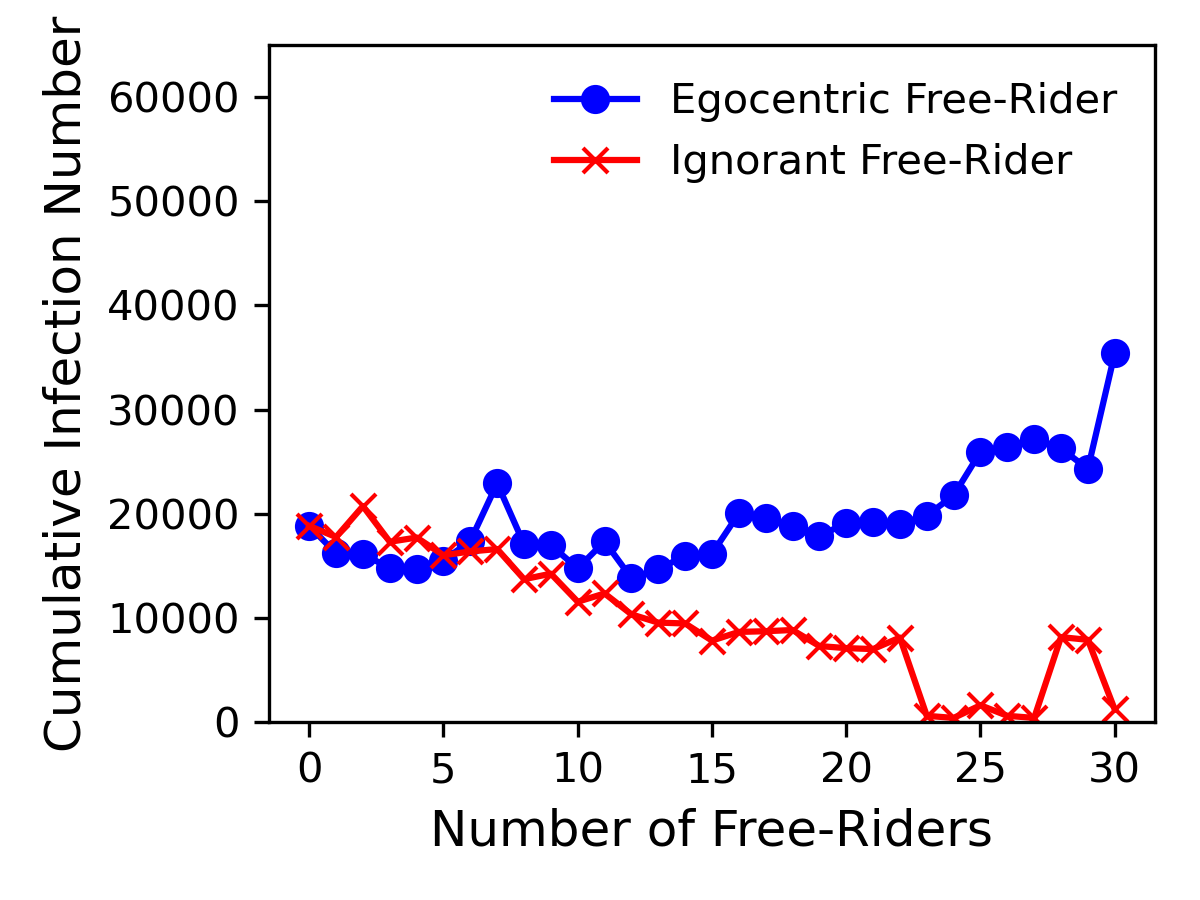}%
    }%
    \subfloat[\footnotesize $\tau^r=10$ and $\zeta^s_t=0.10$]{%
        \includegraphics[width=0.45\columnwidth]{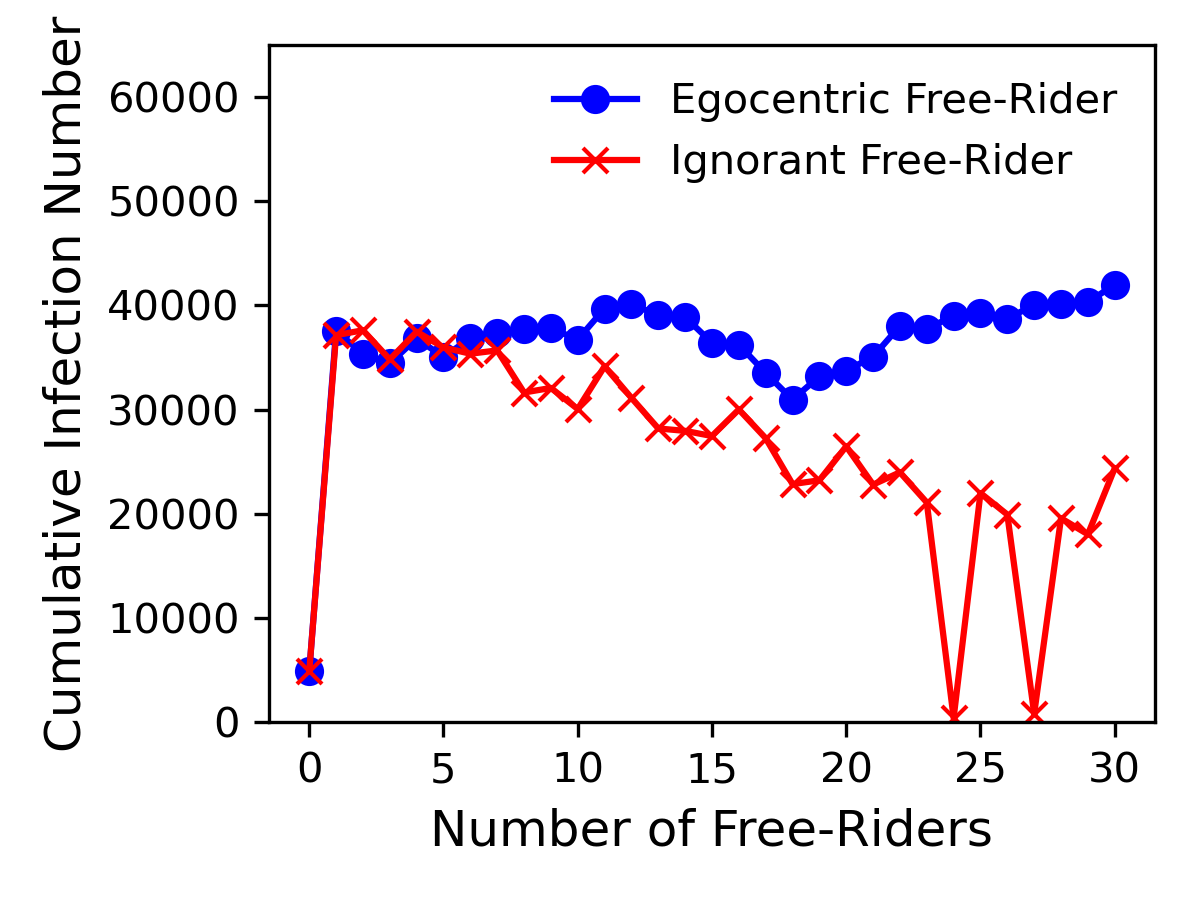}}
    \subfloat[\footnotesize $\tau^r=10$ and $\zeta^s_t=0.15$]{%
        \includegraphics[width=0.45\columnwidth]{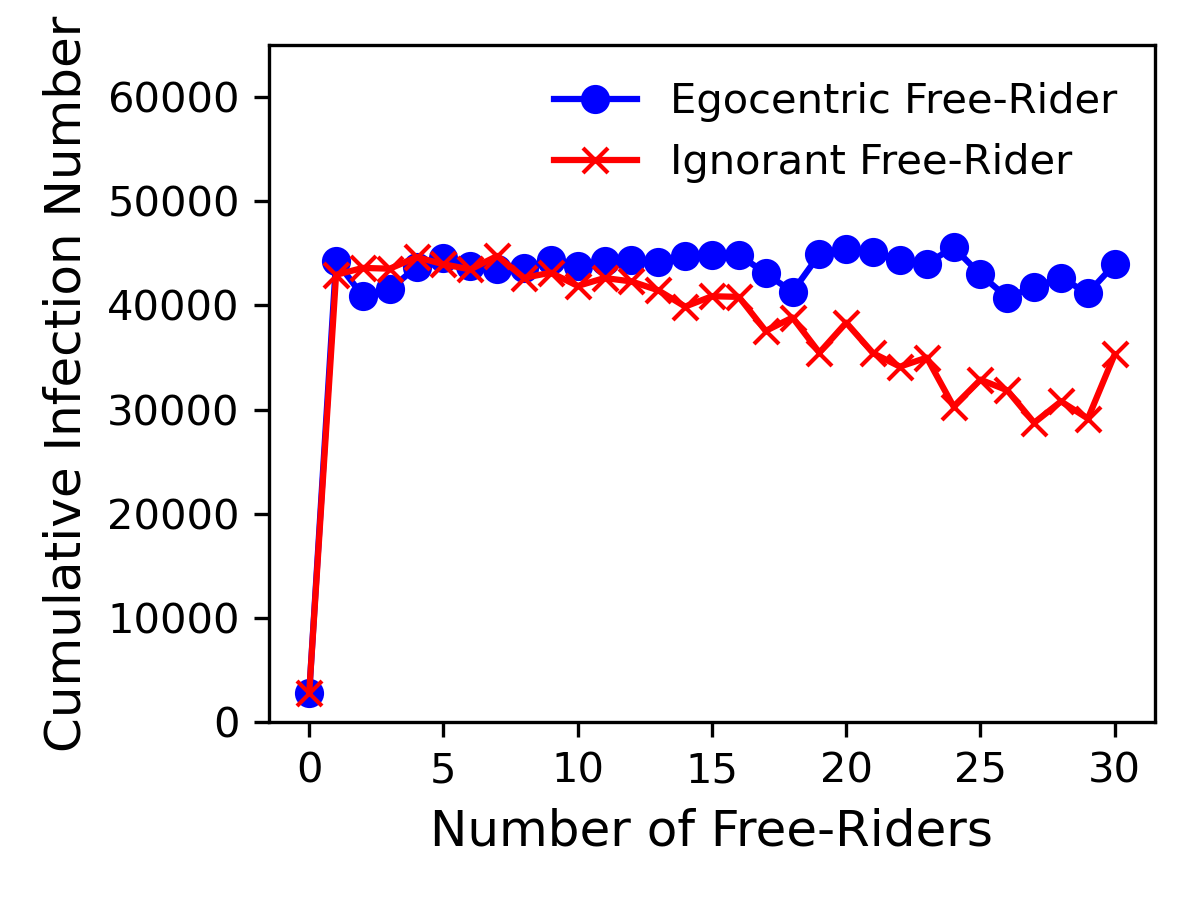}}
    \subfloat[\footnotesize $\tau^r=10$ and $\zeta^s_t=0.20$]{%
        \includegraphics[width=0.45\columnwidth]{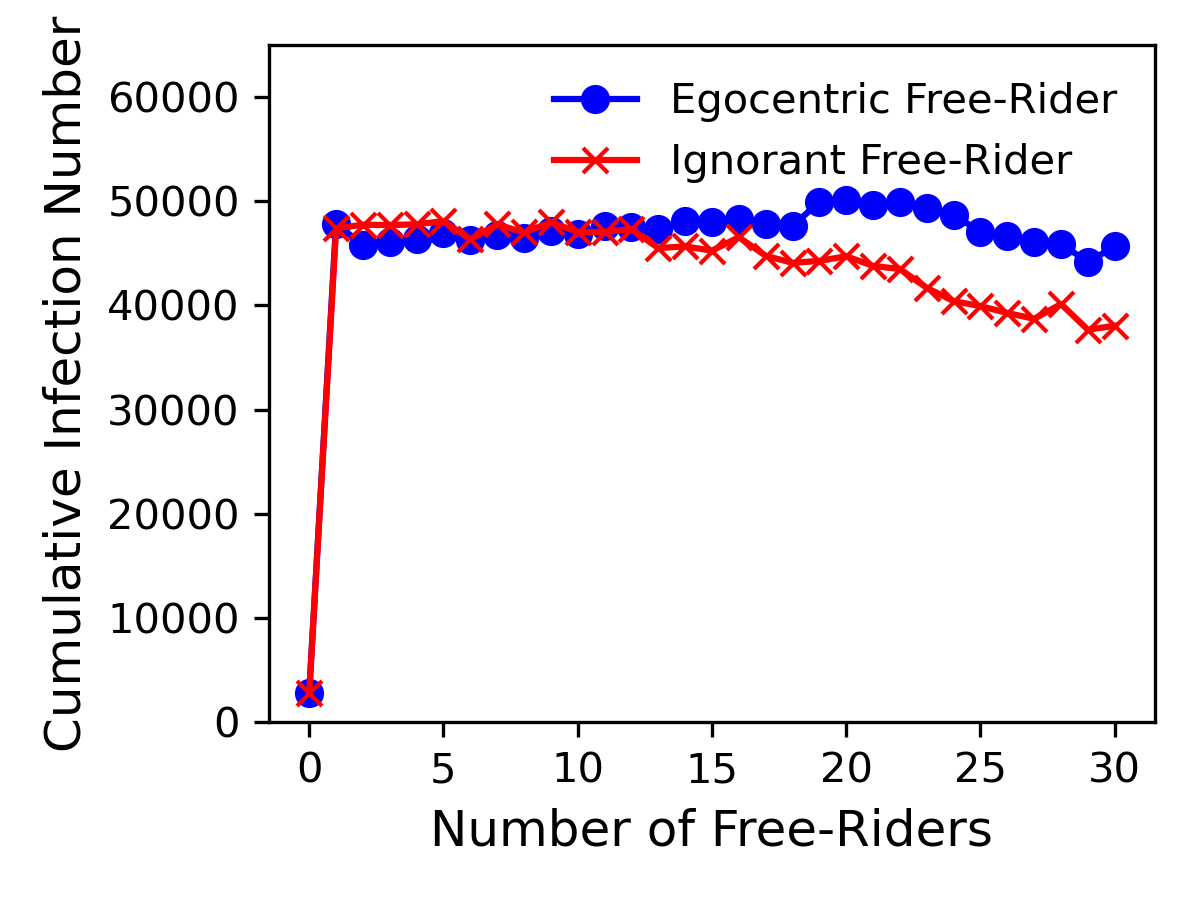}}
    \\
        \subfloat[\footnotesize $\tau^r=15$ and $\zeta^s_t=0.05$]{%
\includegraphics[width=0.45\columnwidth]{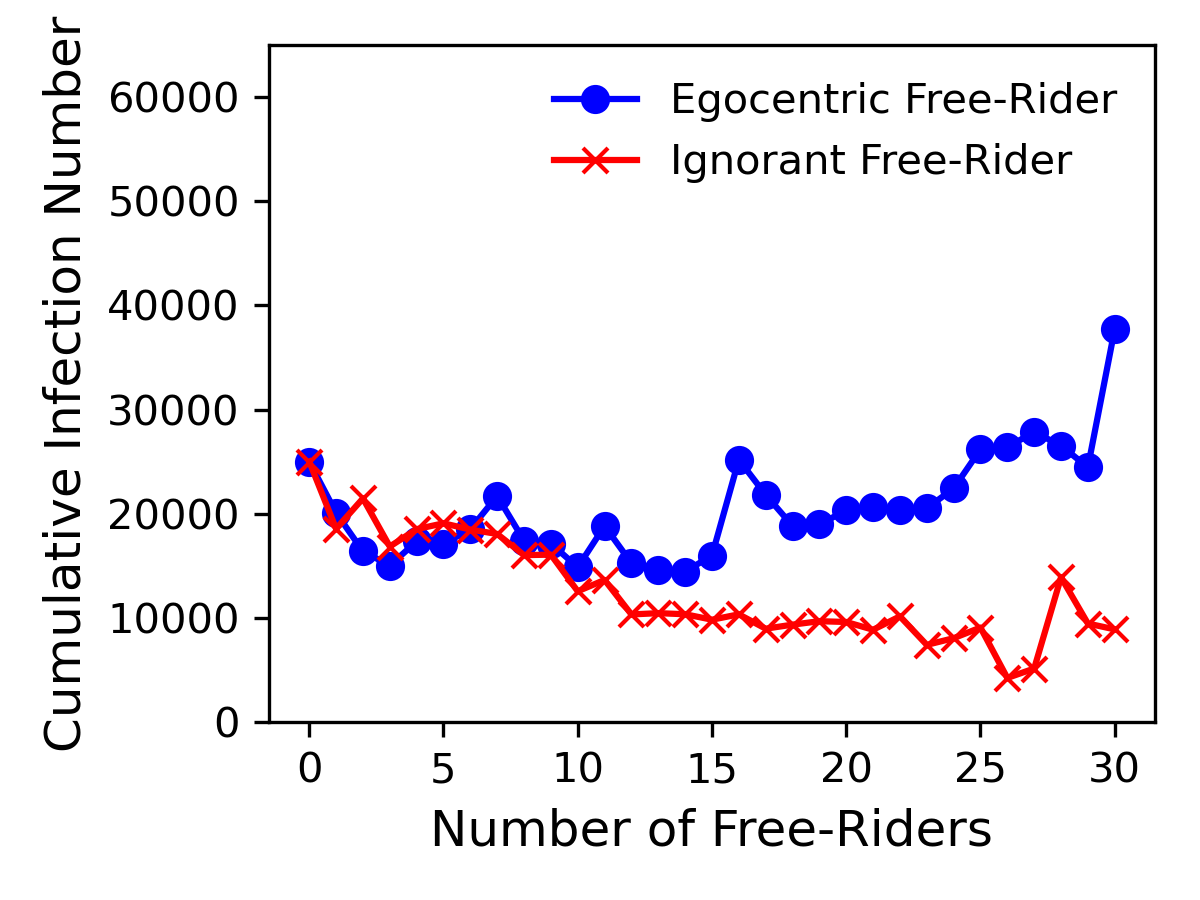}%
    }%
    \subfloat[\footnotesize $\tau^r=15$ and $\zeta^s_t=0.10$]{%
    \includegraphics[width=0.45\columnwidth]{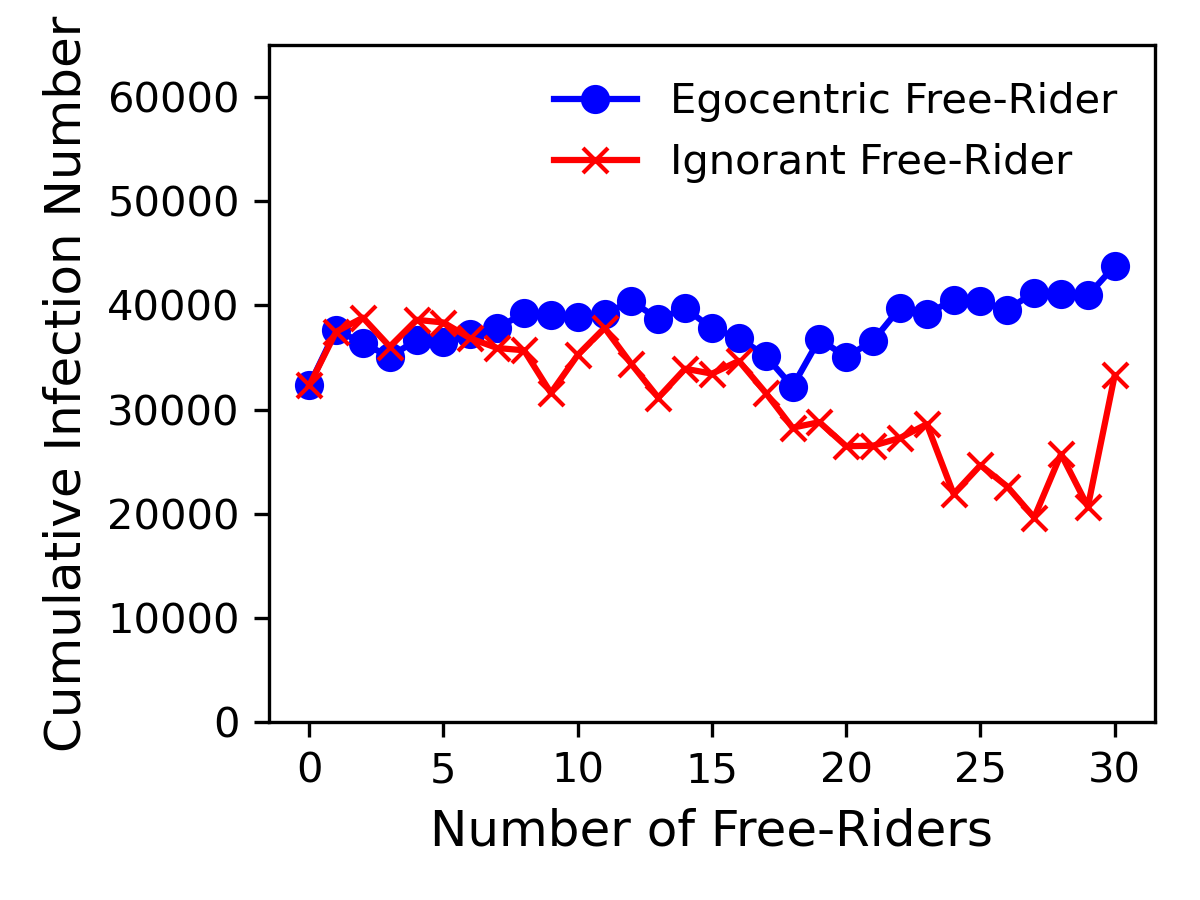}}
    \subfloat[\footnotesize $\tau^r=15$ and $\zeta^s_t=0.15$]{%
        \includegraphics[width=0.45\columnwidth]{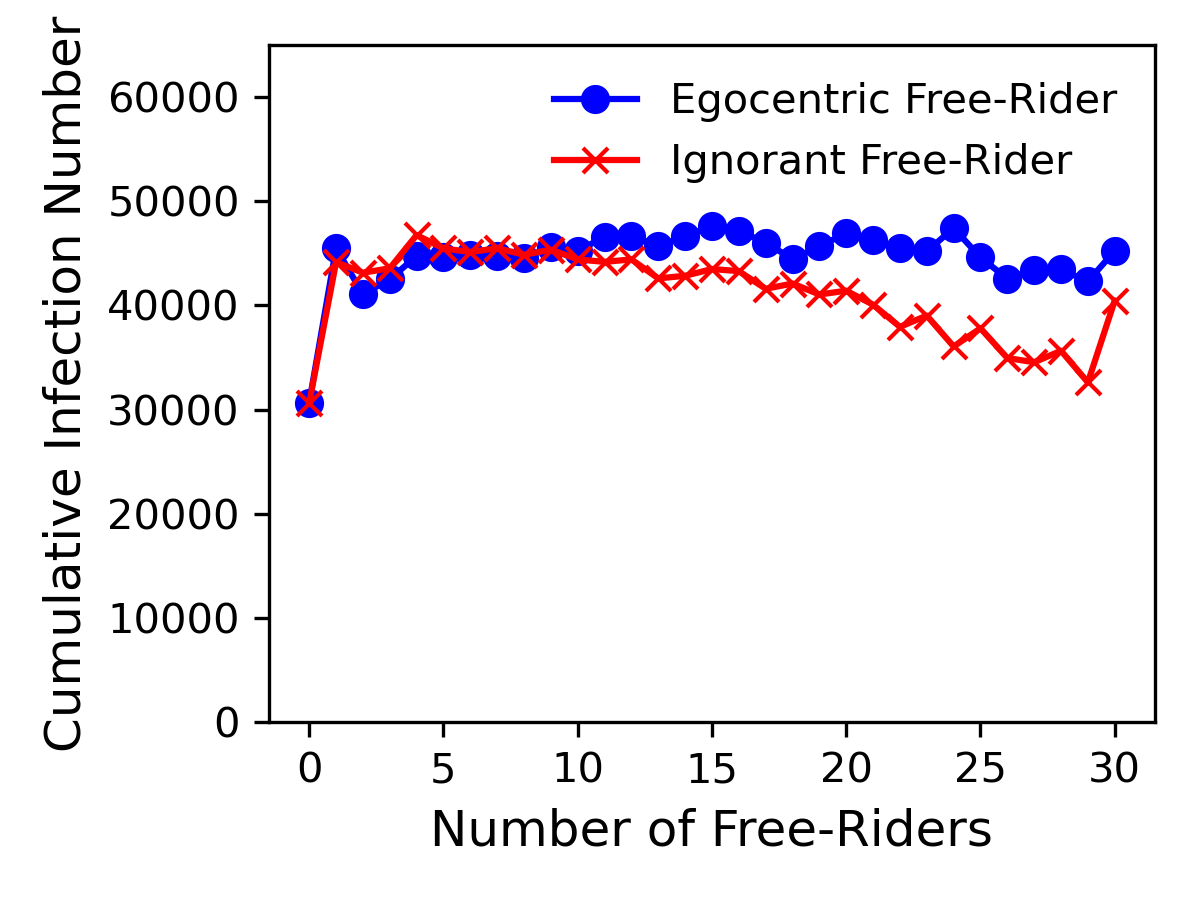}}
    \subfloat[\footnotesize $\tau^r=15$ and $\zeta^s_t=0.20$]{%
        \includegraphics[width=0.45\columnwidth]{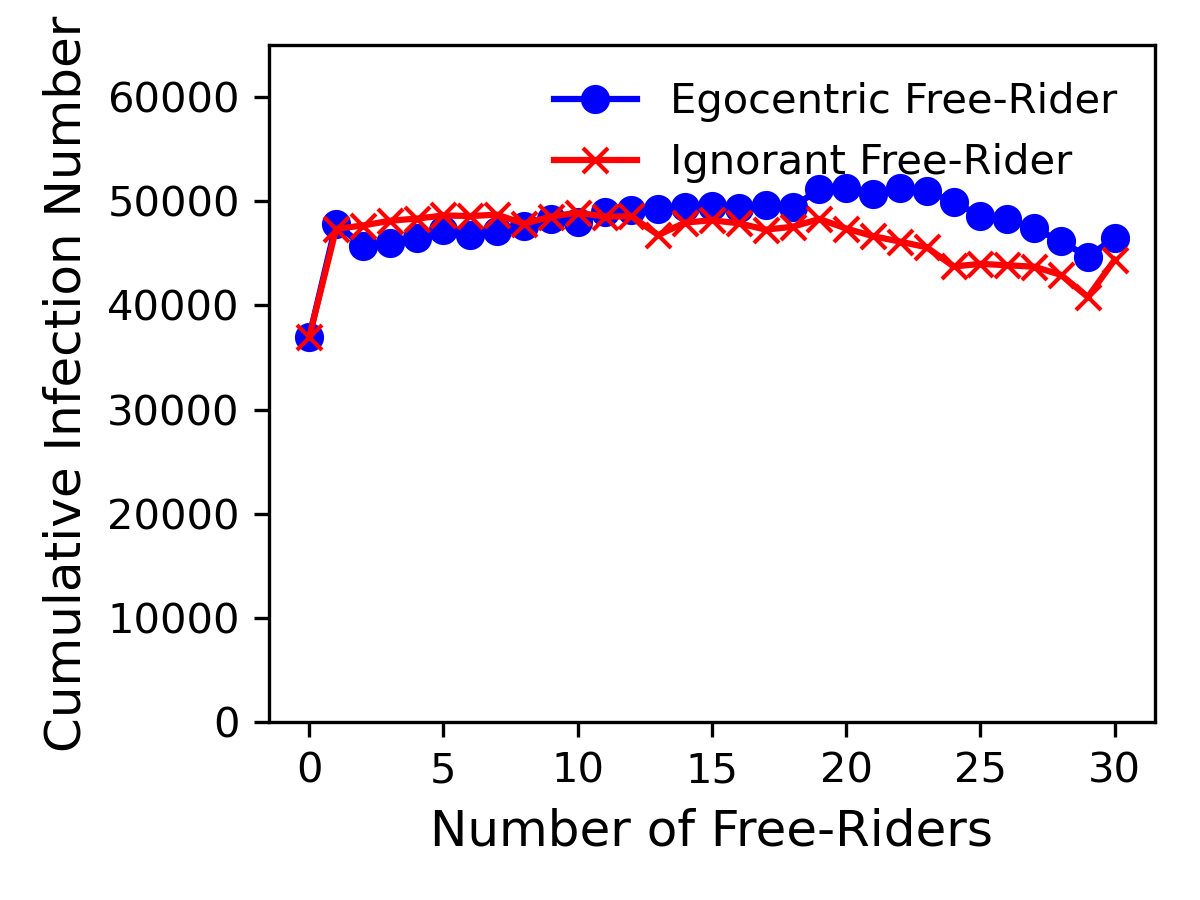}}
    \\
        \subfloat[\footnotesize $\tau^r=20$ and $\zeta^s_t=0.05$]{%
\includegraphics[width=0.45\columnwidth]{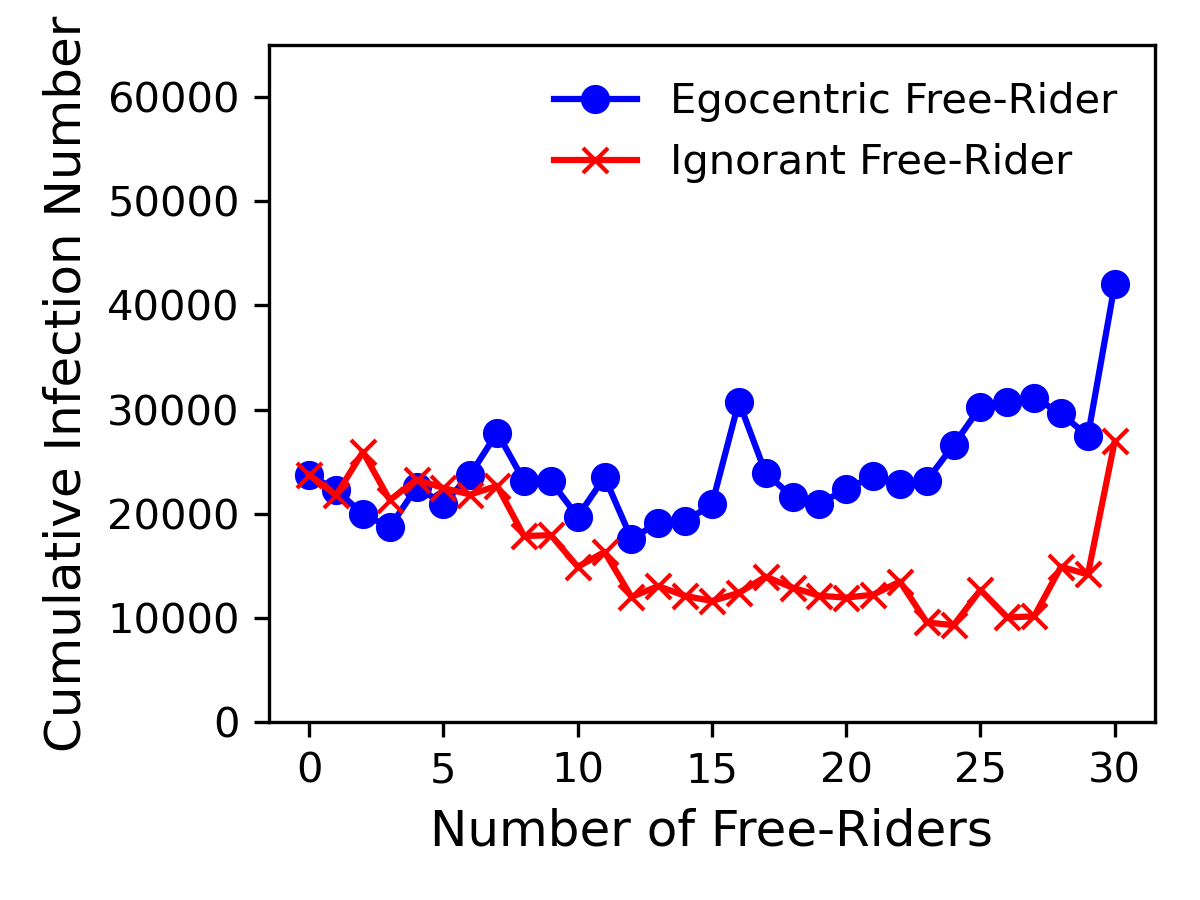}%
    }%
    \subfloat[\footnotesize $\tau^r=20$ and $\zeta^s_t=0.10$]{%
        \includegraphics[width=0.45\columnwidth]{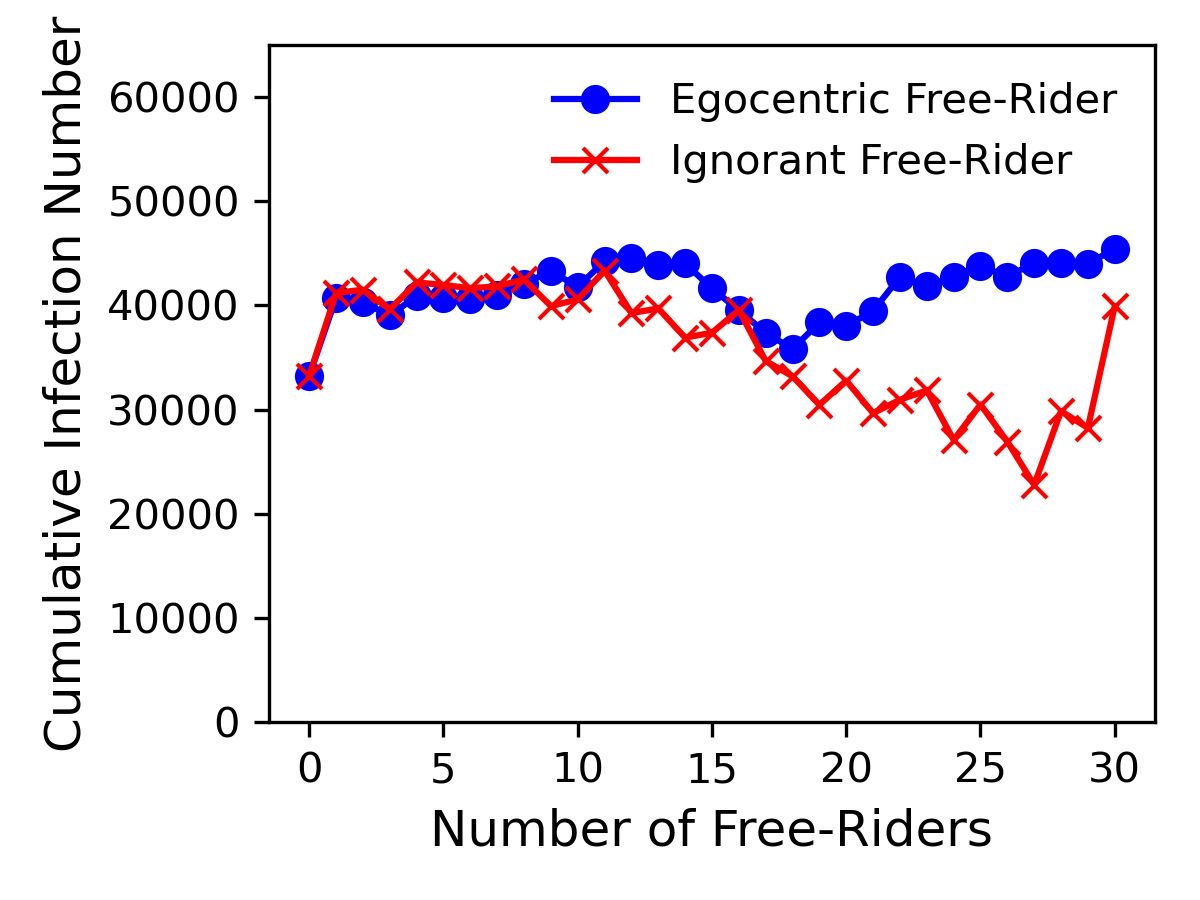}}
    \subfloat[\footnotesize $\tau^r=20$ and $\zeta^s_t=0.15$]{%
        \includegraphics[width=0.45\columnwidth]{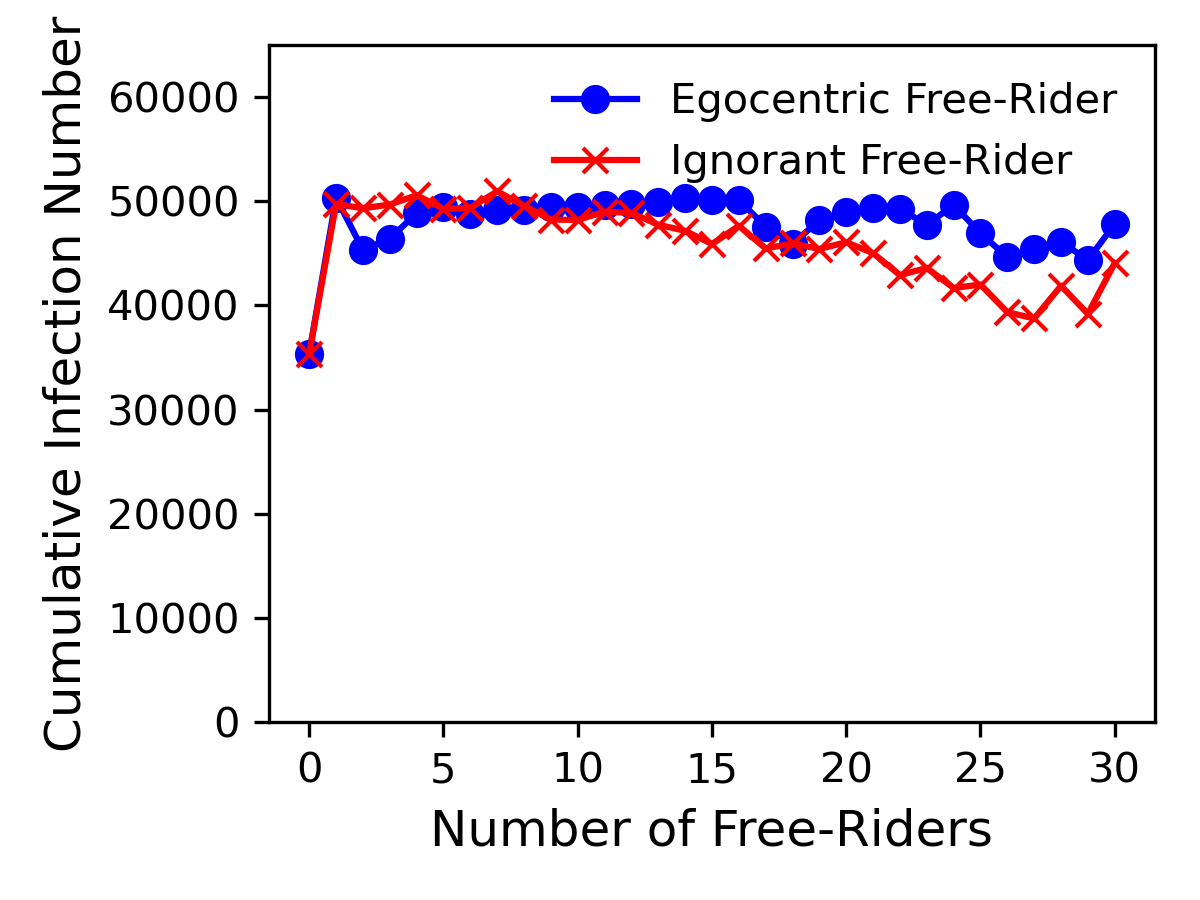}}
    \subfloat[\footnotesize $\tau^r=20$ and $\zeta^s_t=0.20$]{%
        \includegraphics[width=0.45\columnwidth]{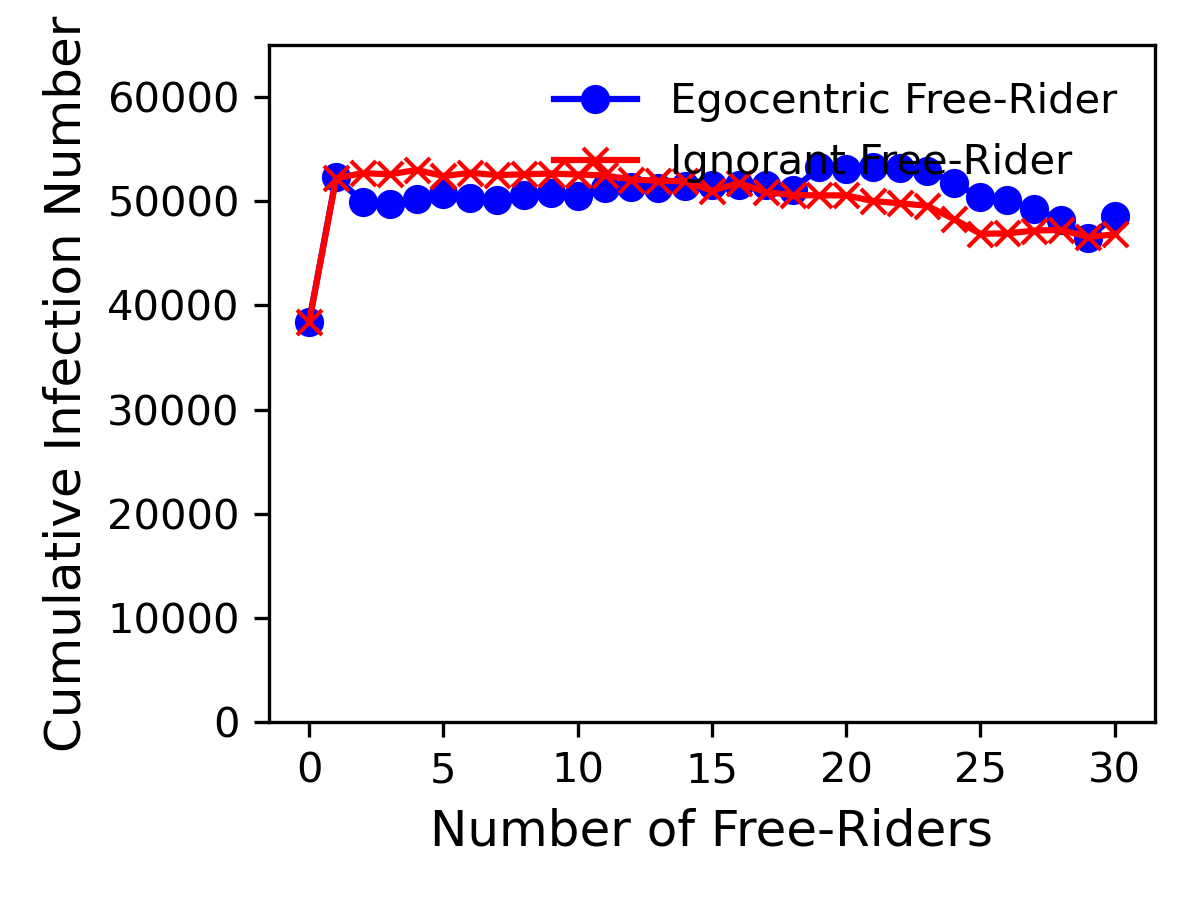}}
	\caption{The cumulative infection occurrences in "free-rider" scenarios given various setups of infection rate and recovery time.}
 \label{infectResilience2}
\end{figure*}

Fig.~\ref{infectResilience2} shows the cumulative total infection number under a varying infection rate and recovery time in the network simulations that are driven by mixed preference mutation styles. The cumulative total infection number increases with higher infection rates and longer recovery time values. The introduction of the "free-rider" leads to more infections in cooperative networks, where more egocentric "free-riders" lead to persistent high infection levels and ignorant "free-riders" lead to a gradual infection number decrease. With the increase in recovery time and infection rates, the cumulative total infection number keeps at higher values with fewer fluctuations. This suggests that severe epidemic spread can broadly affect infections irrespective of "free-rider" numbers.

\begin{figure*}[htp]
    \centering
    \subfloat[\footnotesize $\tau^r=5$ and $\zeta^s_t=0.05$]{%
\includegraphics[width=0.45\columnwidth]{ResiliencePlots/Compare_RiderRewardRT5Inf0.05.png}%
    }%
    \subfloat[\footnotesize $\tau^r=5$ and $\zeta^s_t=0.10$]{%
    \includegraphics[width=0.45\columnwidth]{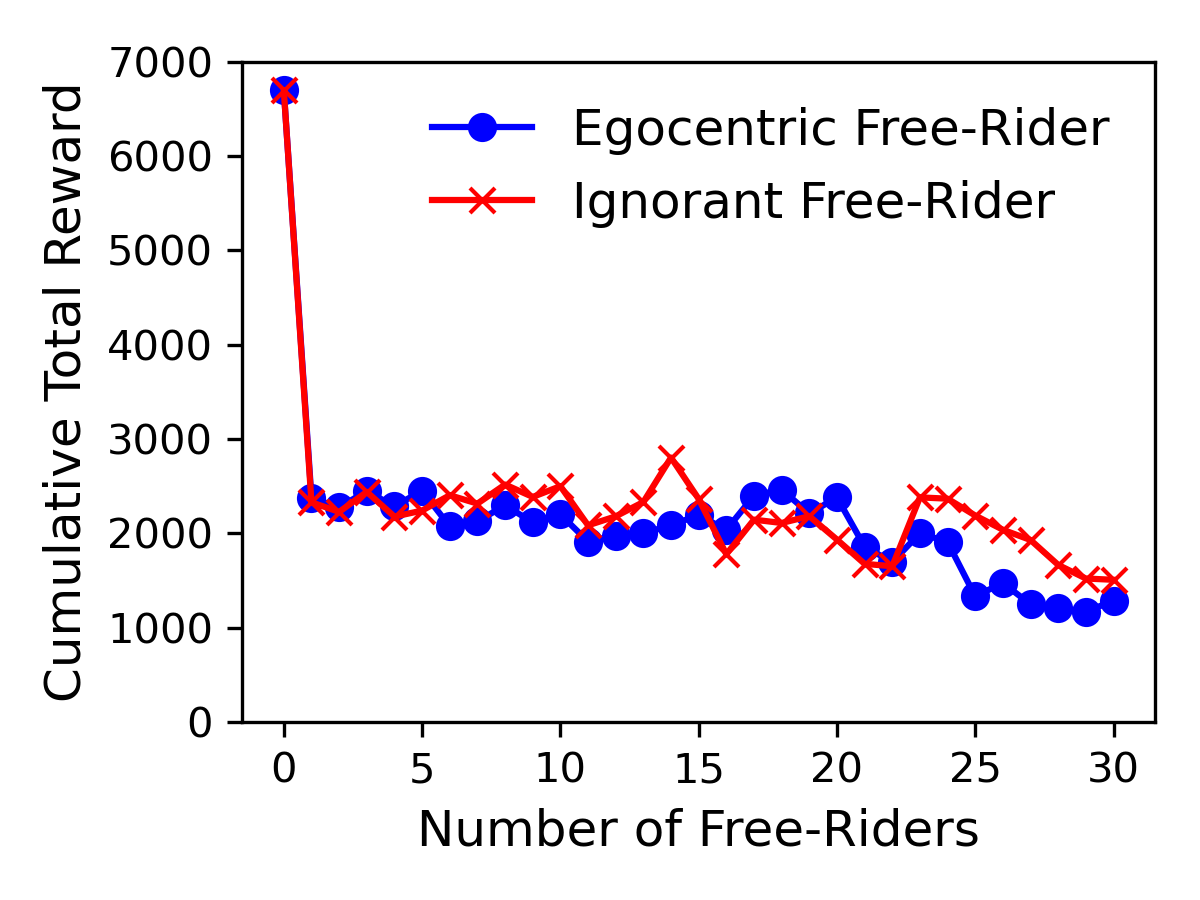}}
    \subfloat[\footnotesize $\tau^r=5$ and $\zeta^s_t=0.15$]{%
        \includegraphics[width=0.45\columnwidth]{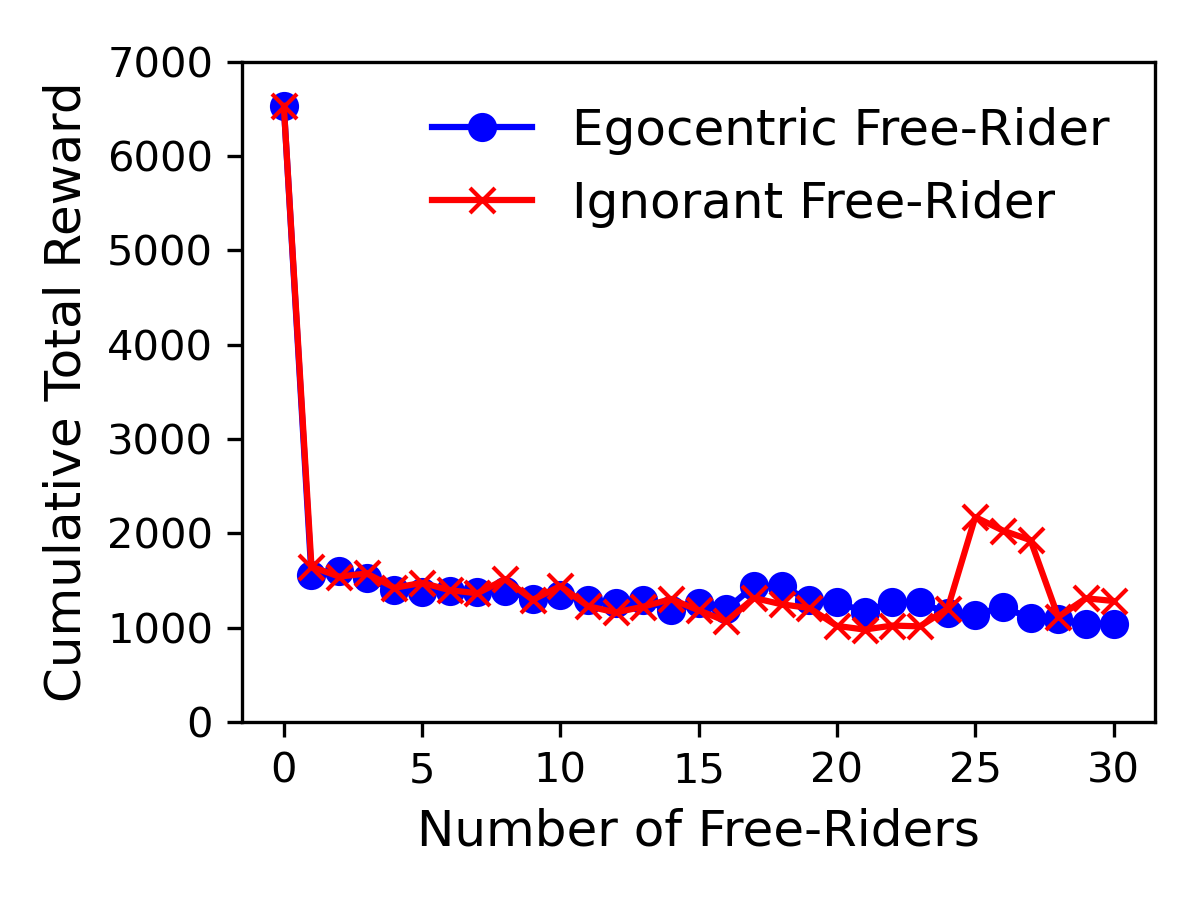}}
    \subfloat[\footnotesize $\tau^r=5$ and $\zeta^s_t=0.20$]{%
        \includegraphics[width=0.45\columnwidth]{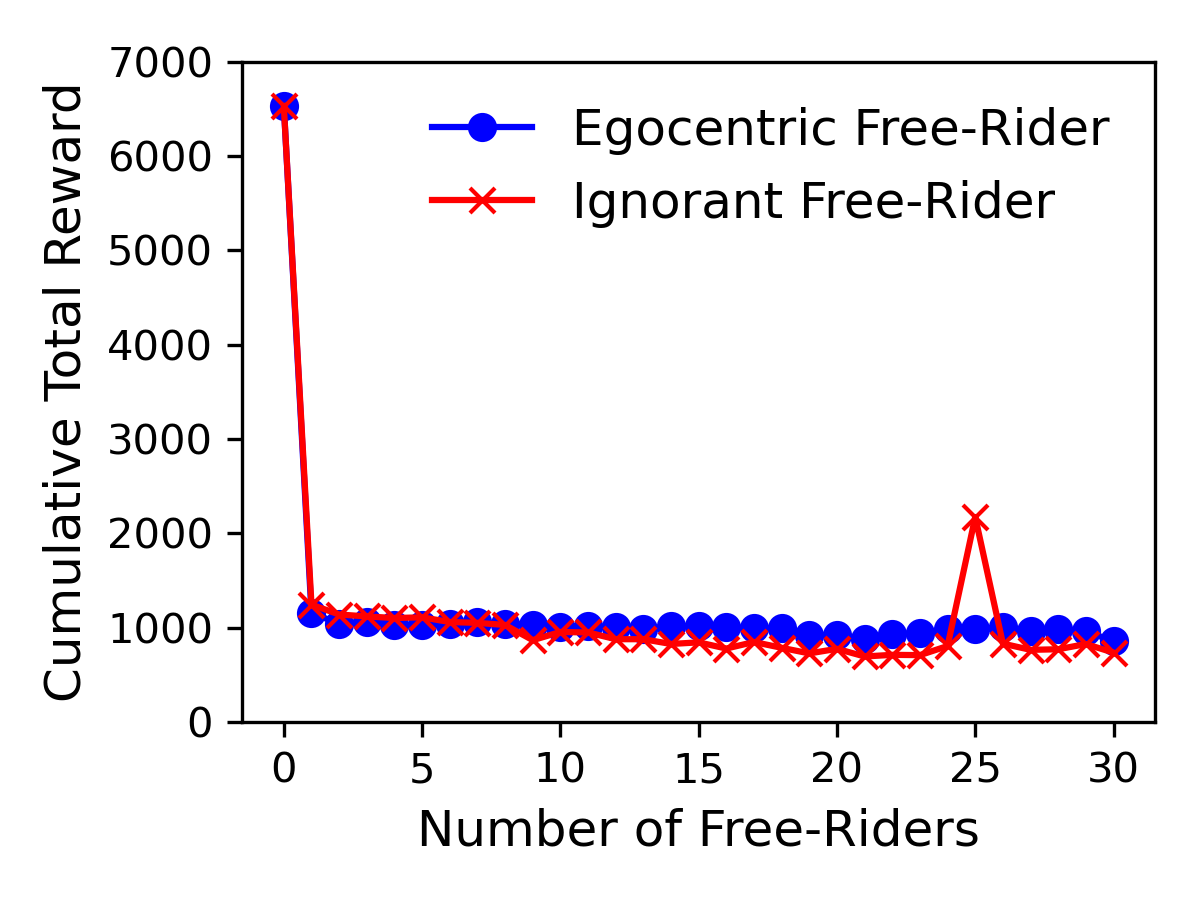}}
    \\
        \subfloat[\footnotesize $\tau^r=10$ and $\zeta^s_t=0.05$]{%
\includegraphics[width=0.45\columnwidth]{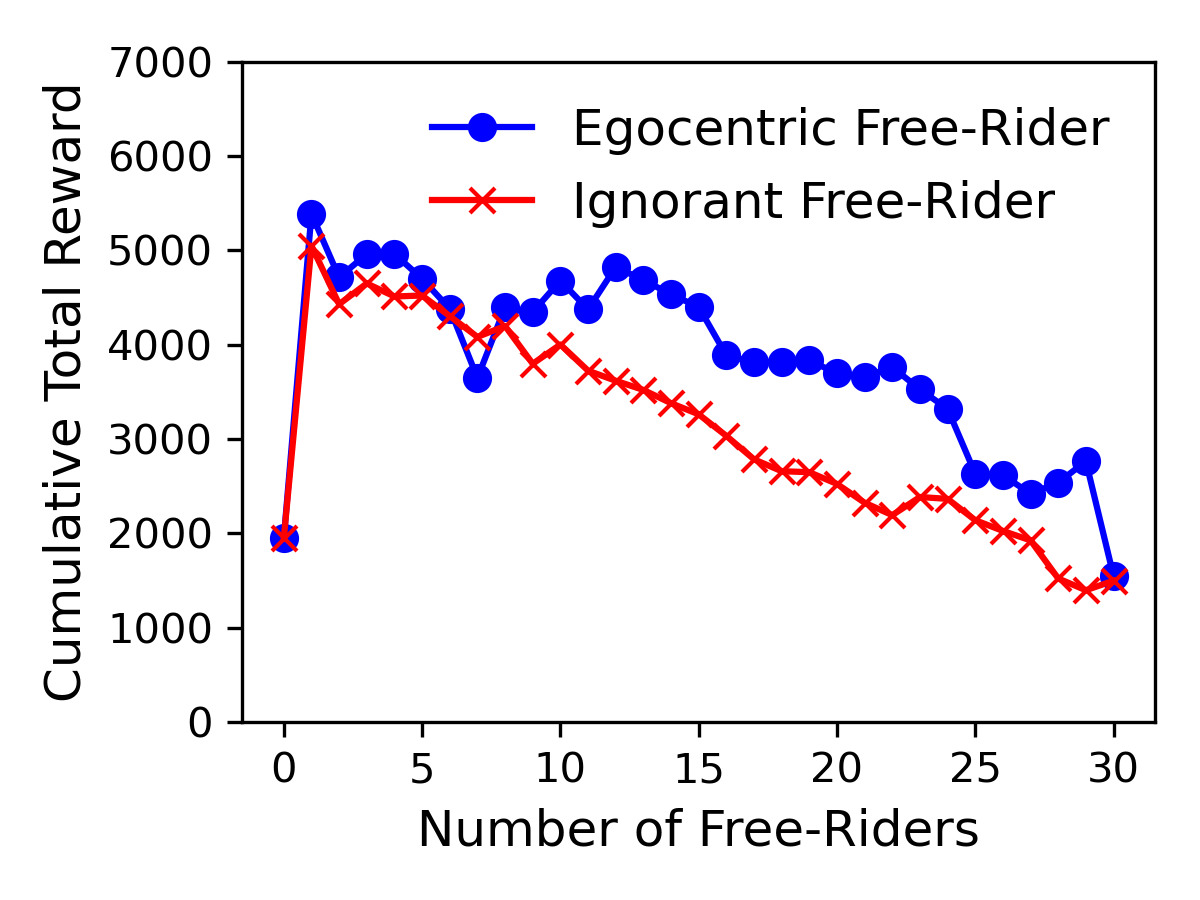}%
    }%
    \subfloat[\footnotesize $\tau^r=10$ and $\zeta^s_t=0.10$]{%
        \includegraphics[width=0.45\columnwidth]{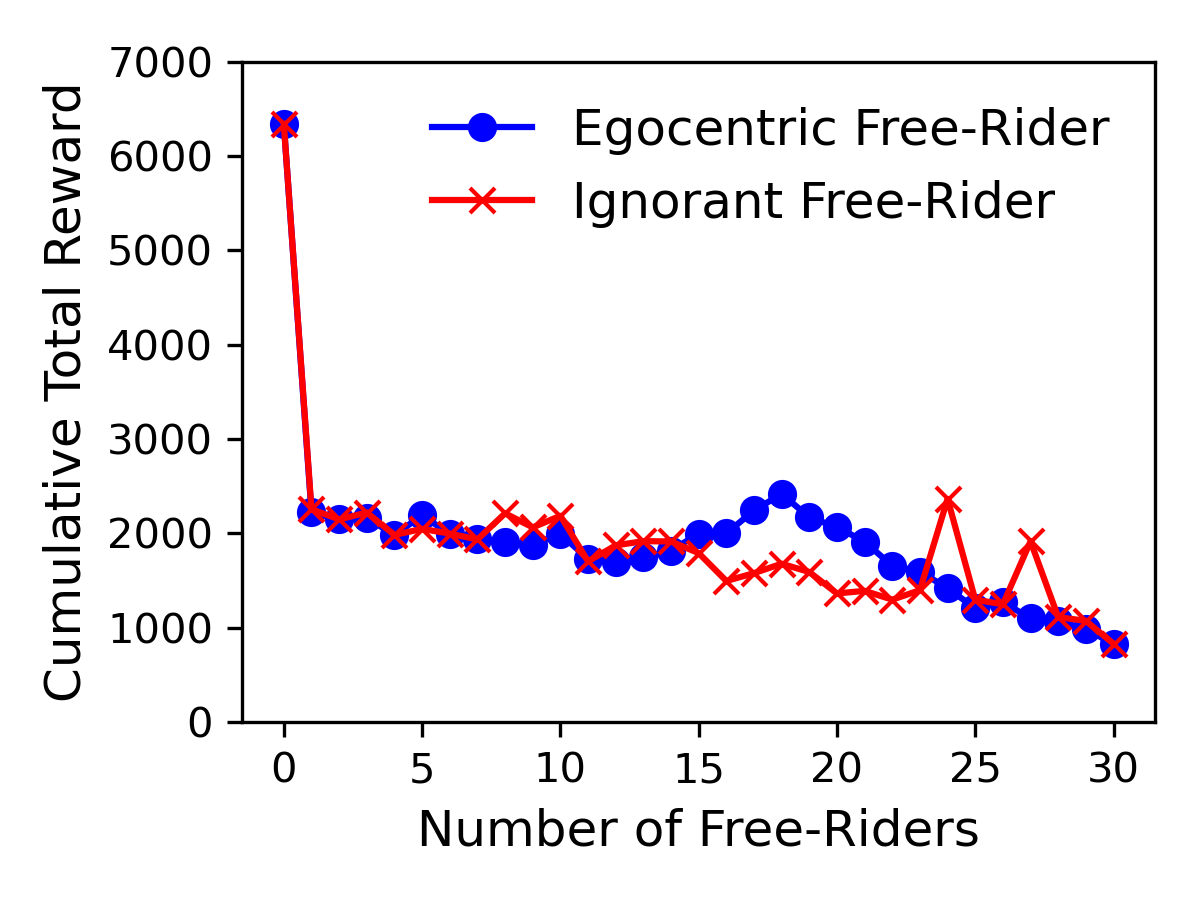}}
    \subfloat[\footnotesize $\tau^r=10$ and $\zeta^s_t=0.15$]{%
        \includegraphics[width=0.45\columnwidth]{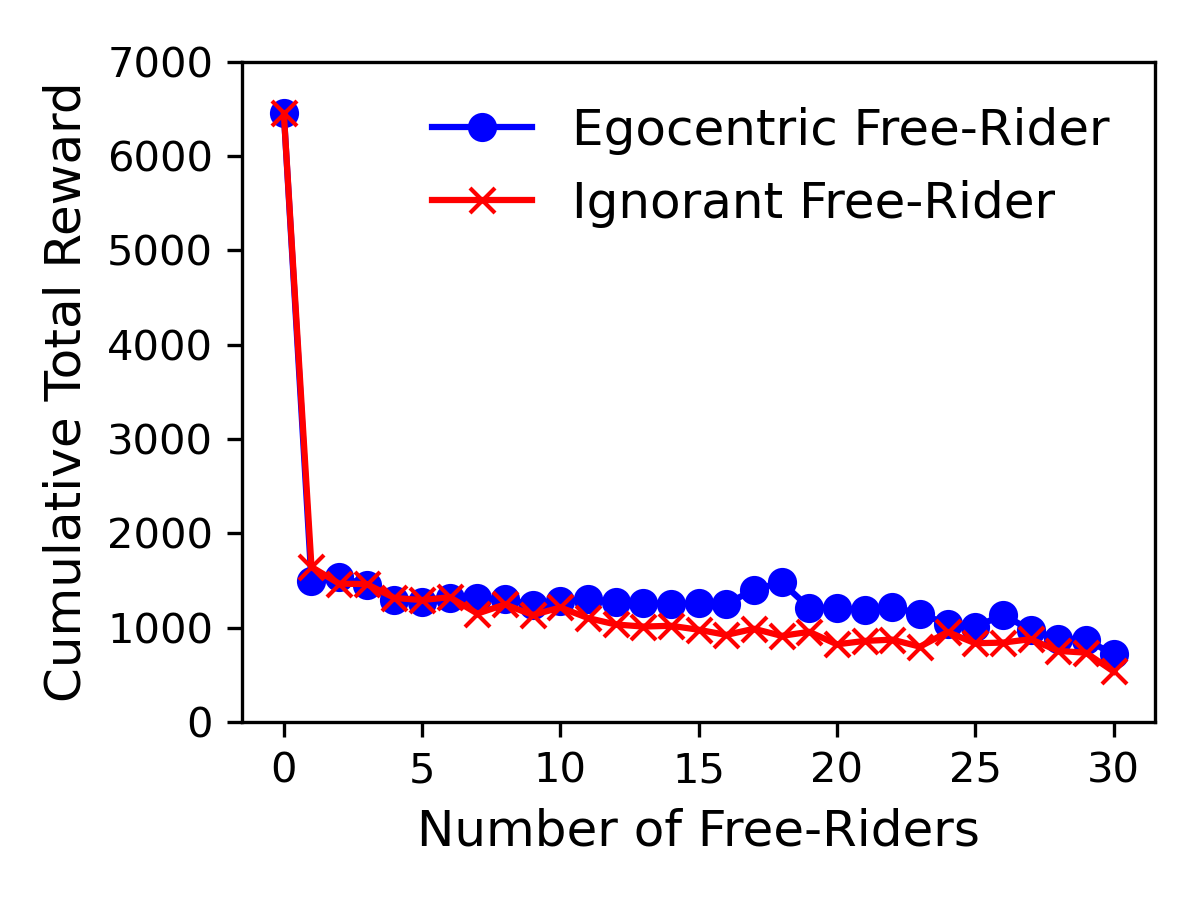}}
    \subfloat[\footnotesize $\tau^r=10$ and $\zeta^s_t=0.20$]{%
        \includegraphics[width=0.45\columnwidth]{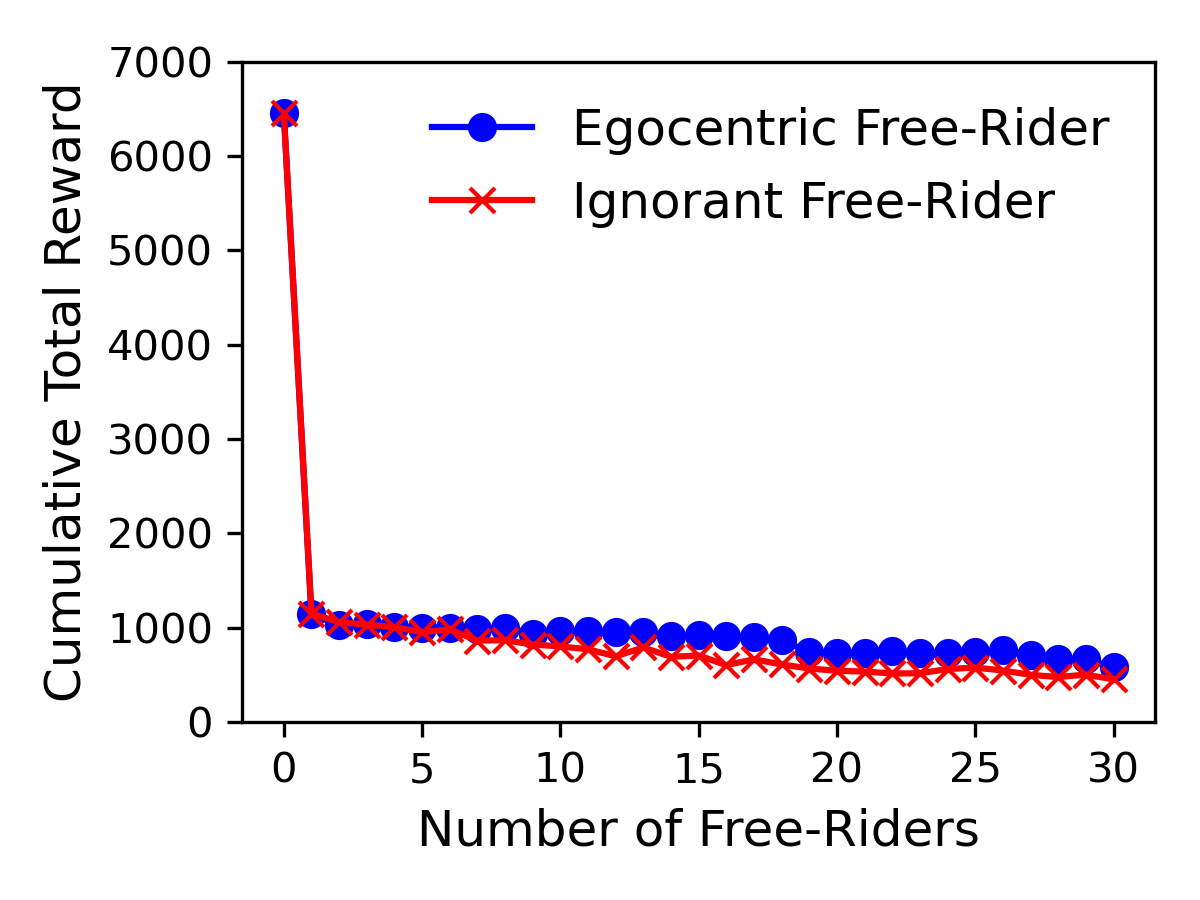}}
    \\
        \subfloat[\footnotesize $\tau^r=15$ and $\zeta^s_t=0.05$]{%
\includegraphics[width=0.45\columnwidth]{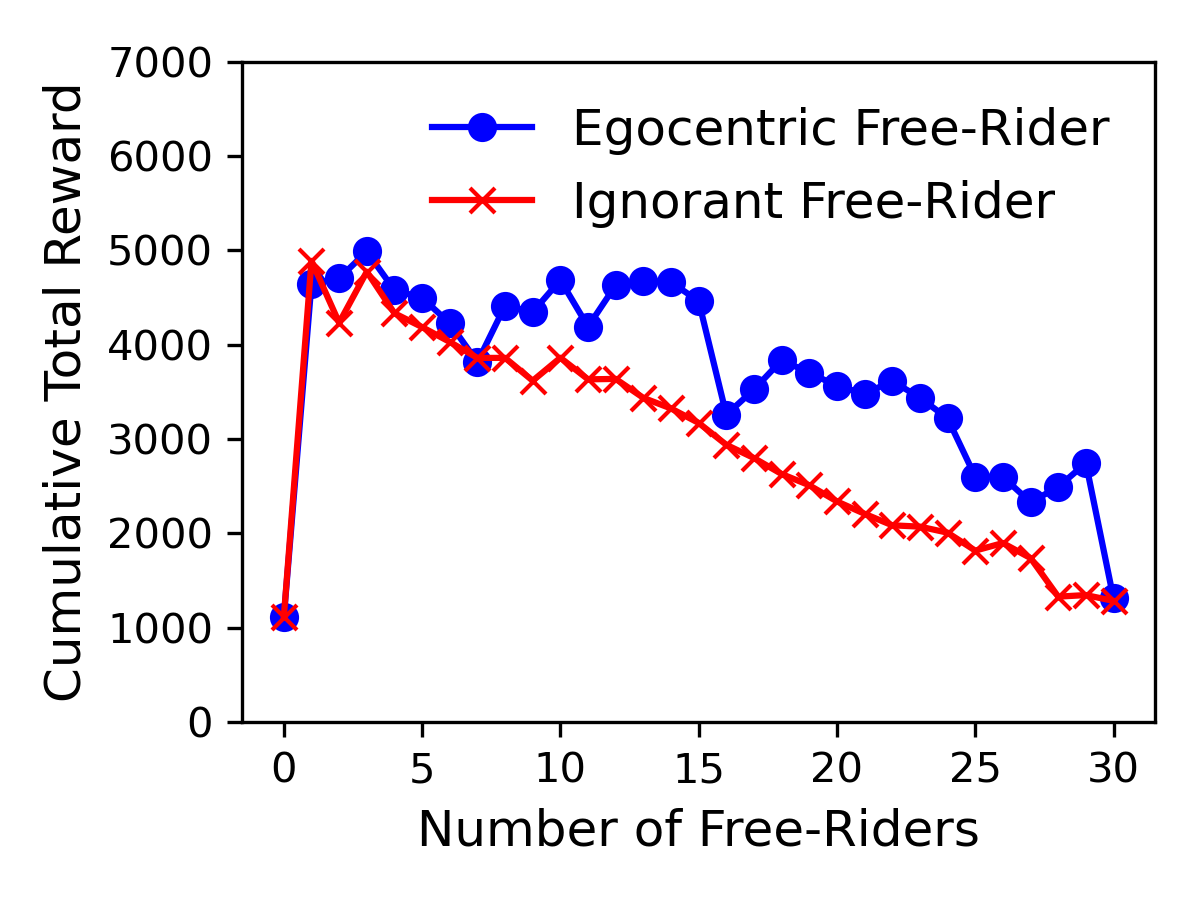}%
    }%
    \subfloat[\footnotesize $\tau^r=15$ and $\zeta^s_t=0.10$]{%
    \includegraphics[width=0.45\columnwidth]{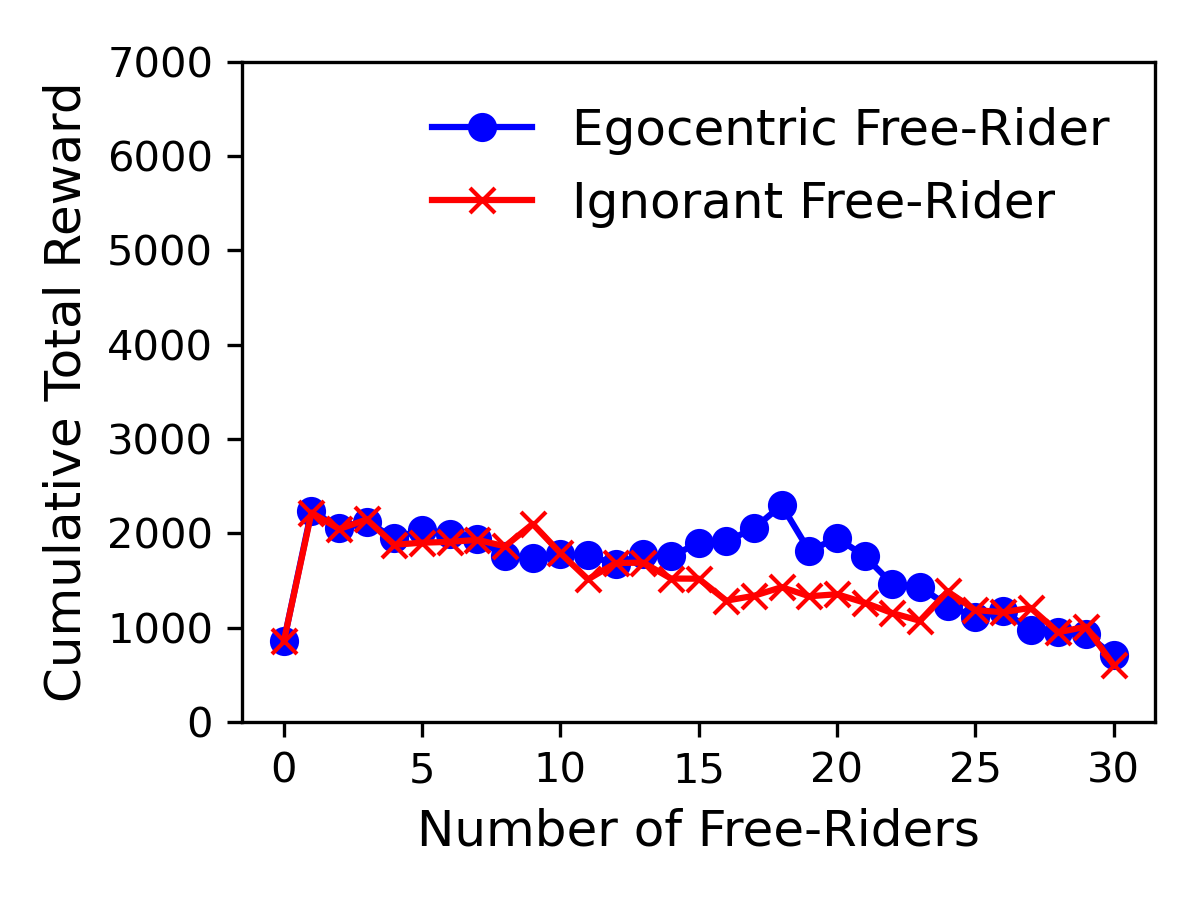}}
    \subfloat[\footnotesize $\tau^r=15$ and $\zeta^s_t=0.15$]{%
        \includegraphics[width=0.45\columnwidth]{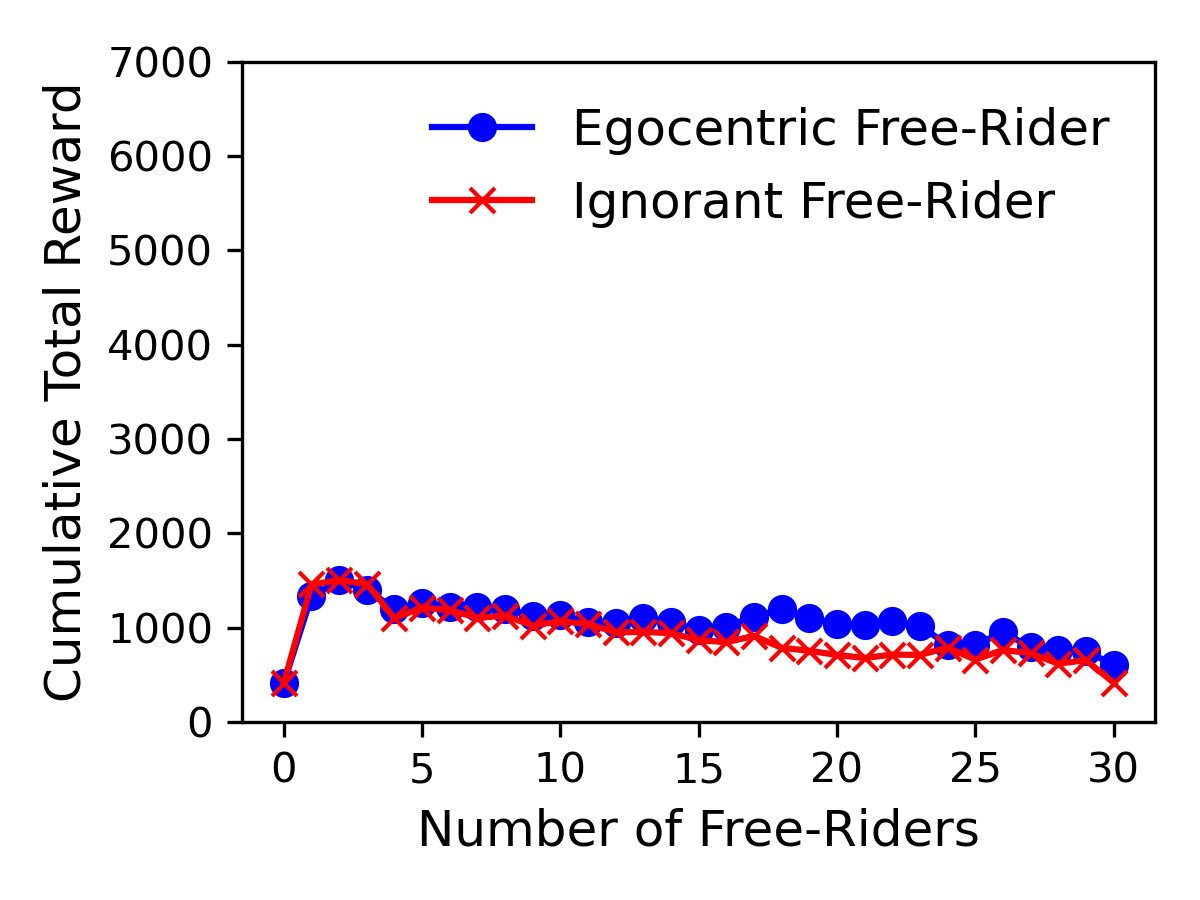}}
    \subfloat[\footnotesize $\tau^r=15$ and $\zeta^s_t=0.20$]{%
        \includegraphics[width=0.45\columnwidth]{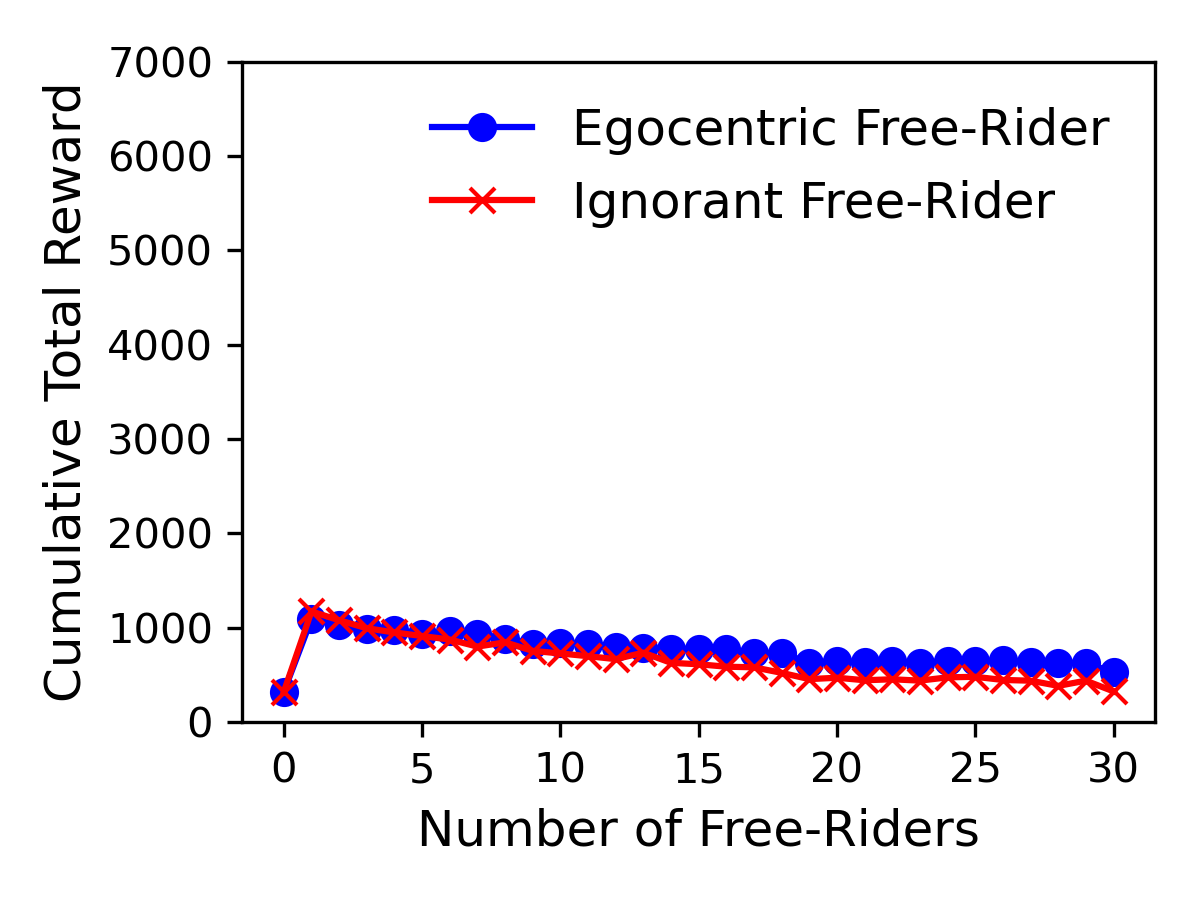}}
    \\
        \subfloat[\footnotesize $\tau^r=20$ and $\zeta^s_t=0.05$]{%
\includegraphics[width=0.45\columnwidth]{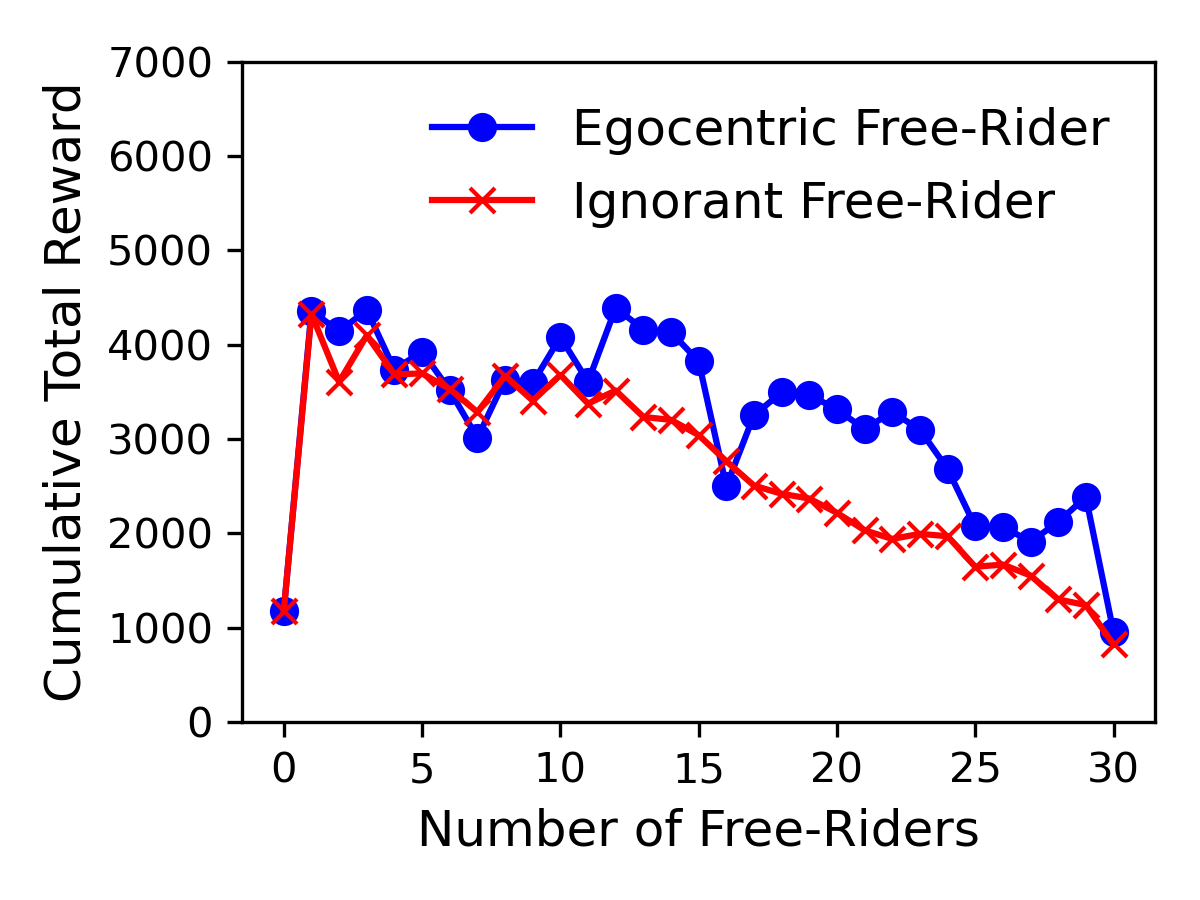}%
    }%
    \subfloat[\footnotesize $\tau^r=20$ and $\zeta^s_t=0.10$]{%
        \includegraphics[width=0.45\columnwidth]{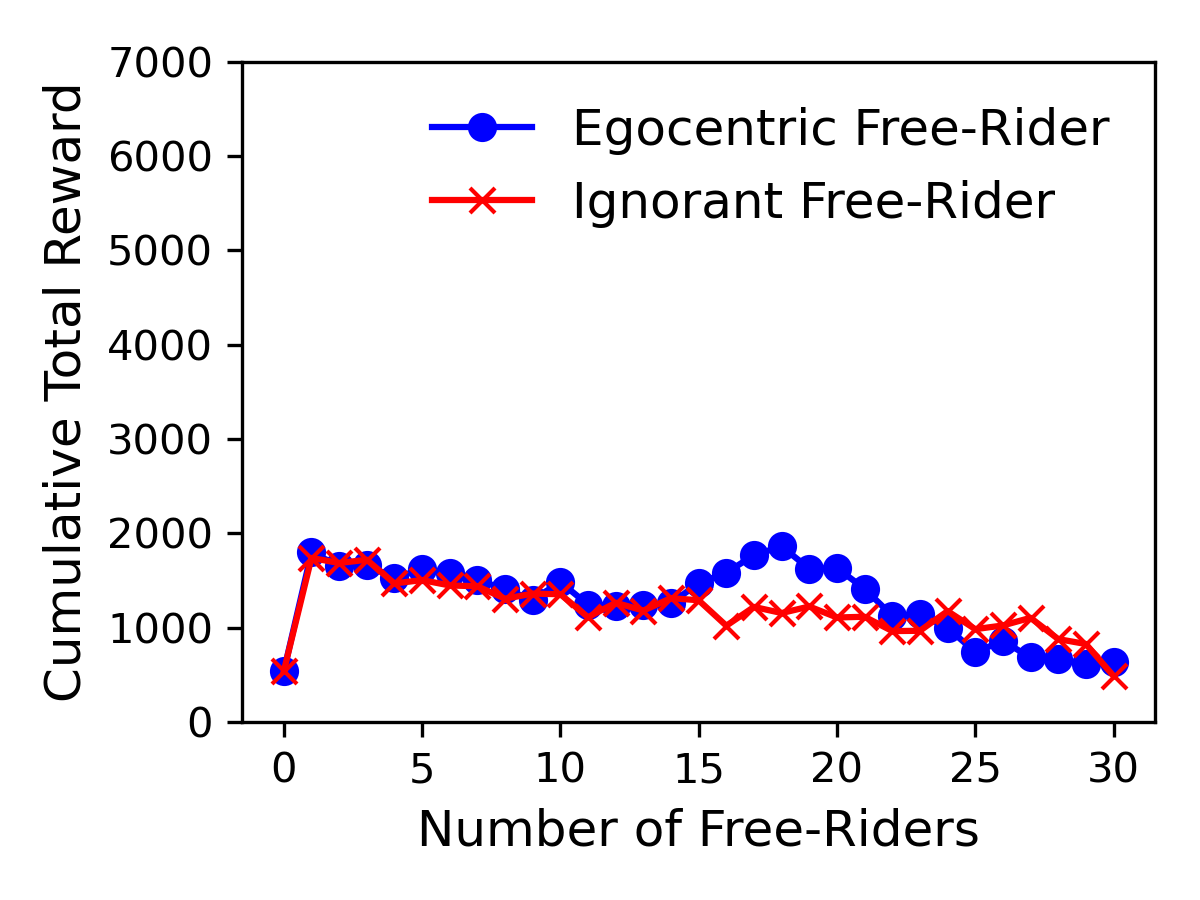}}
    \subfloat[\footnotesize $\tau^r=20$ and $\zeta^s_t=0.15$]{%
        \includegraphics[width=0.45\columnwidth]{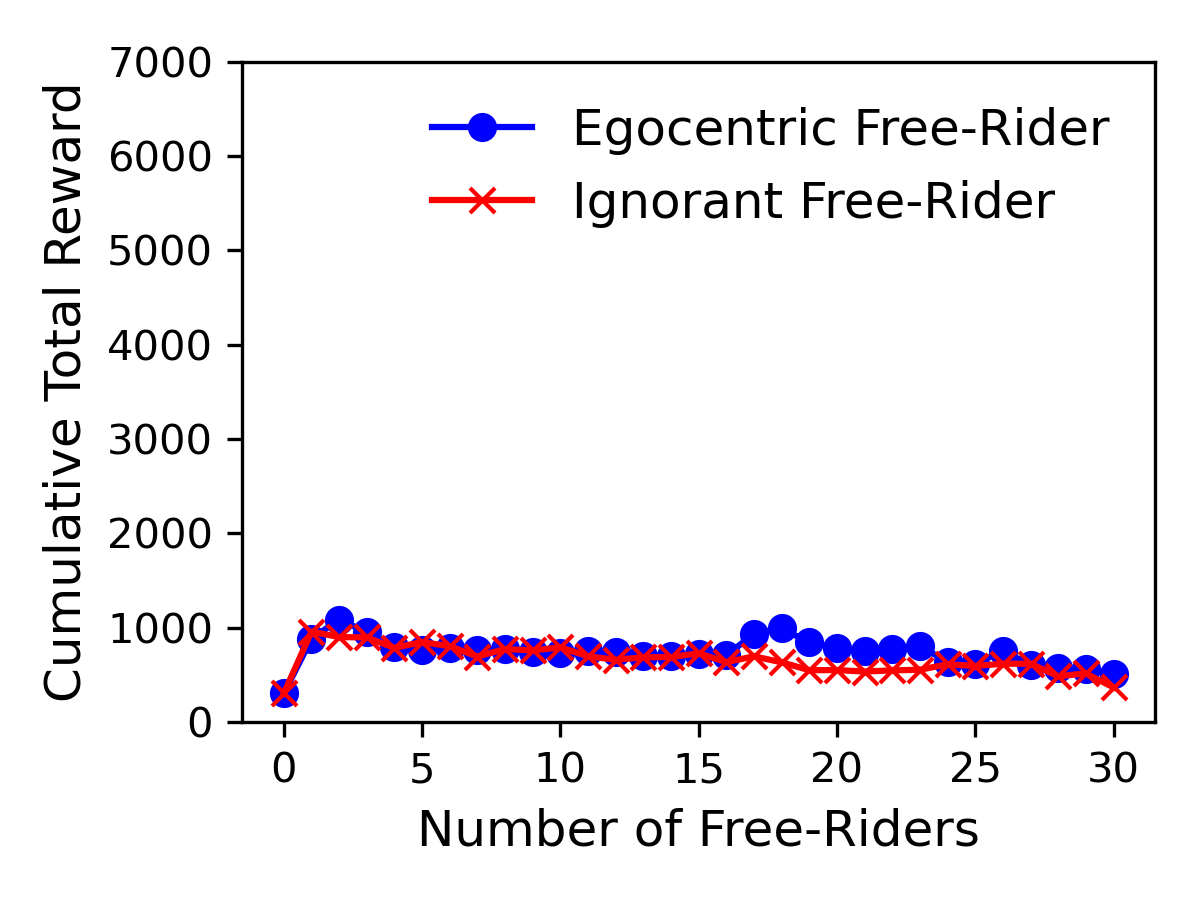}}
    \subfloat[\footnotesize $\tau^r=20$ and $\zeta^s_t=0.20$]{%
        \includegraphics[width=0.45\columnwidth]{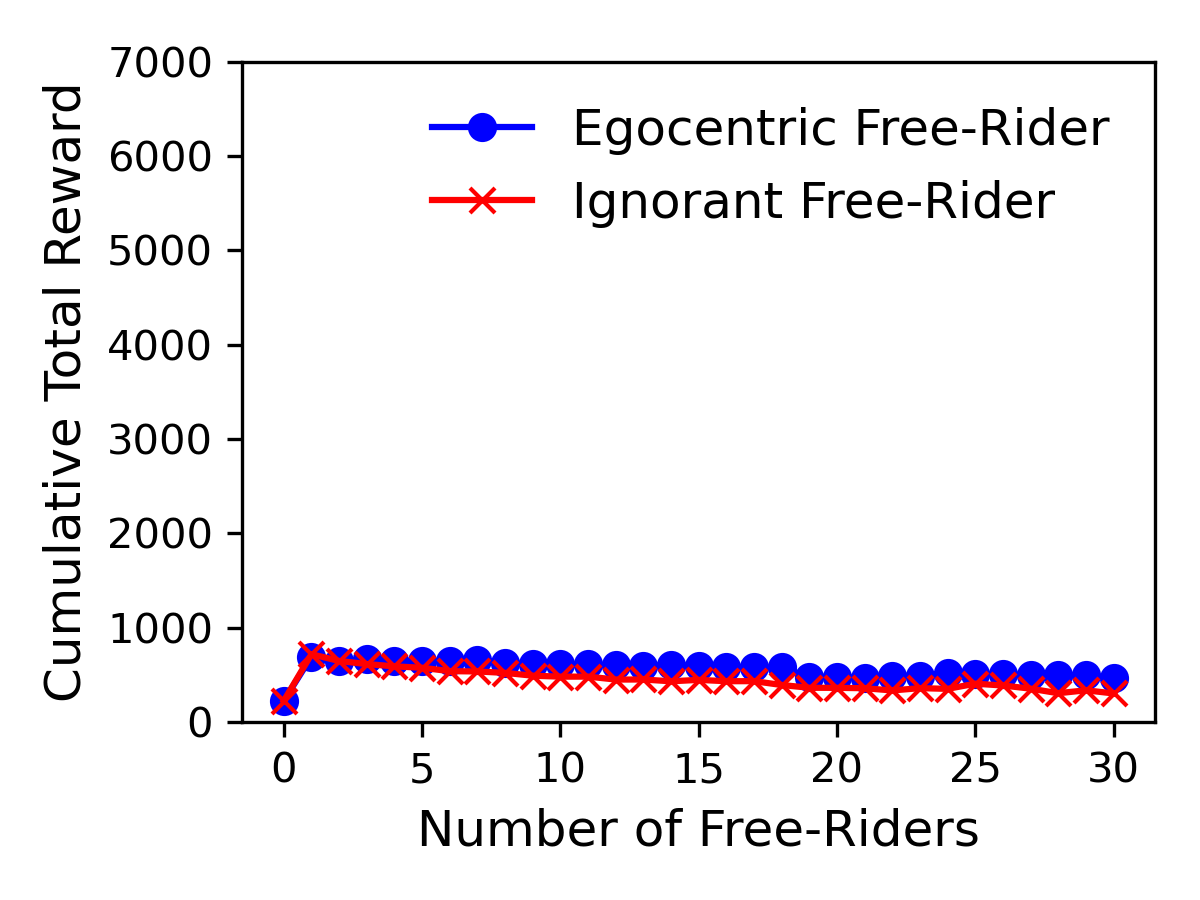}}
	\caption{The cumulative reward in "free-rider" scenarios given various setups of infection rate and recovery time.}
 \label{CumulativeReward2}
\end{figure*}

Fig.~\ref{CumulativeReward2} shows the cumulative total reward of the network simulations driven by mixed preference mutation styles under different infection rates and recovery time values. Given the recovery time at $5$ and $10$, the cumulative total reward decreases when introducing free-riders, which can result from the significant increase of infection numbers (as shown in Fig.~\ref{infectResilience2}). Given the recovery time at $15$ and $20$, the cumulative total reward increases when introducing just one free-rider. The cumulative total reward decreases when we introduce more free-riders. This is because the transformation of the cooperative seed node into a "free-rider" suppresses the cooperative interactions between itself and other infected nodes under a severe epidemic spread.


In summary, under lower infection rates and recovery time, cooperative networks driven by the single cooperative preference mutation styles keep resilient to epidemic spread by suppressing the infection number and preserving the cumulative total reward. In this condition, the introduction of "free-riders" into the cooperation scenario (under a mixed preference mutation style) adversely impact the resilience levels. Under higher infection rates and recovery time, the networks have low resilience levels to more severe epidemic spread despite of their preference mutation styles, either single or mixed. 
The lack of awareness regarding potential epidemic severity escalation during the reinforcement learning phase exposes nodes to greater epidemic risks.

\section{Conclusion}
\label{Conclusion}
This study proposes the Temporal Digital Twin-Oriented Complex Networked System (DT-CNS) model driven by reinforcement learning algorithm. 
Our previous work~\cite{wen2024evolutionary} enables to model DT-CNSs regarding the evolving networks, a dynamic process on the networks and their interrelated changes in DT-CNS. This study extends this framework by modelling temporal interactions given an epidemic outbreak based on the reinforcement learning algorithm.

More specifically, we propose a temporal DT-CNS modelling framework concerning the nodes' temporal decisions on a preference mutation and the social capital limit (maximum number of temporal directed interactions) in response to the interaction patterns and epidemic risks. Under this framework, we consider nodes' heterogeneous features and changeable connection preferences, which combine the effects of preferential attachment and homophily. We introduce different preference mutation styles, which can be cooperative, ignorant and egocentric dependent on their reward designs in the decision-making processes. 
The cooperative nodes maximise their total reward using a collective mind driven by reinforcement learning. The egocentric nodes maximise the individual reward under the assumption of being the only "free-rider" in the cooperation. In contrast, the ignorant nodes make random decisions.

We conduct extensive simulation-based experiments on the temporal DT-CNSs to investigate nodes' temporal infection and interaction patterns in the epidemic outbreak based on heterogeneous mutation styles, including (i) a single preference mutation style where all nodes share the same preference mutation strategy and (ii) a mixed preference mutation style with a varying number of ignorant and egocentric "free-riders" in the cooperation. We also evaluate the resilience of these nodes against intensified epidemic outbreaks by increasing infection rates and extending the recovery time. We find that (i) the full cooperation among nodes results in higher cumulative total rewards and reduced infections compared to the solely egocentric or ignorant "free-rider" scenarios; (ii) an increasing number of "free-riders" in the cooperative networks leads to more infections, while an increasing number of egocentric "free-riders" further escalate the infection numbers; (iii) the fully cooperative networks are resilient to lower infection rates and have shorter recovery times, minimising infections and maintaining rewards, but mixed preference mutation styles in cooperation with "free-riders" weaken the resilience level; (iv) high infection rates and longer recovery times decrease networks' resilience to severe epidemic outbreaks, regardless of preference mutation styles. This implies that the reinforcement learning driven-agents lack awareness about the escalating epidemic severity, exposing nodes to increased epidemic risks.

In summary, this study proposes a temporal DT-CNS framework to model nodes' temporal decisions on their preferences for connecting with others, driven by deep reinforcement learning algorithms. This framework models the temporal directed network interactions while incorporating the impact of epidemic processes. Our future work will enhance this framework by incorporating real-time updates of DT-CNS model components based on mutual real-time information flow between the temporal DT-CNS and real-world data. This will make the DT-CNS a Digital Twin of reality.


\subsection{Acknowledgments}
\noindent This work was supported by the Australian Research Council, Dynamics and Control of Complex Social Networks under Grant DP190101087.



\end{document}